%% file: acl_latex.tex
\pdfoutput=1

\documentclass[11pt]{article}

\usepackage[preprint]{acl}

\usepackage{times}
\usepackage{latexsym}
\usepackage{verbatim}
\usepackage{cleveref}
\crefname{figure}{Fig.}{Figs.}
\crefname{table}{Tab.}{Tabs.}
\crefname{section}{Sec.}{Secs.}

\usepackage[T1]{fontenc}

\usepackage[utf8]{inputenc}

\usepackage{microtype}

\usepackage{inconsolata}

\usepackage{graphicx}

\usepackage[export]{adjustbox}

\usepackage{placeins}

\usepackage{caption}
\usepackage{subcaption}

\usepackage{rotating}

\title{Attribute Diversity Determines the Systematicity Gap in VQA}

\author{Ian Berlot-Attwell\thanks{\hspace{0.2cm}Project lead} \\
  University of Toronto \\
  Vector Institute \\
  \texttt{ianberlot@cs.toronto.edu} \And
  Kumar Krishna Agrawal \\
  University of California, Berkeley \\
  \texttt{kagrawal@berkeley.edu} \And
  A. Michael Carrell \\
  University of Cambridge \\
  \texttt{ac2411@cam.ac.uk} \\\AND
  Yash Sharma\thanks{\hspace{0.2cm}Joint senior authors} \\
  University of T\"ubingen \\
  \texttt{yash.sharma@bethgelab.org} \\\And
  Naomi Saphra\footnotemark[2]\hspace{0.2cm} \\ 
  The Kempner Institute at Harvard University \\
  \texttt{nsaphra@fas.harvard.edu}}

\usepackage{xspace}

\newcommand{\dataset}{CLEVR-HOPE\xspace}
\newcommand{\ho}{HOP\xspace}

\newcommand{\train}{{train}\xspace}
\newcommand{\hoVal}{{complex-IID test}\xspace}
\newcommand{\hoTest}{{complex-OOD test}\xspace}
\newcommand{\atomHo}{{minimal-OOD test}\xspace}
\newcommand{\atomNonHo}{{minimal-IID test}\xspace}

\newcommand{\nmn}{Tensor-NMN\xspace}
\newcommand{\lxmertFt}{LXMERT-p\xspace}
\newcommand{\lxmertScratch}{LXMERT-s\xspace}

\newcommand{\attVal}[1]{\texttt{#1}}
\newcommand{\attType}[1]{\textsc{#1}}

\begin{document}
\maketitle
\begin{abstract}
Although modern neural networks often generalize to new combinations of familiar concepts, the conditions that enable such compositionality have long been an open question. In this work, we study the systematicity gap in visual question answering: the performance difference between reasoning on previously seen and unseen combinations of object attributes. To test, we introduce a novel diagnostic dataset, \dataset. We find that the systematicity gap is not reduced by increasing the quantity of training data, but is reduced by increasing the diversity of training data. In particular, our experiments suggest that the more distinct attribute type combinations are seen during training, the more systematic we can expect the resulting model to be. We release our data and code at \url{https://github.com/ikb-a/systematicity-gap-in-vqa}.
\end{abstract}

\section{Introduction}

{\em Systematicity}, the ability to handle novel combinations of known concepts,
is a type of compositional generalization~\cite{DBLP:journals/jair/HupkesDMB20}. While systematicity is crucial to human intelligence~\cite{fodor1988connectionism}, conventionally trained neural networks often struggle to generalize systematically \citep{csordas2021learning,csordas-etal-2021-devil,csordas2022neural}.

Inspired by prior work investigating compositionality failures in language models~\cite{press-etal-2023-measuring}, we study the \textit{systematicity gap} in visual question answering (VQA): the drop in model performance when reasoning about a combination of properties that was held out from both the text and vision modalities at train time. As an example, let us consider \attType{material} and \attType{shape} as two \emph{attribute types}. If a model was trained without exposure to a particular combination of \textit{attribute values}, e.g., \attVal{rubber} \attVal{sphere}, then we say the model composes systematically if it has high performance at test time on data that includes a \attVal{rubber} \attVal{sphere}. 

Our work empirically demonstrates that systematicity emerges in a neural VQA model if the model is trained with diverse contexts for the attribute values in question (i.e., exposed to many \attType{material}-\attType{shape} combinations). The intuition for this hypothesis is simple: given many training examples of distinct combinations, the model learns how material and shape 
interact, and thus systematically generalizes to an unseen combination of \attType{material} and \attType{shape}. 
In contrast, a model trained on low-diversity data (i.e., only exposed to a few \attType{material}-\attType{shape} combinations) fails to learn rules to recombine them.

Using \dataset, a novel dataset for evaluating systematicity on a variety of held-out object attribute value pairs in a controlled setting, we measure the systematic compositionality of multi-modal transformer and neurosymbolic models. We find that, while systematicity does not improve with more training data, 
it does improve with more \textit{diverse} training data. Specifically, attribute types that include more diverse combinations during training can be composed systematically.

\section{\dataset  Diagnostic Dataset} \label{sec:dataset}
\begin{figure*}[h]
    \centering
    \includegraphics[width=0.9\textwidth]{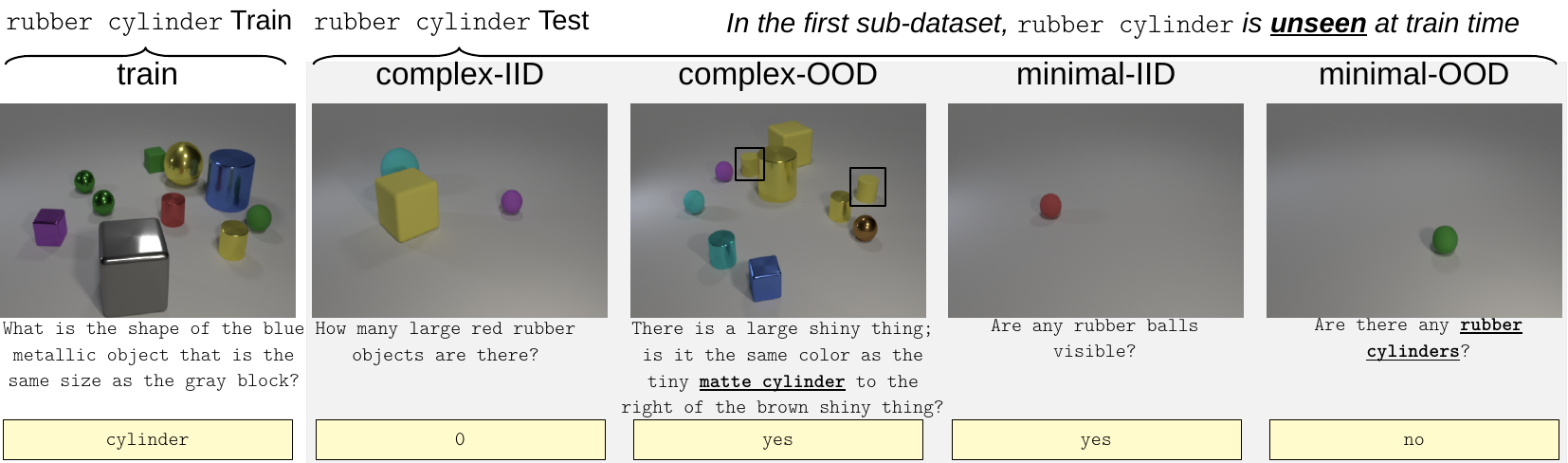}
    \caption{Example image-question pairs for the sub-dataset of \dataset corresponding to \attVal{rubber} \attVal{cylinder}.The 
    test sets are in gray; 
    \attVal{rubber} \attVal{cylinder} is omitted 
    visually \textit{and} textually in the train split and the IID test splits; \attVal{rubber} \attVal{cylinder} only occurs in the OOD splits; 
    occurrences are emphasized in this figure. The train and complex sets are of comparable visual and textual complexity to CLEVR. The minimal sets consist only of existence questions, checking whether a single object matches a given pair of attribute values.}
    \label{fig:hope}
\end{figure*}

Our dataset is based on CLEVR \cite{johnson2017clevr}, a synthetic experimental setting for testing basic visual reasoning skills. CLEVR comprises English questions (such as ``What is the color of the cube on the right side of the yellow sphere?") and corresponding 3D-rendered images of colored blocks. Each block has four attribute types (\attType{size}, \attType{color}, \attType{material}, and \attType{shape}). Our experiments rely on data splits that create distinct in-distribution (ID) and out-of-distribution (OOD) test sets.

We present the CLEVR Held-Out Pair Evaluation (\dataset) dataset for testing the systematicity of VQA models.
\dataset is a controlled setting to test whether VQA models 
generalize to pairs of attribute values that were not seen during either training or fine-tuning.
Within \dataset, we refer to an unseen pair of attribute values as a Held-Out Pair (\ho). The dataset is composed of 29 sub-datasets, each for a different \ho (see \Cref{tab:all_hos} for the list of {\ho}s) 
.

The 29 {\ho}s are selected such that there are 5 {\ho}s from each of the 6 possible pairs of attribute types, with the exception of \attType{Size}+\attType{Material} as CLEVR contains only 4 such combinations.

Each \ho has its own train set and 4 test sets. For \attVal{rubber} \attVal{cylinder},  in~\Cref{fig:hope}, these datasets are:

\noindent{\bf \train}: 560k image-question pairs in the training/finetuning set. The data distribution is similar to CLEVR, but any images or questions involving \attVal{rubber} \attVal{cylinder} have been removed.

\noindent{\bf \hoVal}: Test data sampled from the \train
distribution 
(i.e., \attVal{rubber} \attVal{cylinder} is filtered out). This is a standard IID test set; it's primarily used as a point of comparison for the \hoTest split.

\noindent{\bf \hoTest}: Test data sampled from the CLEVR distribution filtered to always have (i) at least one object matching \attVal{rubber} \attVal{cylinder}, and (ii) \attVal{rubber} \attVal{cylinder} in the question. It is OOD as \attVal{rubber} \attVal{cylinder} is a combination never seen visually or textually in the train data. This split requires the model to behave systematically (i.e., generalize to \attVal{rubber} \attVal{cylinder}) while reasoning over scenes with several objects.

\noindent{\bf \atomNonHo}: Minimal image-question pairs that check whether a model can recognize pairs of attribute values, corresponding to \attVal{rubber} \attVal{cylinder}'s attribute types, that were seen in the \train set. The scene contains a single object, and the question only asks if there is an object matching a given pair of attributes. This split is called IID as it only contains pairs of attribute seen at train time. This split provides a point of comparison for the \atomHo split. 

\noindent{\bf \atomHo}: Minimal image-question pairs that check recognition of 
\attVal{rubber} \attVal{cylinder}. This split tests a model's systematicity independent of reasoning or visual clutter. If a model had some systematic behaviour which degraded when reasoning was required, then we'd expect to see a small performance gap between \atomHo and \atomNonHo, and a large gap between \hoTest and \hoVal.
By construction, always returning false would yield 75\% accuracy (See Appx.~\ref{sec:atom} for details).

Summarizing the split naming convention: ``complex'' splits contain CLEVR-like visual and textual complexity. ``minimal'' splits contain single-object images, and only recognition questions (e.g., ``Are any rubber cylinders visible?''). IID splits contain only the attribute pairs seen at train time. OOD splits contain the HOP in both the question, and in at least one object in the scene. 

\Cref{sec:dataset_details} includes dataset details. 
Note, \dataset omits validation sets to prevent tuning for specific task~\cite{teney2020value}; instead, hyperparameters should be chosen using CLEVR. 

\section{Models \& Training}
\input{model_train}

\section{Results}
\subsection{Evidence of Systematic Behaviour}\label{sec:evidence_of_sys} 
\input{systematicity}

\input{diversity.tex}

\section{Related work}

While compositionality in VQA has been studied, prior work has focused on generalization to new question structures ~\cite{bahdanau2019closure,vani2021iterated,bogin-etal-2021-covr}, task-object combinations~\cite{DBLP:conf/cvpr/WhiteheadWJFS21}, or question-answer combinations~\cite{agrawal2017c}, rather than new attribute combinations. Systematicity has often been investigated through synthetic datasets to control for the model's exposure to particular attribute combinations. ~\citet{lake2018generalization} introduced the SCAN benchmark to evaluate compositionality in sequence-to-sequence models, revealing a lack of systematicity. 
Followup~\cite{patel-etal-2022-revisiting, jiang-etal-2022-mutual} and concurrent~\cite{zhou-etal-2023-data} seq2seq works have shown that 
the conceptual diversity of the training set significantly affects systematicity---our work extends these findings to the multi-modal domain of 
VQA.

The closest prior work 
is the CLEVR-CoGenT dataset: ~\citet{johnson2017clevr} created a train-test CLEVR split where at train time 
cubes and cylinders are restricted to limited color palettes, that are
reversed at test time. They observed that model performance declined on held-out attribute combinations. 
But, unlike \dataset, CLEVR-CoGenT does not change the question distribution at train time--- 
held-out combinations can leak by appearing in text at train time. 
Furthermore, CLEVR-CoGenT has only a single train set with held-out \attType{color}-\attType{shape} combinations---whereas \dataset expands the set of held-out combinations to 29 train sets, covering all possible pairs of attribute types.  
\dataset also independently assesses each \ho, including in a minimal setting. 
In combination, these improvements allow us to study the impact of train-time diversity.

Our results align with concurrent work on the effects of training diversity in VQA: \citet{rahimi2023d3} modify CLEVR to study the 
related question of productivity, 
 concluding that increasing the diversity of question combinations increases productivity. Unlike our work, they do not use a transformer architecture, instead studying MAC~\cite{DBLP:conf/iclr/HudsonM18}, FiLM~\cite{DBLP:conf/aaai/PerezSVDC18}, and Vector-NMN~\cite{bahdanau2019closure}. Additionally, as they study a fundamentally different question, their dataset only alters the question distribution---their image distribution is unchanged between train and test time. Given that systematicity and productivity are both aspects of compositional generalization~\cite{DBLP:journals/jair/HupkesDMB20}, the growing evidence across task settings and facets of compositionality \citep{oren-etal-2021-finding,levy-etal-2023-diverse} suggests 
a 
close relationship between train-time diversity and compositional generalization as a broad phenomenon.

We encourage further study, and release our code and data to facilitate these efforts. Confirming our findings in newer vision-language models, quantifying them with scaling laws, exploitation in data curation pipelines, and using these effects to predict model reliability, 
are all important future work. 

\section{Conclusions} 
Using \dataset, we demonstrate that several models exhibit a degree of systematic generalization to held-out object attribute pairs. Furthermore, we illustrate that 
the systematicity gap (the difference between in- and out-of-distribution performance) does not improve with more data, but does with more attribute diverse data
---i.e., the number of attribute pairs of the same type seen at train time. 

\FloatBarrier

\section*{Limitations}

First and foremost, while the synthetic nature of \dataset allows for a more controlled study of models, 
it raises the question whether the observed results will hold in other vision-language tasks, or in more complex and diverse real-world settings. Based on similar findings in text domains~\cite{patel-etal-2022-revisiting, jiang-etal-2022-mutual,zhou-etal-2023-data, oren-etal-2021-finding,levy-etal-2023-diverse}, and evidence of a link between productivity and diversity in VQA \cite{rahimi2023d3}, we anticipate that our findings would apply in other multimodal settings like image or caption generation -- although we leave that to future work. The investigation of scaling laws in which these diversity requirements breakdown is an important direction for future work.

The second major limitation arises from the choice of models. LXMERT uses a pretrained F-RCNN \cite{NIPS2015_14bfa6bb} for object detection, which we do not alter. As the F-RCNN is pretrained, it may already possess implicit knowledge of the attributes (e.g., shape), and may contribute systematic structure to LXMERT. Any such visual knowledge or biases are therefore given to both \lxmertFt and \lxmertScratch. In contrast, note that the language component of \lxmertScratch is randomly initialized---whereas~\cite{tan-bansal-2019-lxmert} initialized their language transformer with BERT \cite{devlin-etal-2019-bert} when pretraining from scratch. Similarly, Tensor-NMN uses a frozen pretrained ResNet \cite{DBLP:conf/cvpr/HeZRS16} as its vision backbone, and its language components and modules are initialized from scratch. A related limitation is that \lxmertFt may have been exposed to the held-out attribute during its pretraining phase; we control for this via the \lxmertScratch experiments where no vision-language pretraining is performed. 

More generally, the models we studied are not the strongest. However, despite this we find that they yield high accuracy across the board (see Appx. Fig.~\ref{fig:matrix_acc_ho0_5}), which indicates to us that analyzing the systematicity gap for these models still provides meaningful insight. While we do not know for certain whether our results will extrapolate to stronger models, we recommend analysis for stronger models as an important direction for future work, and enable such by releasing our data and code. The literature studying the impact of data distribution on learning often proposes and demonstrates an effect in toy settings initially, and it is this area that our paper falls within.

Beyond stronger models, it may be interesting to confirm our findings on other NMN variants such as N2NMN~\cite{DBLP:conf/iccv/HuARDS17}, or on scene-graph based architectures such as XNM~\cite{DBLP:conf/cvpr/ShiZL19} or the Neural State Machine~\cite{DBLP:conf/nips/HudsonM19} which were also designed to encourage compositionality.

Finally, due to resource limitations, we only evaluate Tensor-NMN on 6 of the 29 total {\ho}s, one for each attribute type combination.

\section*{Ethics Statement}

We judge that our work has very low risk. The primary risk is of using our dataset to measure model systematicity in models that are not trained on our train/test split. We have provided a highly specific diagnostic dataset that is designed to provide a data split for testing generalization claims, and our OOD set is not useful to measure generalization in arbitrary VQA models. This concern is documented in the dataset datasheet in Section \ref{sec:dataset_datasheet} of the Appendix.

Our dataset has the disclosed bias that it only contains English questions, however this is reasonably overshadowed by the synthetic nature of the text (see the Limitations section). We also believe that given the functional forms of the questions, it should be fairly straightforward to convert the functional forms into synthetic natural language equivalents for a given target language. As to the images, any cultural biases in the specific choices of shape, colour, size, materials, position, and count of the objects are inherited from the original CLEVR \cite{johnson2017clevr}. 

Apart from the dataset, our contributions are findings on the impact of train-time diversity on model systematicity. To the best of our understanding, this finding does not introduce any new capabilities to the models in question, nor does it affect the accessibility or impact of these models.

\FloatBarrier 

\section*{Acknowledgements}

Resources used in preparing this research were provided, in part, by the Department of Computer Science at the University of Toronto, the Province of Ontario, the Government of Canada through CIFAR, companies sponsoring the Vector Institute (\url{www.vectorinstitute.ai/partnerships/current-partners/}), the Hyundai Motor Company (under the project Uncertainty in Neural Sequence Modeling), the Samsung Advanced Institute of Technology (under the project Next Generation Deep Learning: From Pattern Recognition to AI), and by a gift from the Chan Zuckerberg Initiative Foundation to establish the Kempner Institute for the Study of Natural and Artificial Intelligence.

Ian Berlot-Attwell is funded by a Natural Sciences and Engineering Research Council of Canada Postgraduate Scholarship-Doctoral, and a Vector Institute Research Grant. A. Michael Carrell is funded in part by a Microsoft Research scholarship. The authors thank the International Max Planck Research School for Intelligent Systems (IMPRS-IS) for supporting Yash Sharma. 

We appreciate the invaluable ideas and discussion from Preetum Nakkiran, Spencer Frei, Nicholas Schiefer, and Ferenc Huszár who helped shape the early stages of this work, and Preetum in particular for enabling this collaboration. We would also like to thank Jonathan Shi for his early contributions to our code base, including help with, and modification to, the CLEVR dataset generation code.   
We would also like to thank Frank Rudzicz for being a wonderful PhD supervisor, and for his suggestions and advice throughout the work.

\FloatBarrier

\bibliography{anthology,custom}

\appendix
\section{Extended Related Work}
\label{extended-rw}

While compositionality in VQA has been studied, prior work has focused on generalization to new question structures ~\cite{bahdanau2019closure,vani2021iterated,bogin-etal-2021-covr}, task-object combinations~\cite{DBLP:conf/cvpr/WhiteheadWJFS21}, or question-answer combinations~\cite{agrawal2017c}, rather than new attribute combinations. One reason for this gap is that, with natural data, it is hard to control for the model's exposure to particular attribute combinations. 
By using a controlled synthetic setting, we can guarantee that generalization behavior is systematic based on the data split.

Systematicity has often been investigated through synthetic datasets. ~\citet{lake2018generalization} introduced the SCAN benchmark to evaluate compositionality in sequence-to-sequence models, revealing a lack of systematicity. 
Followup~\cite{patel-etal-2022-revisiting, jiang-etal-2022-mutual} and concurrent~\cite{zhou-etal-2023-data} seq2seq works have shown that 
the conceptual diversity of the training set significantly affects systematicity---our work extends these findings to the multi-modal domain of 
VQA.

The closest prior work 
is the CLEVR-CoGenT dataset: ~\citet{johnson2017clevr} created a train-test CLEVR split where at train time 
cubes and cylinders are restricted to limited color palettes, that are
reversed at test time. They observed that model performance declined on held-out attribute combinations. 
But, unlike \dataset, CLEVR-CoGenT does not change the question distribution at train time---
held-out combinations can leak by appearing in text at train time. 
Furthermore, CLEVR-CoGenT has only a single train set with held-out \attType{color}-\attType{shape} combinations---whereas \dataset expands the set of held-out combinations to 29 train sets, covering all possible pairs of attribute types.  
\dataset also independently assesses each \ho, including in a minimal setting. 
In combination, these improvements allow us to study the impact of train-time diversity.

Beyond CLEVR-CoGenT, our results align with concurrent work on the effects of training diversity in VQA: \citet{rahimi2023d3} modify CLEVR to study the 
related question of productivity. Specifically, generalization to questions with more reasoning steps, and generalization to new question combinations (e.g., answering counting questions about \attVal{shape}, when all train-time counting questions are about \attVal{color} or \attVal{size}). They conclude that increasing the diversity of question combinations increases productivity. Unlike our work, they do not use a transformer architecture, instead studying MAC~\cite{DBLP:conf/iclr/HudsonM18}, FiLM~\cite{DBLP:conf/aaai/PerezSVDC18}, and Vector-NMN~\cite{bahdanau2019closure}. Additionally, as they study a fundamentally different question, their dataset only alters the question distribution---their image distribution is unchanged between train and test time.

Given that systematicity and productivity are both aspects of compositional generalization~\cite{DBLP:journals/jair/HupkesDMB20}, the growing evidence across task settings and facets of compositionality \citep{oren-etal-2021-finding,levy-etal-2023-diverse} suggests 
a 
close relationship between train-time diversity and compositional generalization as a broad phenomenon.
\section{\dataset: Additional details}
\label{sec:dataset_details}

The full list of held-out pairs ({\ho}s) can be found in Table \ref{tab:all_hos}. The {\ho}s were selected by choosing two attribute values from each of \attVal{large cyan rubber cylinder}, \attVal{small brown rubber sphere}, \attVal{small red metal cylinder}, \attVal{large gray metal cube}, and \attVal{small purple rubber sphere}. 

Note that there are only 4 possible \attType{material}-\attType{size} combinations, as there are only 2 \attType{sizes} and 2 \attType{materials}. We include all 4 of these, as well as 5 {\ho}s for every other pair of attribute types.

Before selecting the 5 4-tuples from which we created the {\ho}s in \dataset, we first created a small set of minimal test questions for testing how well a given model comprehends a given attribute in isolation---CLEVR-PRELIM. For example, for the color \attVal{cyan} we had two types of tests. First, tests similar to the \atomHo tests (i.e., a single object and rephrasings of ``Are any cyan objects visible?''). Second, counting tests---all questions were rephrases of ``What number of cyan objects are there?'', and images had varying numbers of \attVal{cyan} objects. Specifically, we fixed the position of 5 objects, and created 6 images, each with a different number of objects matching the attribute---i.e., 0, 1, 2, 3, 4, or 5 \attVal{cyan} objects. 

Note that, unlike \dataset which studies pairs of attributes values, CLEVR-PRELIM evaluates \textit{only} attribute values in isolation.

Using CLEVR-PRELIM, we performed a zero-shot evaluation of \citet{tan-bansal-2019-lxmert}'s VQA2.0 \cite{balanced_vqa_v2} fine-tuned LXMERT checkpoint. From this preliminary study we found that zero-shot model performance was generally poor (e.g., over all attribute values of all types, the highest count performance was 49.1\%). Given our interest in studying the impact of the amount of training data, we created our first 4-tuple by individually selecting each attribute value; specifically choosing the attribute value that zero-shot LXMERT had the lowest performance on---this created the 4-tuple \attVal{Large cyan rubber cylinder}. The remaining four tuples were selected uniformly at random. Ultimately, as we did not see any significant difference between a small sample of 6 {\ho}s (those created from attribute pairs in \attVal{large cyan rubber cylinder}) and a larger sample of 23  {\ho}s (those created from random 4-tuples), we present results aggregated over all 29 {\ho}s.

Note that as two 4-tuples were \attVal{rubber} \attVal{spheres} and \attVal{small spheres}, we added the {\ho}s \attVal{rubber cube} and \attVal{small cube} so that we would maintain five \attType{material}-\attType{shape} and five \attType{size}-\attType{shape} pairs.

For each \ho in \dataset, the approximate size of the corresponding splits is outlined below:

\begin{itemize}
    \item \train set: 62k images, and 560k image-question pairs
    \item \hoVal set: 13k images, 120k image-question pairs
    \item \hoTest set: 15k images, 15k image-question pairs
    \item \atomNonHo set: 2576-3200 images, 8640-11970 image-question pairs (depending on \ho)
    \item \atomHo set: 448-3840 images, 448-3840 image-question pairs (depending on \ho) 
\end{itemize}

\noindent To reduce the resources required to generate the dataset, images are reused throughout the dataset. Specifically, the images are reused across the \train sets for the {\ho}s, and reused from the original CLEVR \cite{johnson2017clevr} training set.

Similarly, each of the test sets reuse images across {\ho}s. Note that while the \hoVal and \hoTest sets do not reuse eachother's images, the \atomNonHo and \atomHo sets do for images that do not involve the \ho under consideration.

To ensure that CLEVR can be fairly used for hyperparameter tuning, and to prevent any data leakage, \textit{no} CLEVR validation or test images are reused in CLEVR-HOPE.

For further information, including distribution and maintenance, see the CLEVR-HOPE Datasheet in Section \ref{sec:dataset_datasheet}. The datasheet follows the format outlined by \citet{gebru2021datasheets}, and is modified from the template by \citet{datasheetTemplate}.

\subsection{\dataset: \atomHo set and \atomNonHo set}
\label{sec:atom}

All images in the \atomHo and \atomNonHo sets contain only a single object. All questions ask whether there are any objects matching the attribute value pair. E.g., for  
the \ho \attVal{rubber cyan}, some question variants include ``Are there any cyan matte things?'' and ``Are any cyan matte things visible?''.

\begin{figure*}[h]
    \centering
    \includegraphics[width=\textwidth]{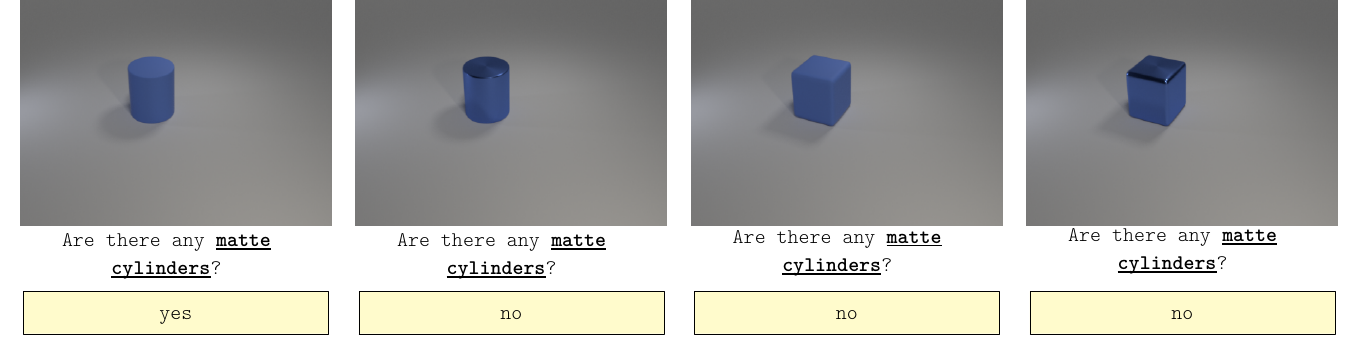}
    \caption{Four example image-question pairs for the \atomHo split of the sub-dataset of \dataset corresponding to the first held-out attribute pair---i.e., \attVal{rubber} \attVal{cylinder}. Note how the first image matches \attVal{rubber} \attVal{cylinder} (\attType{material}=\attVal{rubber}, and \attType{shape}=\attVal{cylinder}), and the next three image have one attribute value (\attType{material}=\attVal{metal}), the other attribute value (\attType{shape}=\attVal{cube}), or both (\attType{material}=\attVal{metal}, and \attType{shape}=\attVal{cube}) attribute values not matching \attVal{rubber} \attVal{cylinder}. This pattern repeats throughout the dataset, with the choice of distractor values, object position, lightning, question-phrasing and the value of the attribute-types not in \ho, all chosen randomly, but fixed within each set of 4 images.}
    \label{fig:atom_dataset}
\end{figure*}

These splits are designed to test the model in a systematic manner: each image matching the \ho has 3 corresponding images that do not match the \ho. These 4 images share identical question phrasing. The non-matching images maintain the object position, lighting, and the attribute values that are irrelevant to the \ho, but change the first attribute value in the \ho, the second attribute value in the \ho, or both attribute values in the \ho, respectively. See~\Cref{fig:atom_dataset} for an example.

Note that the question template is taken directly from the original CLEVR dataset generation code. The main change is the aforementioned systematic design, and that the images used contain only a single object, whereas the original CLEVR requires at least 3 objects in any scene.

The \atomNonHo split is created in the same way, but testing all other attribute-value pairs of the same type as the \ho. Note that the distractor attribute values in the negative examples were selected uniformly at random. Since this may create the held-out pair (and indeed, \textit{must} do so for one of the four size-material images), after the initial creation of the \atomNonHo split, we filter it to remove any image-question pairs where the object in the image matches the \ho.

\section{Training details}
\label{sec:train_details}

All subsets of the \train sets (i.e., of size 25k, 200k, and 560k) are created by taking the first however many indices. This corresponds to a random subset of images for 25k, which is consecutively randomly expanded. This is so because the image-question pairs are unsorted, apart from all questions for any given image having contiguous indices. Note that we fix the number of gradient updates across subset sizes, i.e., smaller subsets are trained for more epochs so that the total number of gradient updates is the same.

\begin{table}
    \centering
    \begin{tabular}{c|c|c}
         Hyperparameter & \lxmertFt & \lxmertScratch \\
         \hline 
         Learning Rate & 5e-5 & 1e-5 \\
         Gradient Updates & 218,750 & 481,000 \\
         Batch size & 32 & 32 \\
    \end{tabular}
    \caption{Key hyperparameter values used for LXMERT}
    \label{tab:hyperparameters}
\end{table}

For LXMERT, the maximum sequence length is increased to 49 so that \dataset questions are not truncated.

For \lxmertFt, we follow~\citet{tan-bansal-2019-lxmert}'s procedure for finetuning their pretrained LXMERT checkpoint on a VQA dataset. As part of their procedure, the pretrained F-RCNN \cite{NIPS2015_14bfa6bb} object detector is \textit{not} altered in any way. 

\lxmertFt hyperparameters were modified from the hyperparameters used by \citet{tan-bansal-2019-lxmert} for finetuning LXMERT for VQA. Specifically, \citet{tan-bansal-2019-lxmert} finetuned LXMERT for the VQA tasks of VQAv2 \cite{balanced_vqa_v2}, NLVR2 \cite{suhr-etal-2019-corpus}, and GQA \cite{hudson2019gqa} with a batch size of 32, 4 epochs, and a learning rate of either 1e-5 or 5e-5. We ultimately used a learning rate of 5e-5, and increased the epochs to 10 as we found it yielded better performance.

For \lxmertScratch we randomly initialize all LXMERT weights (this \textit{excludes} the pretrained F-RCNN object detector), and apply the LXMERT finetuning procedure (albeit with different hyperparamters) to train this randomly initialized model. 

Both LXMERT models contain 209 million trainable parameters, in addition to the frozen F-RCNN object detector (65 million frozen parameters).

\lxmertScratch hyperparameter tuning was performed via grid search over learning rate (1e-4, 5e-5, 1e-5) and training steps (218750, 481000, 700000). Note that we ultimately used 481k gradient update steps, as its validation accuracy (95.47\%) was extremely close to 700k (96.99\%), with nearly half the training time.

The LXMERT hyperparameters used are summarized in~\Cref{tab:hyperparameters}.

\nmn is trained from scratch following the process used by \citet{bahdanau2019closure}.Following their work, image features are extracted from the conv4 layer of a frozen ResNet101 \cite{DBLP:conf/cvpr/HeZRS16}. 
\nmn is trained in a 3 stage process---initially the program generator and execution engine are trained in a supervised manner, following which they are trained together using REINFORCE. The default hyperparameters for CLEVR from \citet{bahdanau2019closure} are used. 

The \nmn model contains 42 million trainable parameters, in addition to the frozen ResNet101 image feature extractor (27 million frozen parameters -- less than the full ResNet101 as only the conv4 features are used).

Models were trained on a mixture of 16GB Nvidia Tesla T4 GPUs, and 8GB Nvidia GeForce RTX 2070 GPUs. Each run was trained on a single GPU, with the experiments spread over approximately 44 GPUs. We upper bound the number of GPU hours of compute used at approximately 24k, 32k, and 66k for the \lxmertFt, \lxmertScratch and \nmn experiments respectively.

\section{LXMERT Detailed Results}

\begin{figure}[h]
    \centering
    \includegraphics[width=0.5\textwidth]{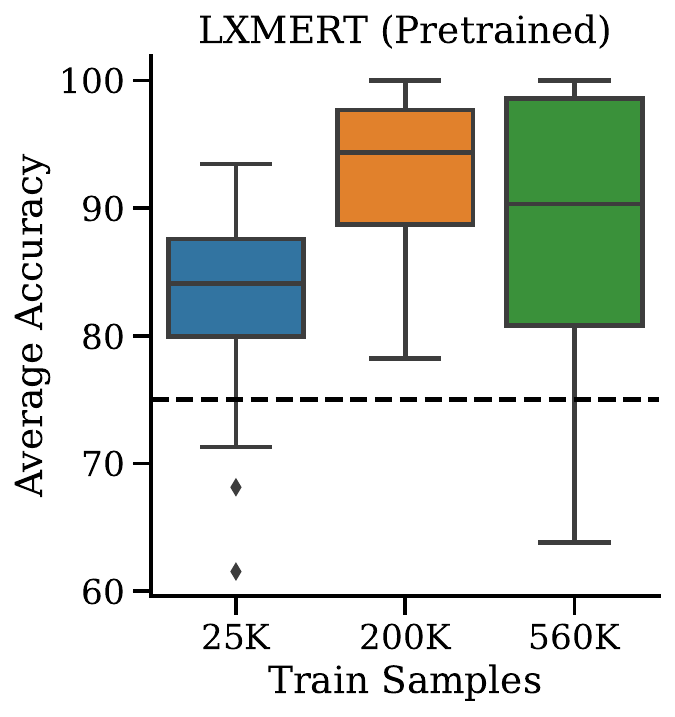}\\
    \includegraphics[width=0.5\textwidth]{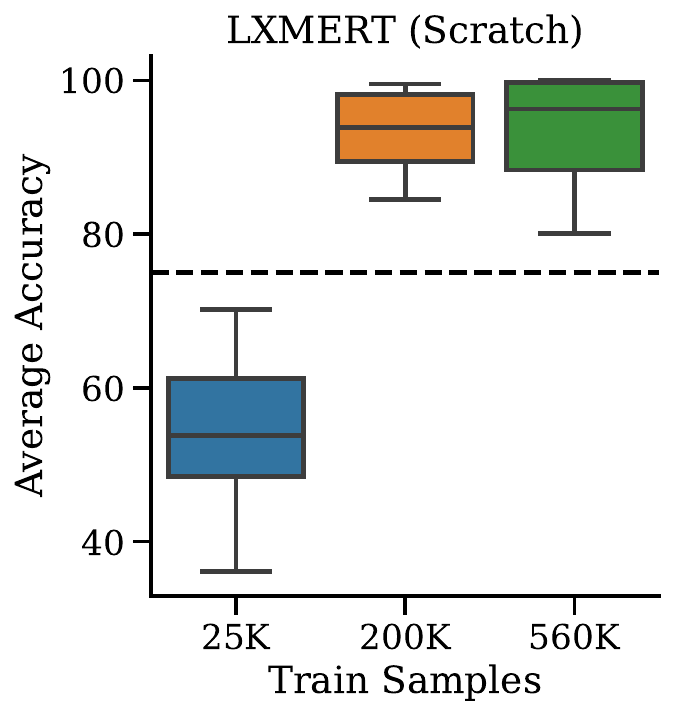}\\
    \caption{Box plot of \textbf{\atomHo} set performance on all 29 {\ho}s. The average performance for each \ho is produced by averaging over 3 trials. The variation captured by this boxplot is from the difference in average performance between {\ho}s, rather than from the variation within the 3 trials.}
    \label{fig:atom_ho}
\end{figure}

\begin{figure}[h]
    \centering
    \includegraphics[width=0.5\textwidth]{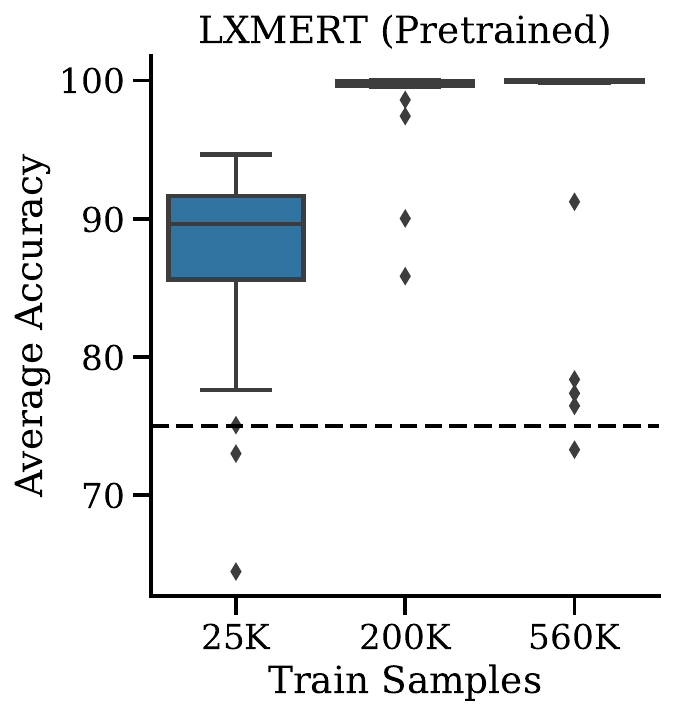}\\
    \includegraphics[width=0.5\textwidth]{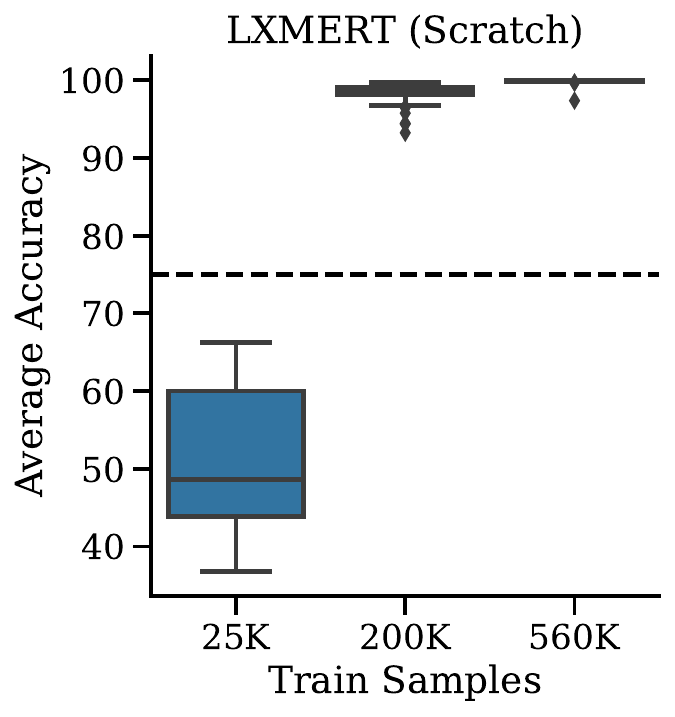}\\
    \caption{Box plot of \textbf{\atomNonHo} set performance on all 29 {\ho}s. The average performance for each \ho is produced by averaging over 3 trials. The variation captured by this boxplot is from the difference in average performance between {\ho}s, rather than from the variation within the 3 trials.}
    \label{fig:atom_non_ho}
\end{figure}

LXMERT performance on \atomHo can be found in~\Cref{fig:atom_ho}. Performance on \atomNonHo can be found in~\Cref{fig:atom_non_ho}. All plots mark 75\%---this baseline performance is achieved on the \atomHo split by always predicting false (i.e., the most common class). Always predicting false on \atomNonHo yield a baseline performance between 66\% and 75\%, depending on the \ho. 

\begin{figure}[h]
    \centering
    \includegraphics[width=0.5\textwidth]{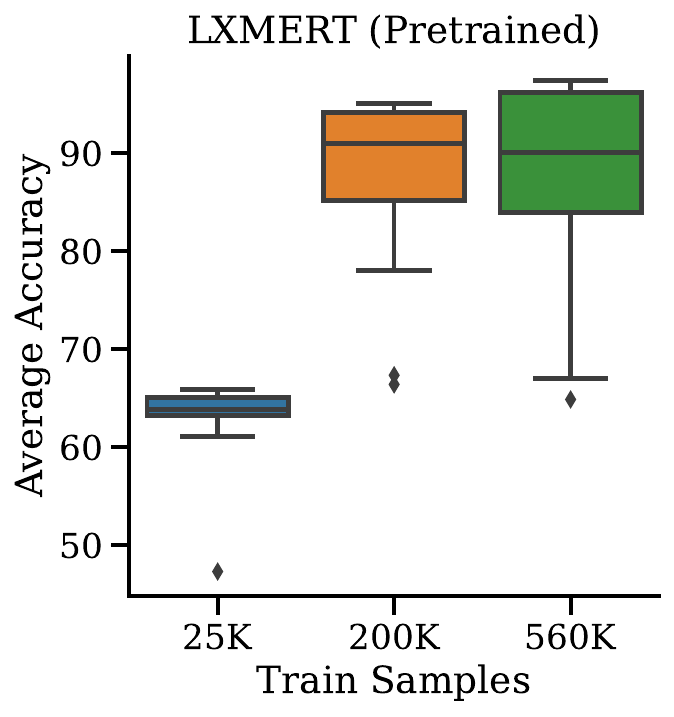}\\
    \includegraphics[width=0.5\textwidth]{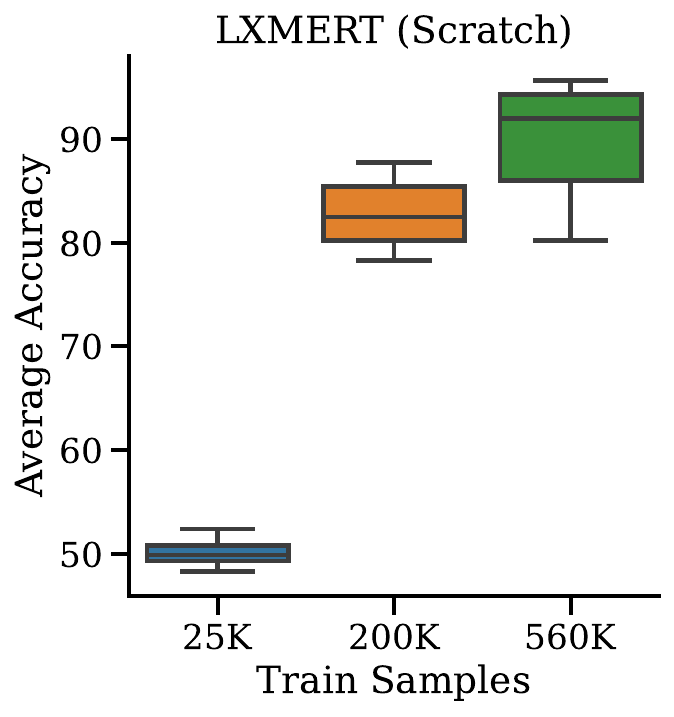}\\
    \caption{Box plot of \textbf{\hoTest} set performance on all 29 {\ho}s. The average performance for each \ho is produced by averaging over 3 trials. The variation captured by this boxplot is from the difference in average performance between {\ho}s, rather than from the variation within the 3 trials.}
    \label{fig:ho_test}
\end{figure}

\begin{figure}[h]
    \centering
    \includegraphics[width=0.5\textwidth]{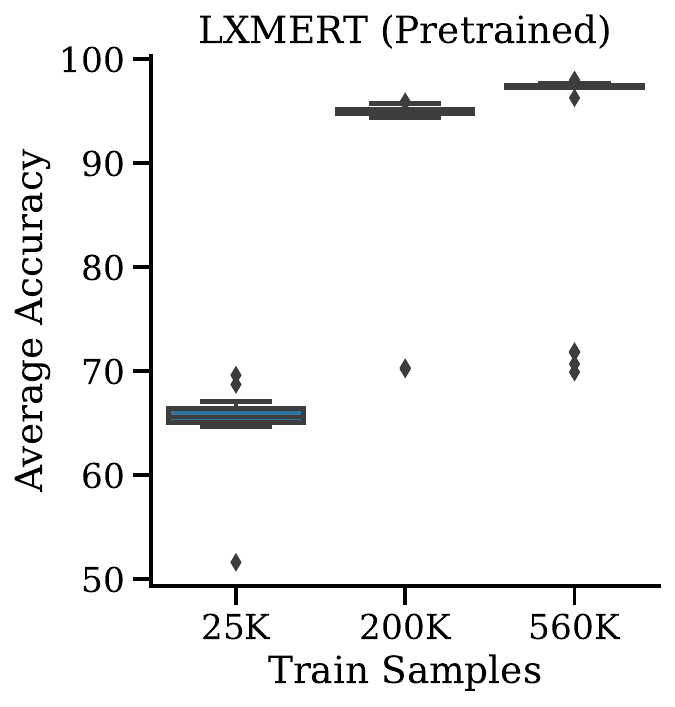}\\
    \includegraphics[width=0.5\textwidth]{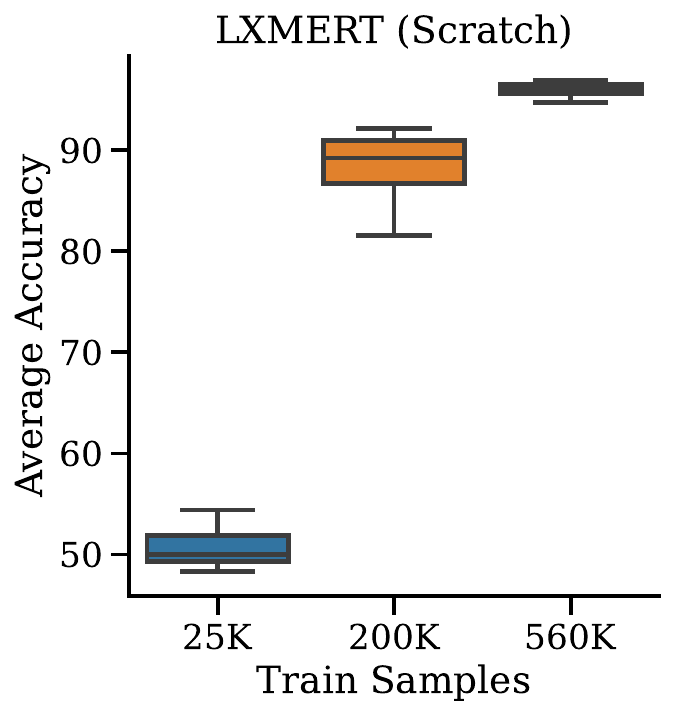}\\
    \caption{Box plot of \textbf{\hoVal} set performance on all 29 {\ho}s. The average performance for each \ho is produced by averaging over 3 trials. The variation captured by this boxplot is from the difference in average performance between {\ho}s, rather than from the variation within the 3 trials.}
    \label{fig:ho_val}
\end{figure}

LXMERT performance on \hoTest can be found in~\Cref{fig:ho_test}. Performance on \hoVal can be found in~\Cref{fig:ho_val}. 

\begin{figure*}[h]
    \centering
    \includegraphics[width=0.8\textwidth]{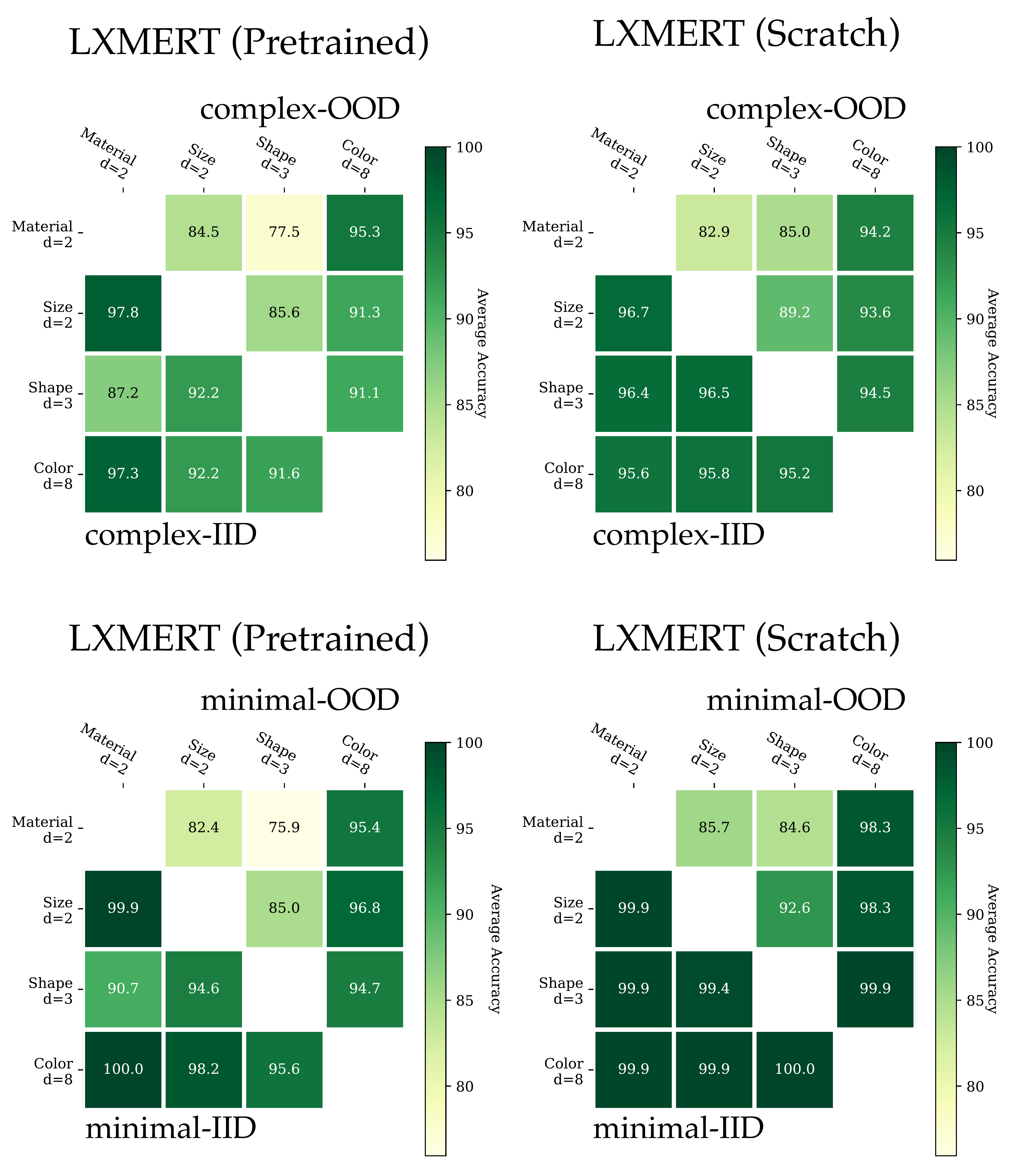}
    \caption{Model accuracies for {\ho}-0 through 28. Note that the LXMERT models often struggle on both IID and OOD questions when \attType{Material}-\attType{Shape} combinations are held out at train time.}
    \label{fig:matrix_acc_ho0_28}
\end{figure*}

For LXMERT trained on the largest \train sets (560k), we plot the complex and minimal model accuracies, averaged by the attribute types of the {\ho}s, in~\Cref{fig:matrix_acc_ho0_28}.

The exact average accuracies and standard deviations over 3 runs are in Tables~\ref{tab:summary_lxmert_ft_acc_aggregate_val} through \ref{tab:summary_lxmert_scratch_acc_aggregate_atom-ho}.

\section{\nmn Detailed Results}

As \nmn was only evaluated on the first 6 {\ho}s, we include the subset of LXMERT models trained on the same {\ho}s for comparison.

\begin{figure}[h]
    \centering
    \includegraphics[width=0.5\textwidth]{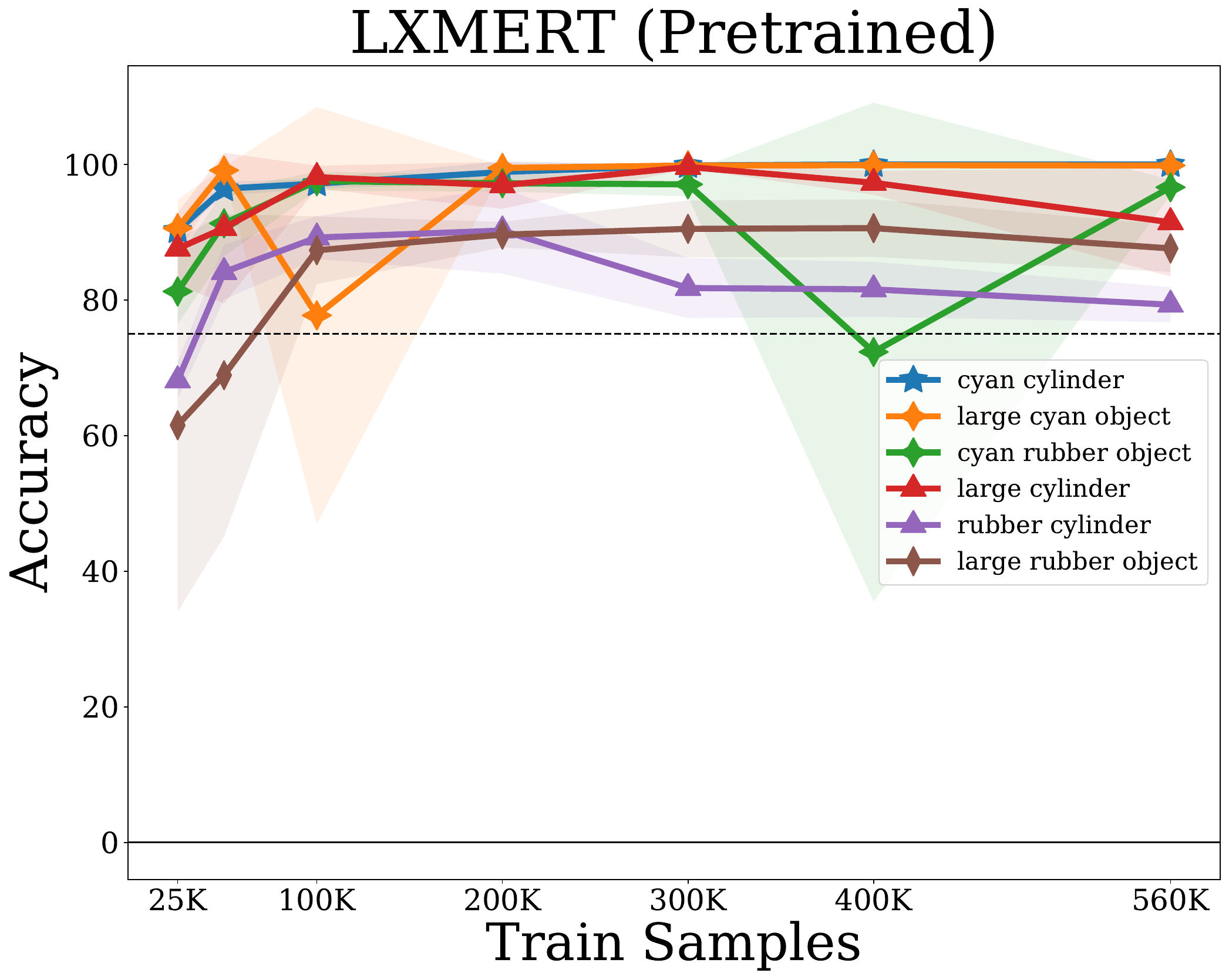}\\
    \includegraphics[width=0.5\textwidth]{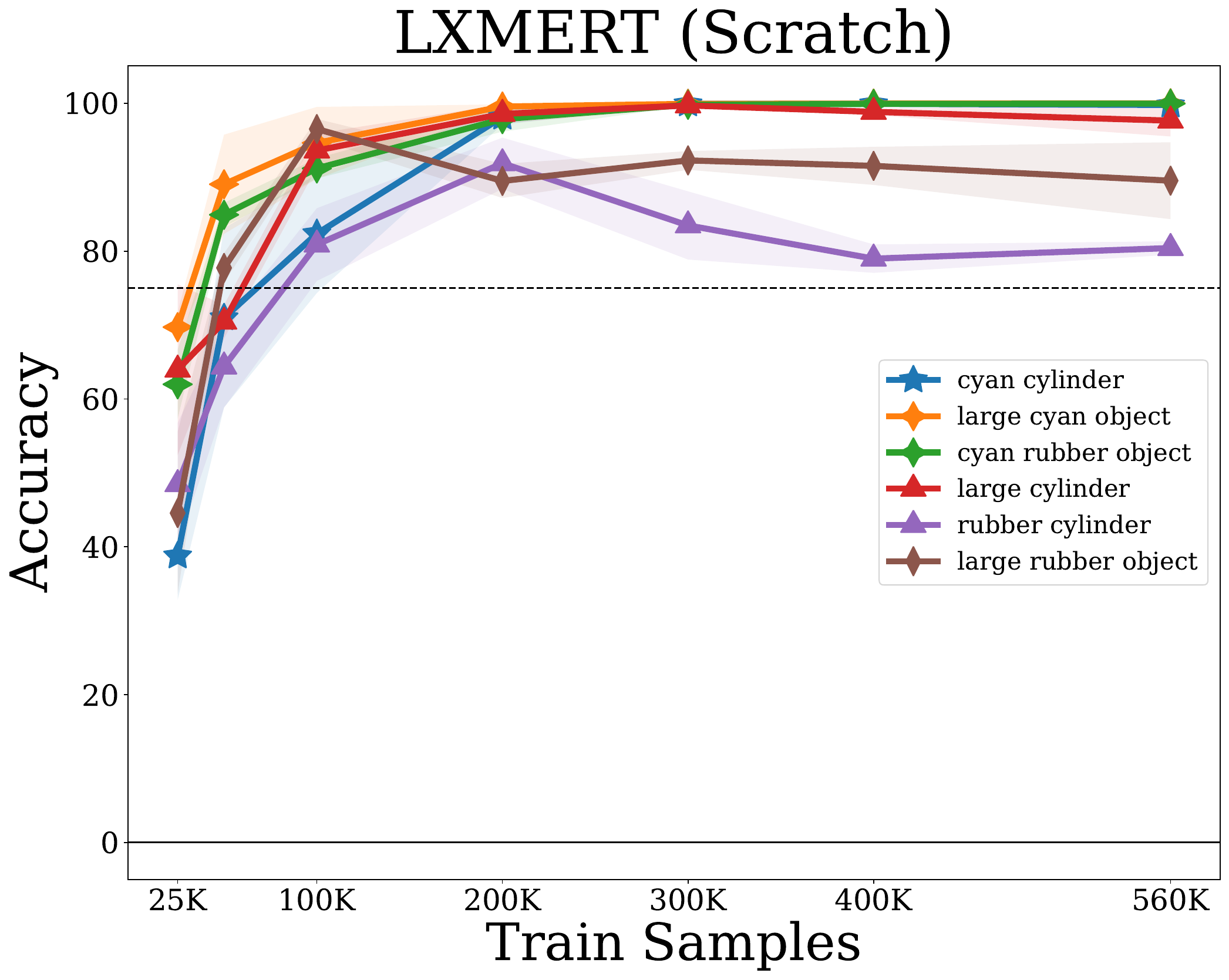}\\
    \includegraphics[width=0.5\textwidth]{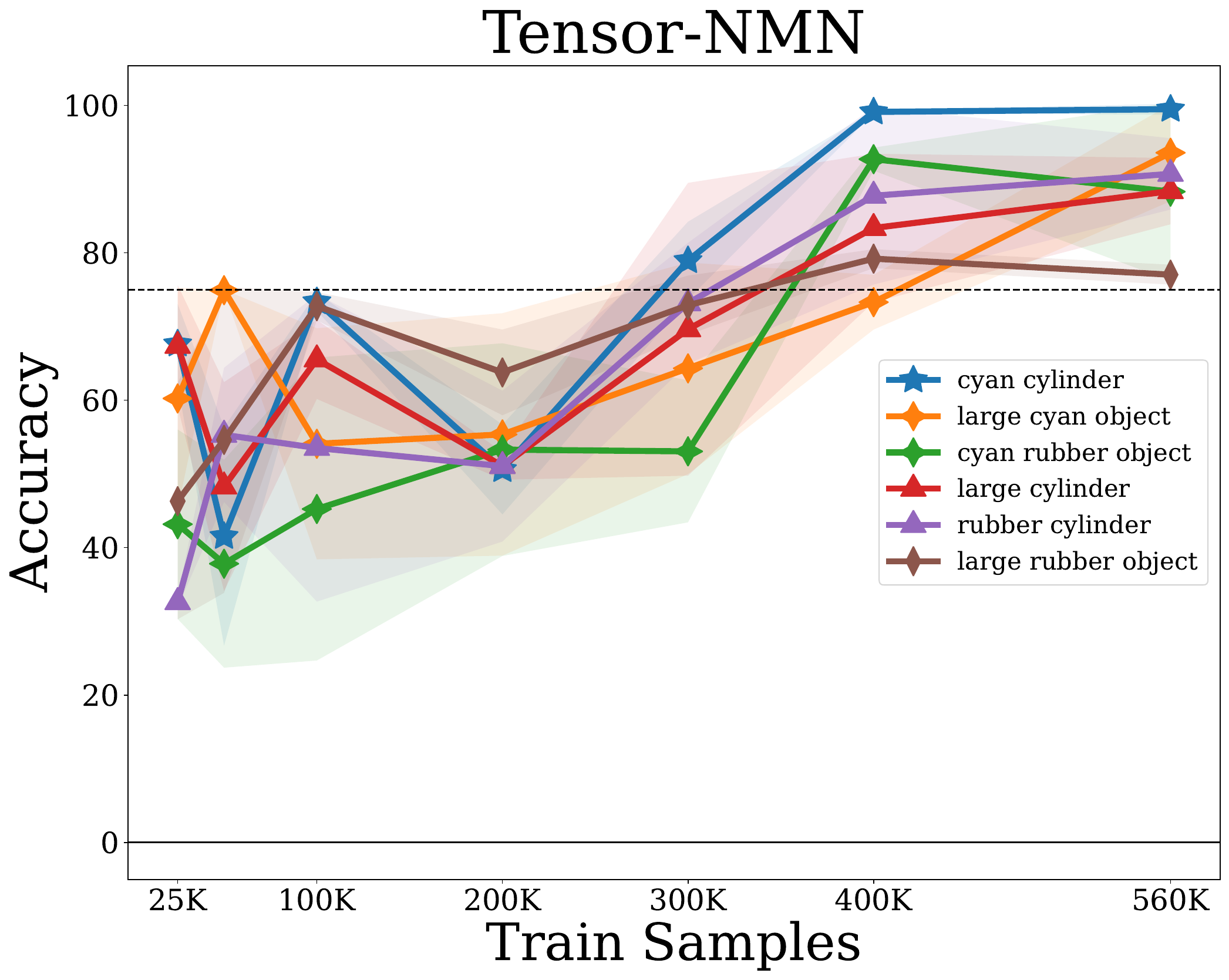}\\
    \caption{Average \textbf{\atomHo} set \nmn performance for the first 6 {\ho}s over 3 trials. For comparison, we also plot the average LXMERT model performances (i.e.,~\Cref{fig:atom_ho}), but restricted to only the first 6 {\ho}s. An area corresponding to 1 standard deviation is shaded.}
    \label{fig:ho05_atom_ho}
\end{figure}

\begin{figure}[h]
    \centering
    \includegraphics[width=0.5\textwidth]{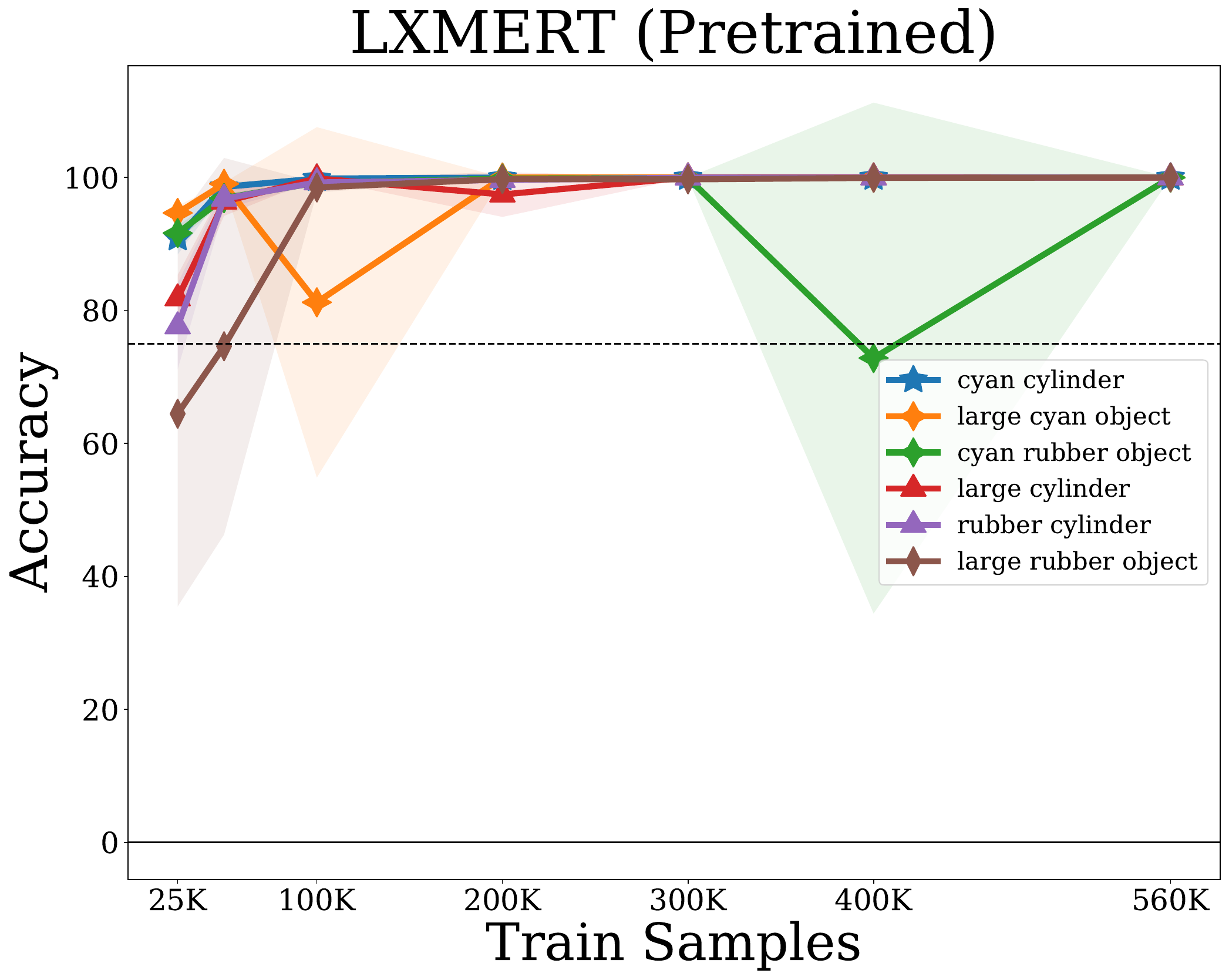}\\
    \includegraphics[width=0.5\textwidth]{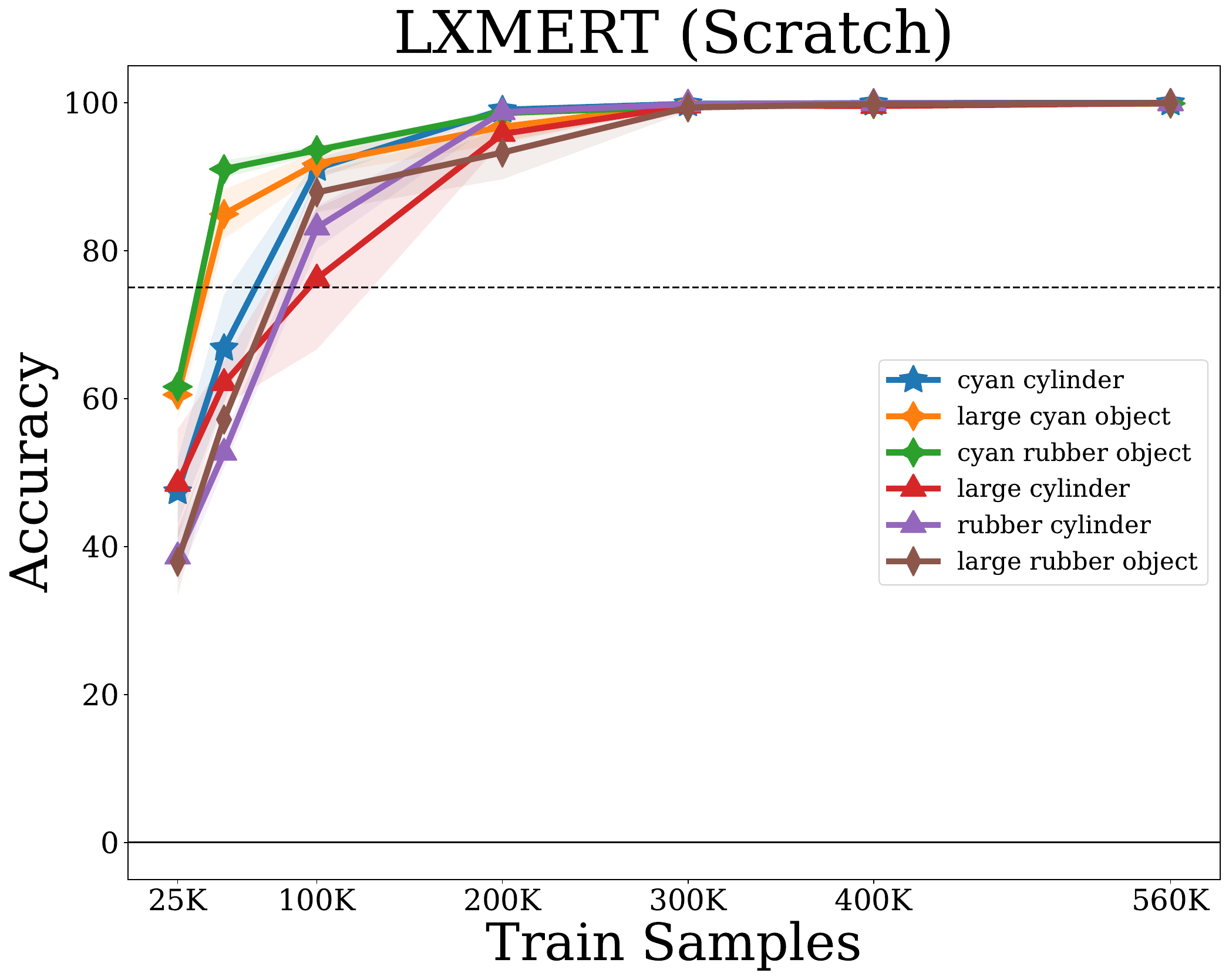}\\
    \includegraphics[width=0.5\textwidth]{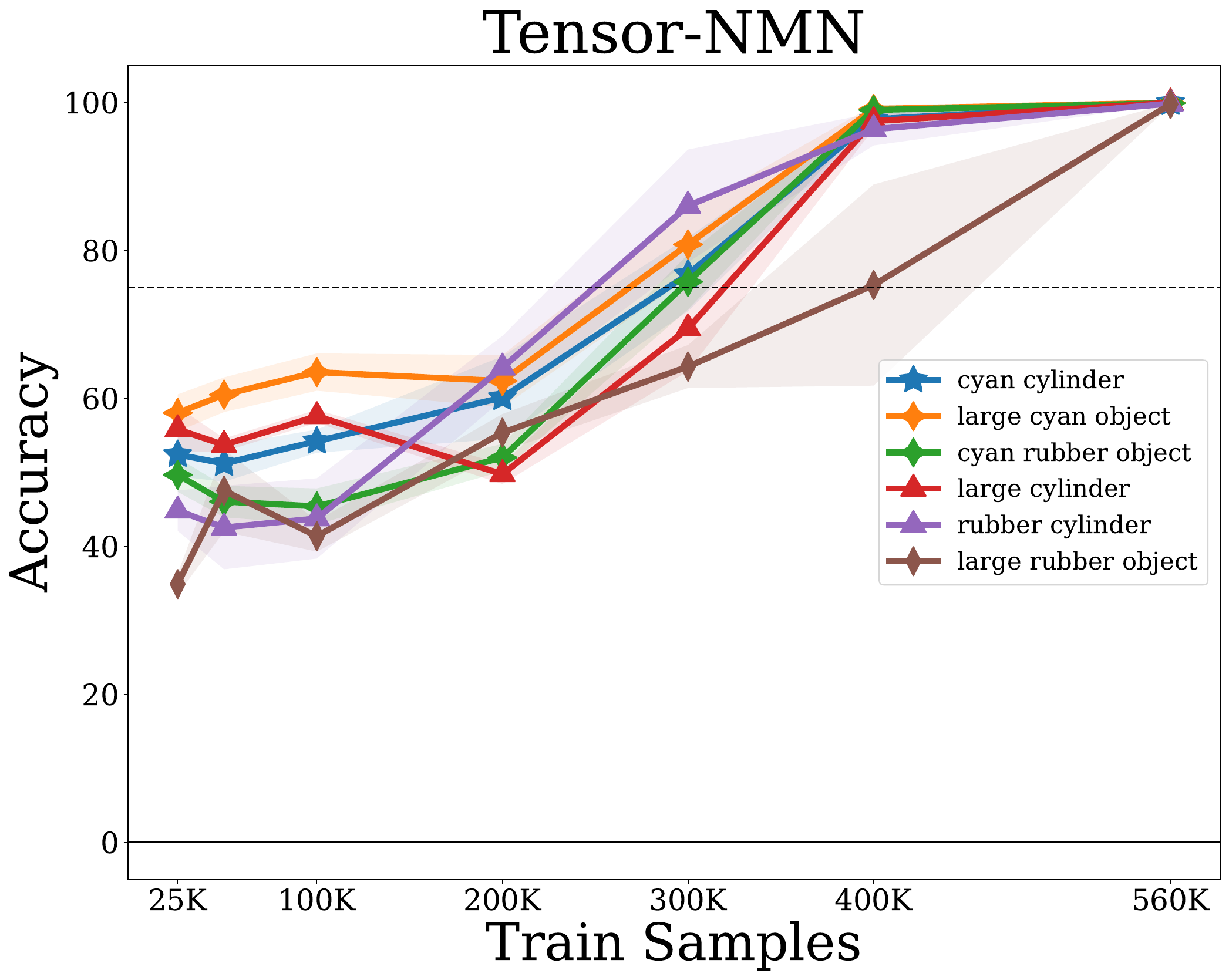}\\
    \caption{Average \textbf{\atomNonHo} set \nmn performance for the first 6 {\ho}s over 3 trials. For comparison, we also plot the average LXMERT model performances (i.e.,~\Cref{fig:atom_non_ho}), but restricted to only the first 6 {\ho}s. An area corresponding to 1 standard deviation is shaded.}
    \label{fig:ho05_atom_non_ho}
\end{figure}

Model performance on \atomHo can be found in~\Cref{fig:ho05_atom_ho}. Performance on \atomNonHo can be found in~\Cref{fig:ho05_atom_non_ho}. All plots mark 75\%---this baseline performance is achieved on the \atomHo split by always predicting false (i.e., the most common class). Always predicting false on \atomNonHo yield a baseline performance between 66\% and 75\%, depending on the \ho. 

\begin{figure}[h]
    \centering
    \includegraphics[width=0.5\textwidth]{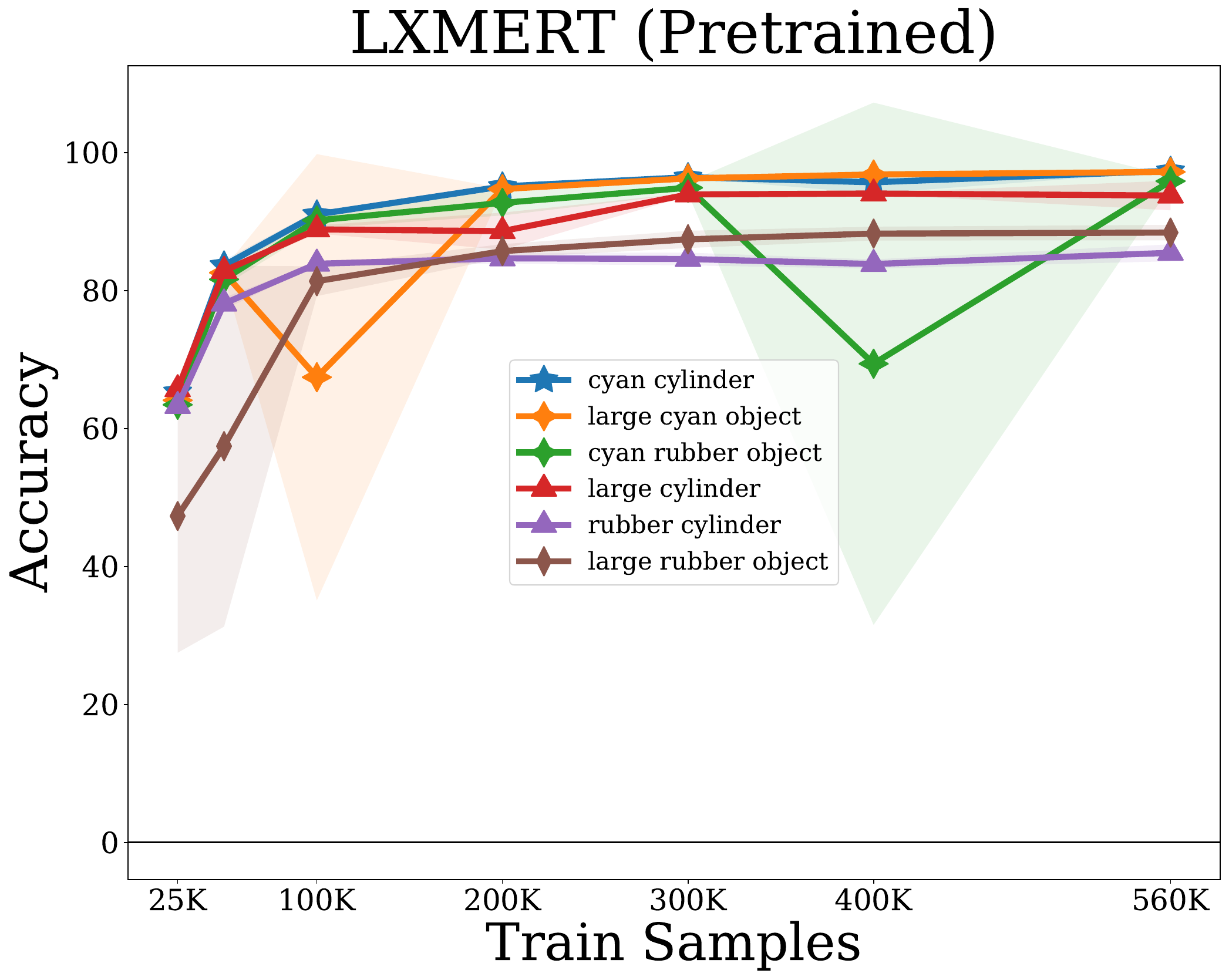}\\
    \includegraphics[width=0.5\textwidth]{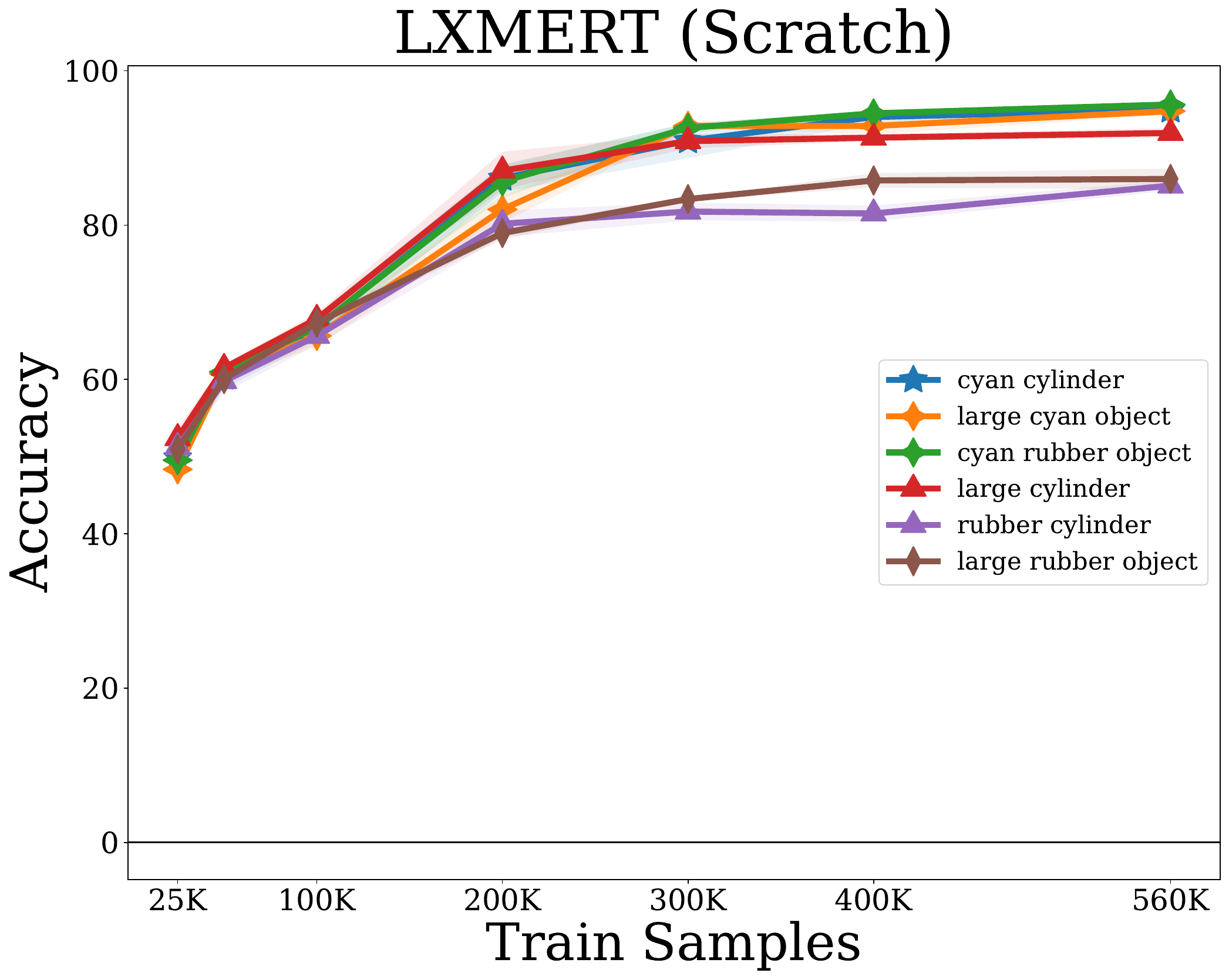}\\
    \includegraphics[width=0.5\textwidth]{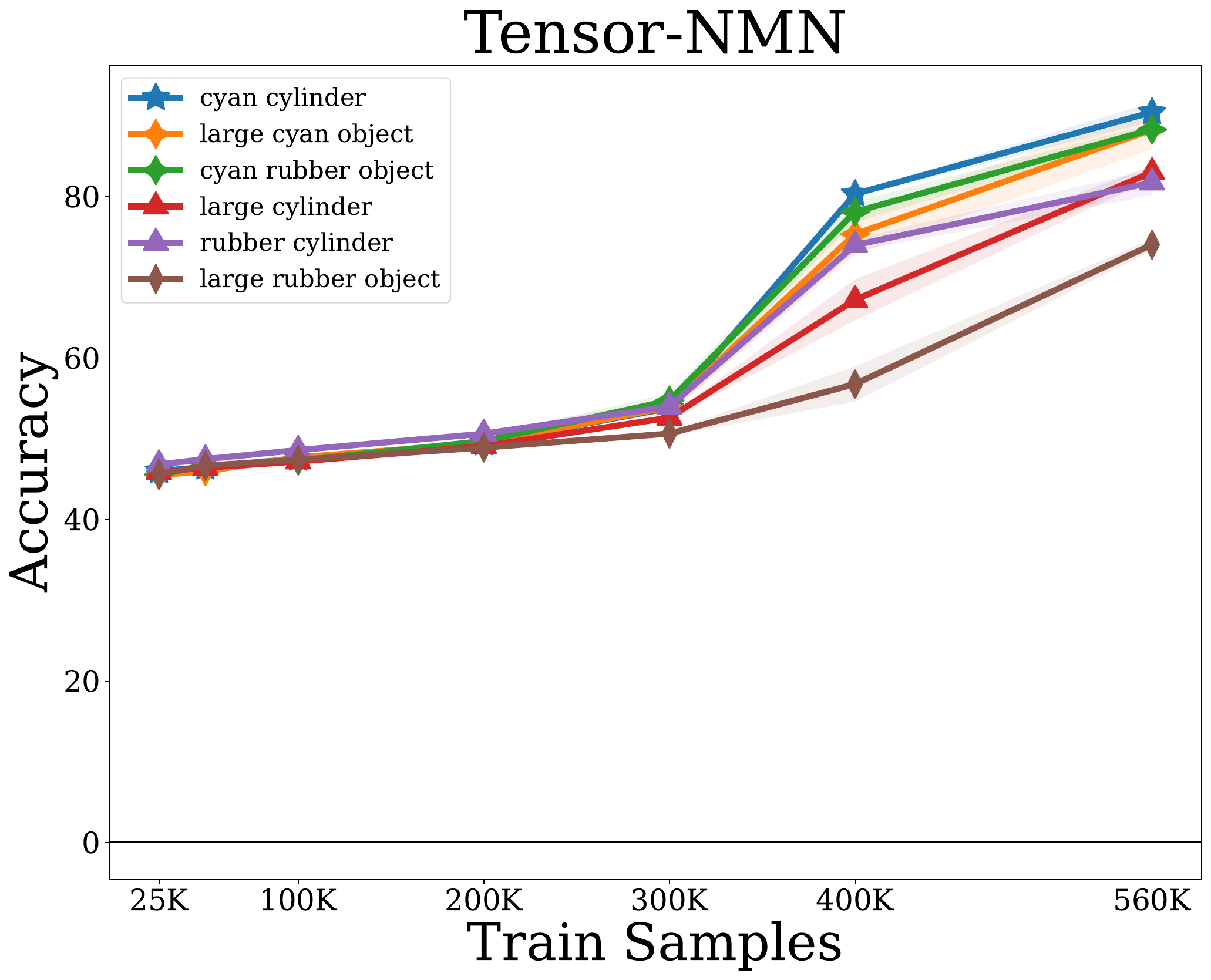}\\
    \caption{Average \textbf{\hoTest} set \nmn performance for the first 6 {\ho}s over 3 trials. For comparison, we also plot the average LXMERT model performances (i.e.,~\Cref{fig:ho_test}), but restricted to only the first 6 {\ho}s. An area corresponding to 1 standard deviation is shaded.}
    \label{fig:ho05_ho_test}
\end{figure}

\begin{figure}[h]
    \centering
    \includegraphics[width=0.5\textwidth]{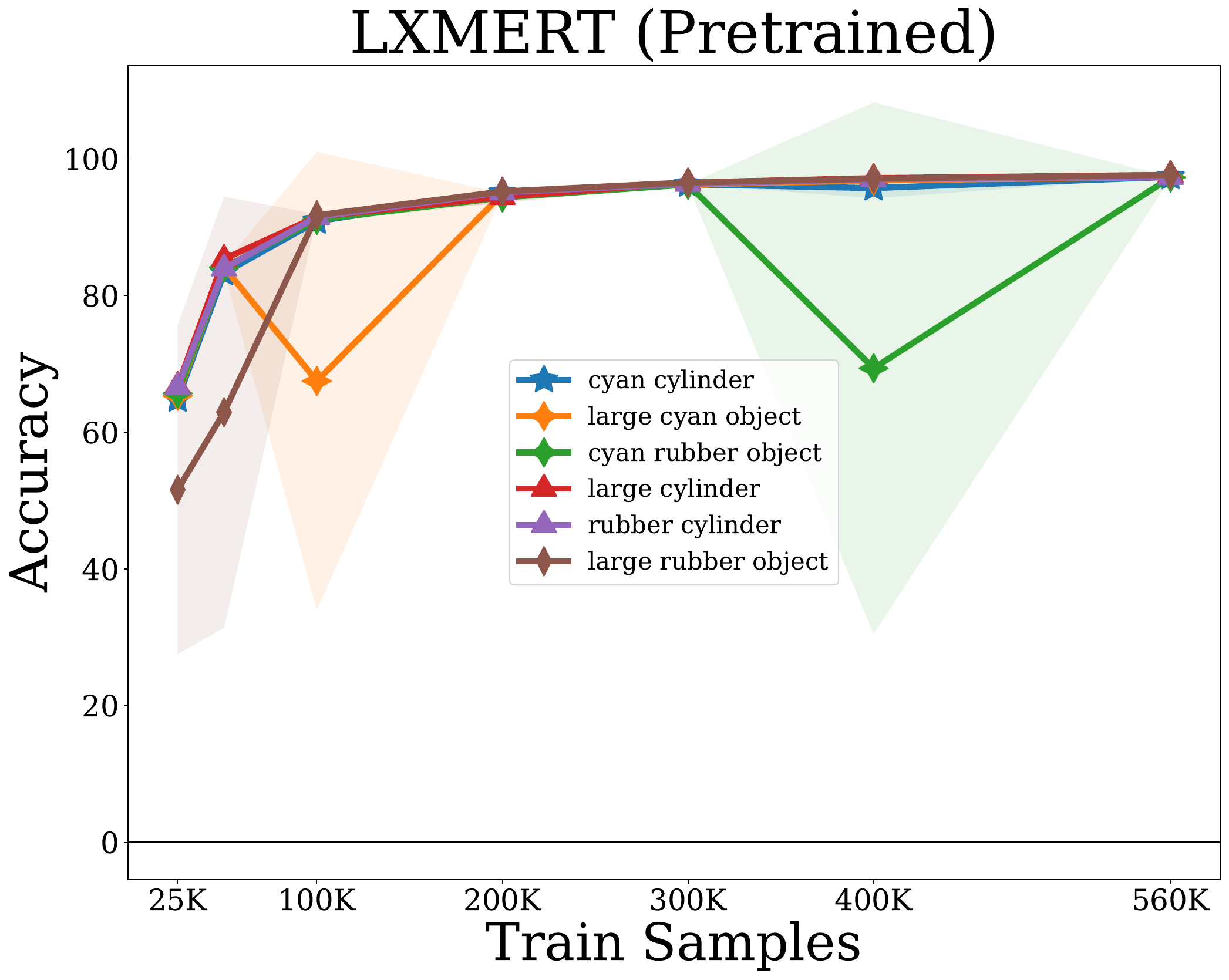}\\
    \includegraphics[width=0.5\textwidth]{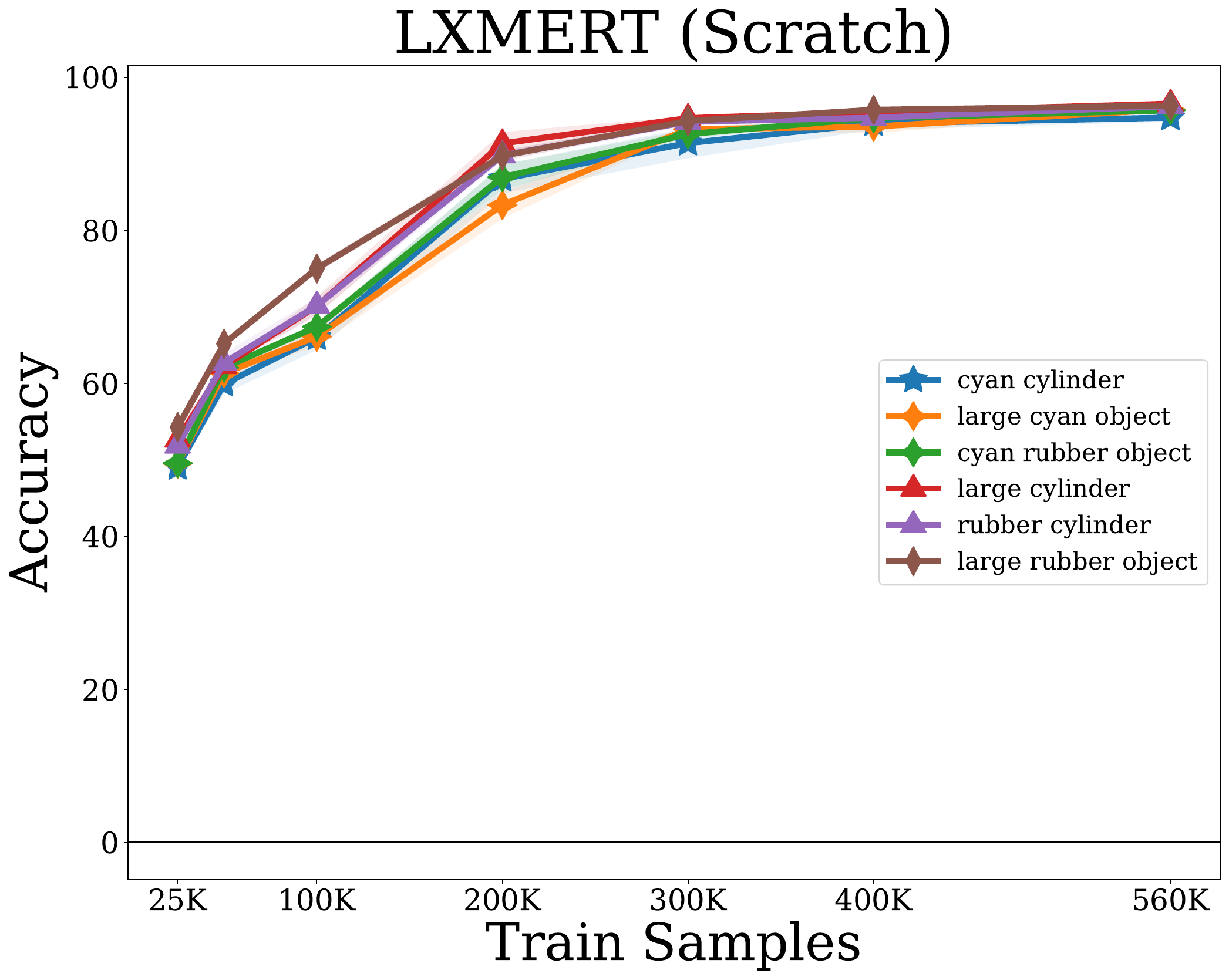}\\
    \includegraphics[width=0.5\textwidth]{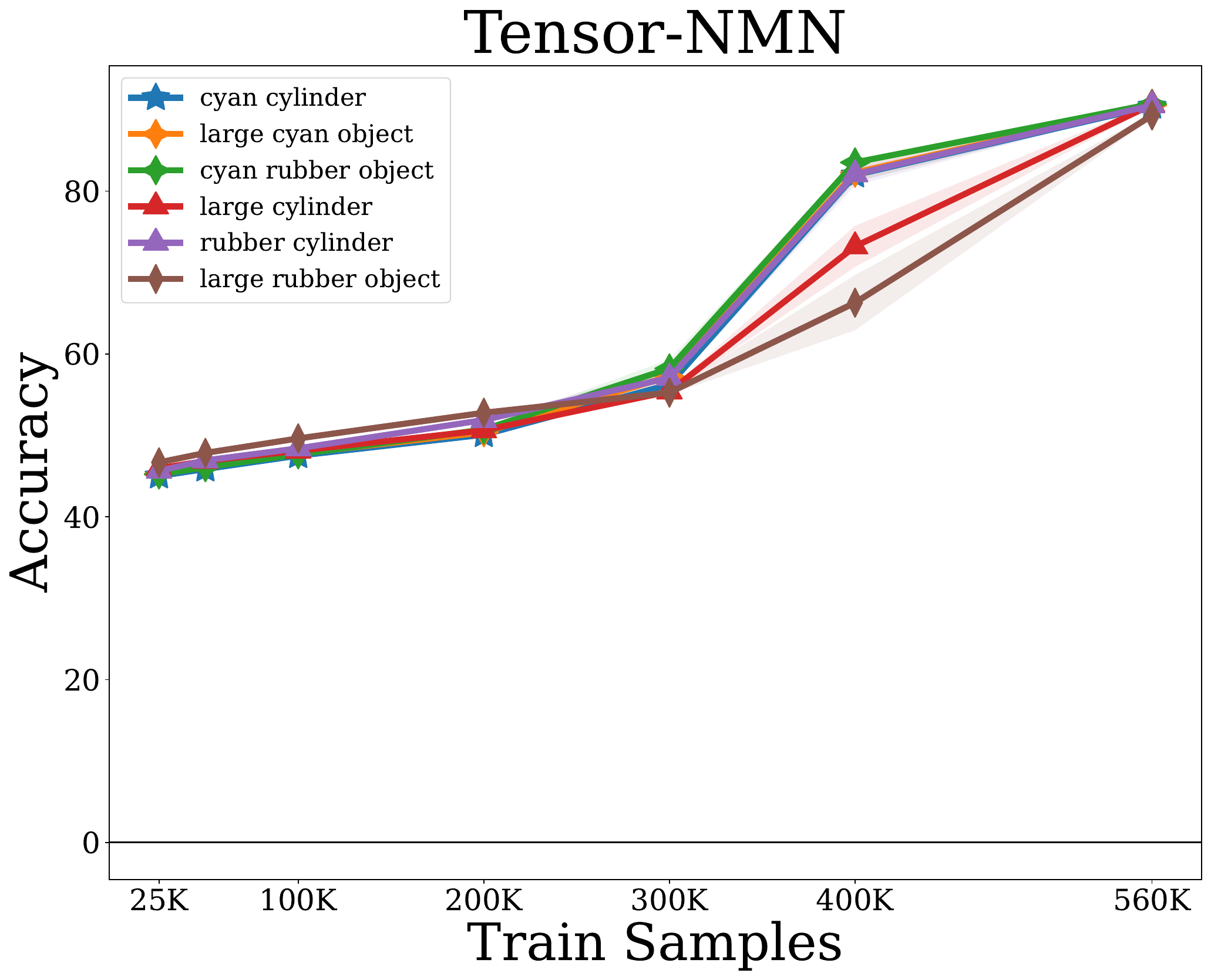}\\
    \caption{Average \textbf{\hoVal} set \nmn performance for the first 6 {\ho}s over 3 trials. For comparison, we also plot the average LXMERT model performances (i.e.,~\Cref{fig:ho_val}), but restricted to only the first 6 {\ho}s. An area corresponding to 1 standard deviation is shaded.}
    \label{fig:ho05_ho_val}
\end{figure}

Model performance on \hoTest can be found in~\Cref{fig:ho05_ho_test}. Performance on \hoVal can be found in~\Cref{fig:ho05_ho_val}. 

\begin{figure*}[h]
    \centering
    \includegraphics[width=\textwidth]{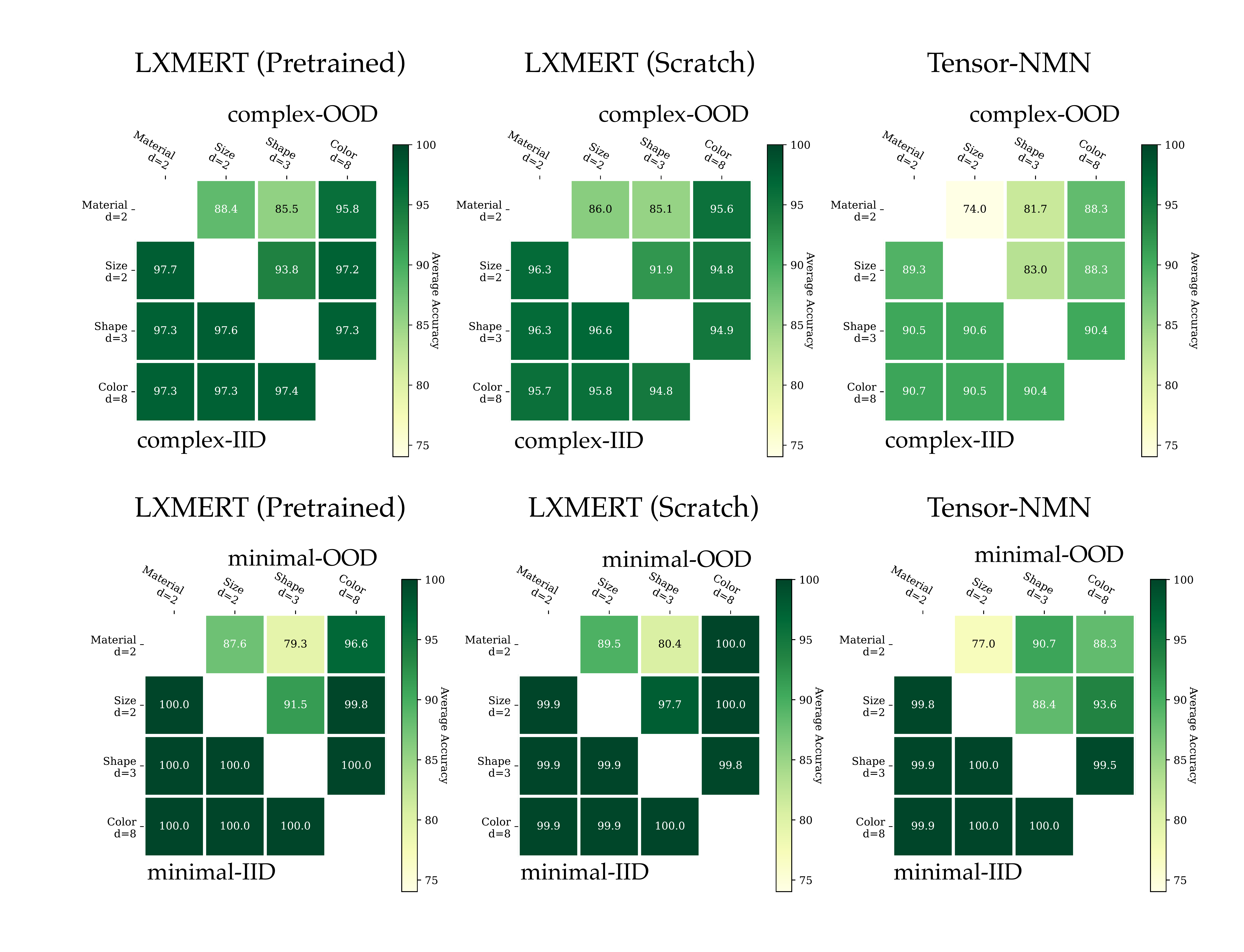}
    \caption{Model accuracies for \textit{only} the first 6 {\ho}s. Note that while the LXMERT models struggle with \attType{Material}-\attType{Shape} combinations on OOD questions, \nmn does not.}
    \label{fig:matrix_acc_ho0_5}
\end{figure*}

For \nmn trained on the largest \train sets (560k), we plot the complex and minimal model accuracies, averaged by the attribute types of the {\ho}s. The results are visualized in~\Cref{fig:matrix_acc_ho0_5}. Again, we include the corresponding subset of LXMERT models for comparison.

The exact average accuracies and standard deviations over 3 runs are in Tables~\ref{tab:summary_nmn_acc_aggregate_val} through \ref{tab:summary_nmn_acc_aggregate_atom-ho}.

\section{Systematicity Gap}\label{sec:Appx._sysGap}

\begin{figure}[h]
    \centering
    \includegraphics[width=0.5\textwidth]{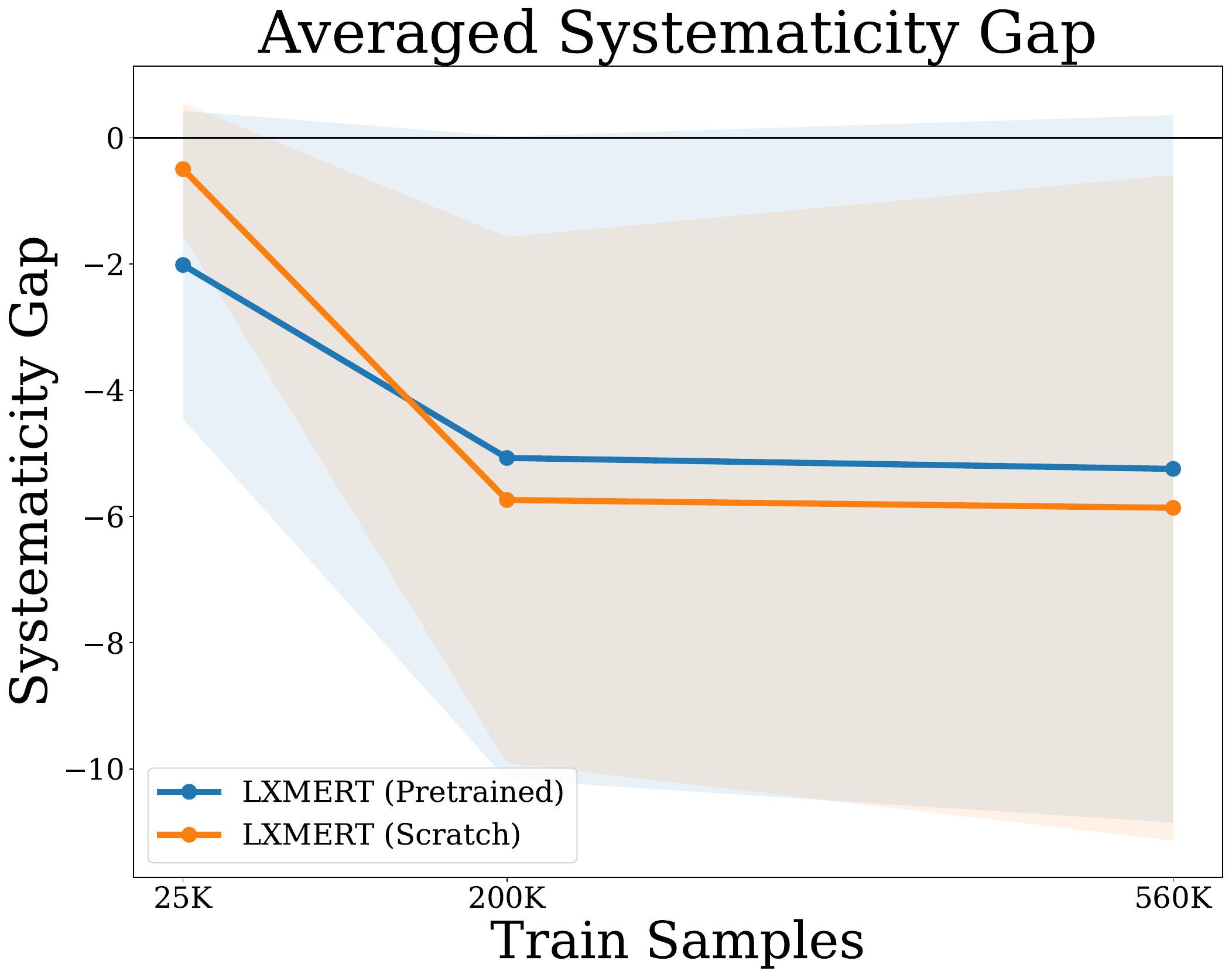}
    \caption{Average systematicity gap on \textbf{complex} examples (i.e., \hoTest accuracy minus \hoVal accuracy) with 1 standard deviation; averaged over 3 runs on each of the 29 {\ho}s. Note that the systematicity gap plateaus, suggesting that the performance drop when generalizing to unseen combinations does not improve with additional training data. }
    \label{fig:sys_gap_complex}
\end{figure}

\begin{figure}[h]
    \centering
    \includegraphics[width=0.5\textwidth]{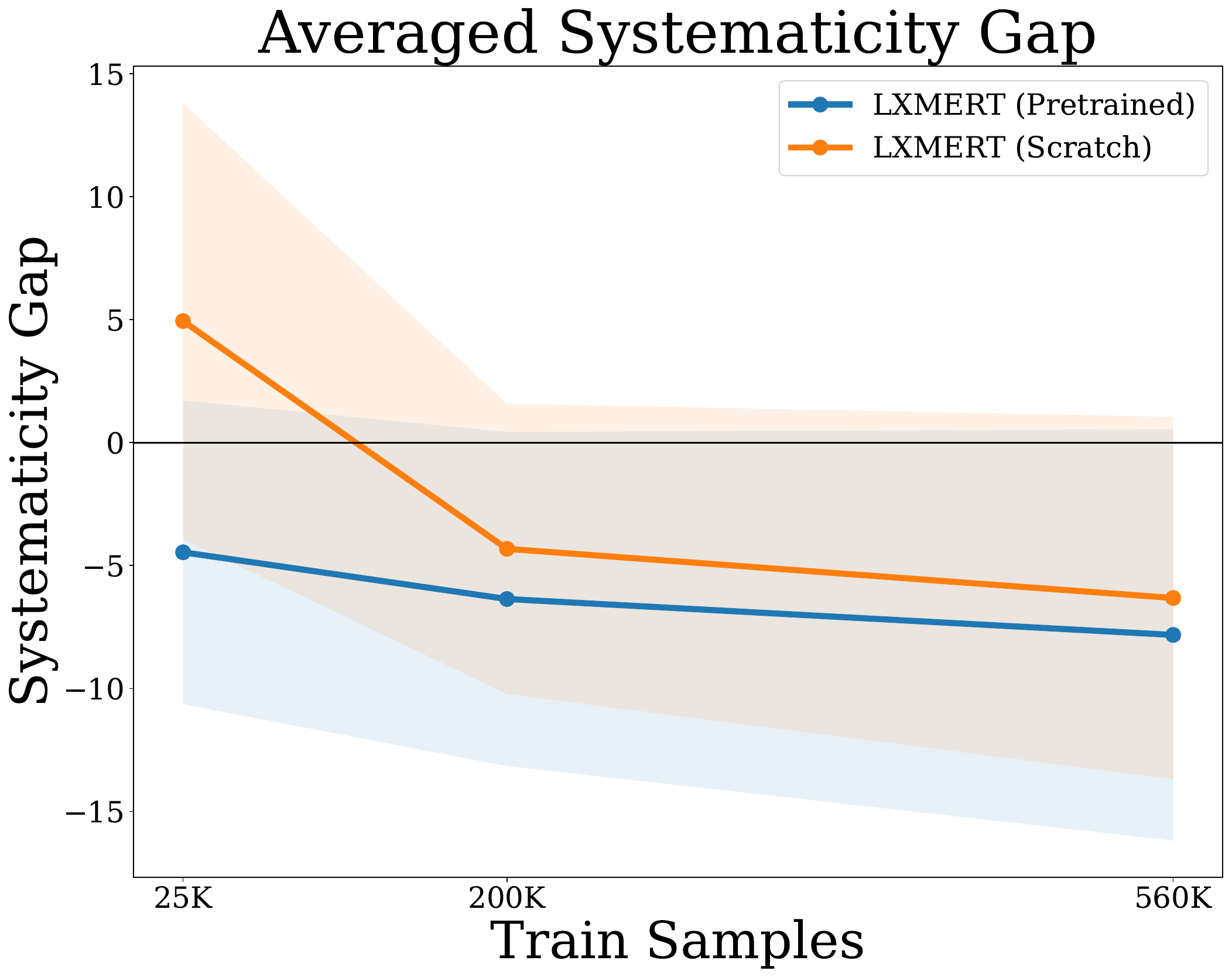}
    \caption{Average systematicity gap on \textbf{minimal} examples (i.e., \atomHo accuracy minus \atomNonHo accuracy) with 1 standard deviation; averaged over 3 runs on each of the 29 {\ho}s.  
    }
    \label{fig:sys_gap_minimal}
\end{figure}

\begin{figure}[h]
    \centering
    \includegraphics[width=0.5\textwidth]{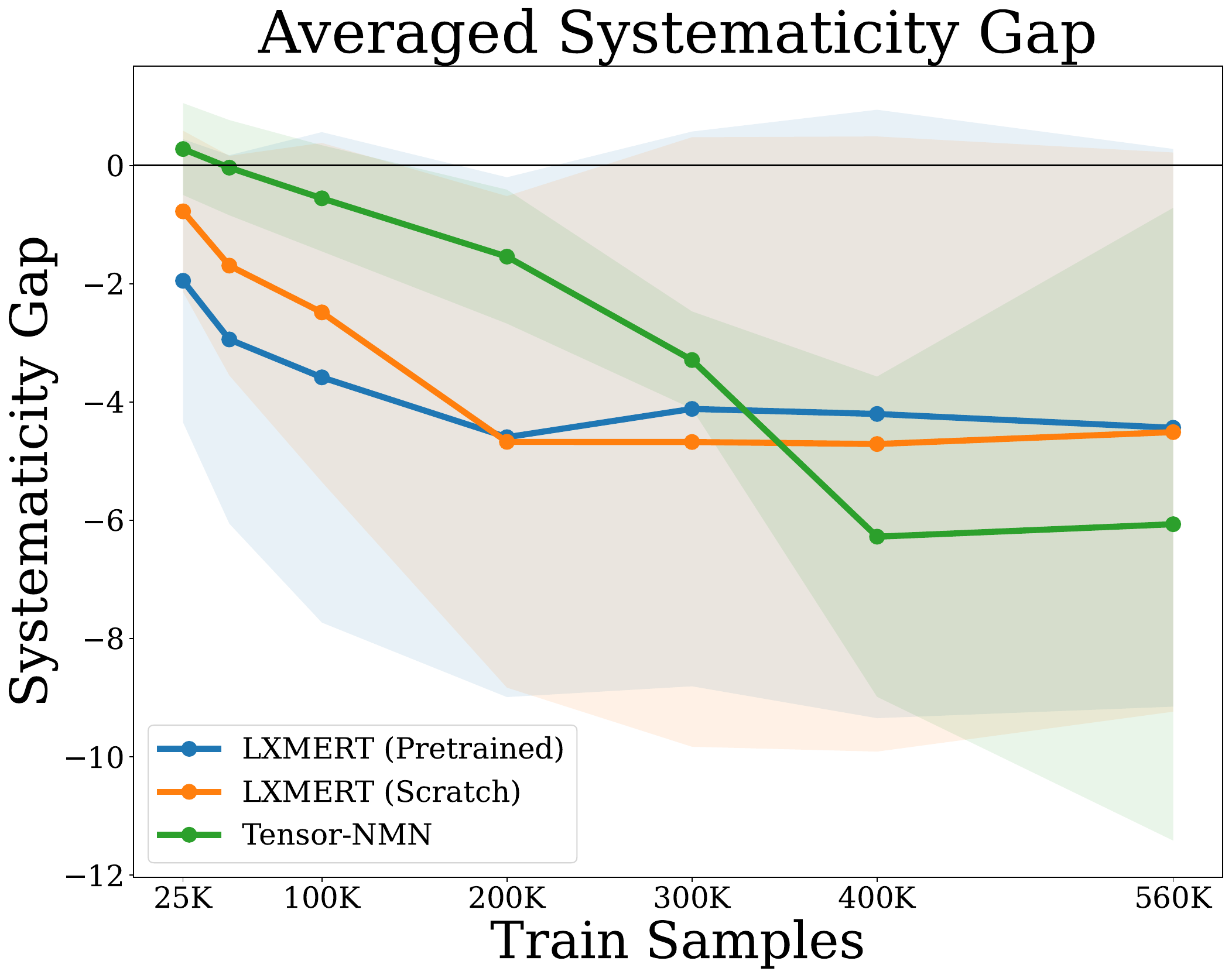}
    \caption{Average systematicity gap on \textbf{complex} examples (i.e., \hoTest accuracy minus \hoVal accuracy) with 1 standard deviation; averaged over 3 runs on \textit{only} the first 6 {\ho}s. Note that the systematicity gap plateaus, suggesting that the performance drop when generalizing to unseen combinations does not improve with additional training data.}
    \label{fig:ho05_sys_gap_complex}
\end{figure}

\begin{figure}[h]
    \centering
    \includegraphics[width=0.5\textwidth]{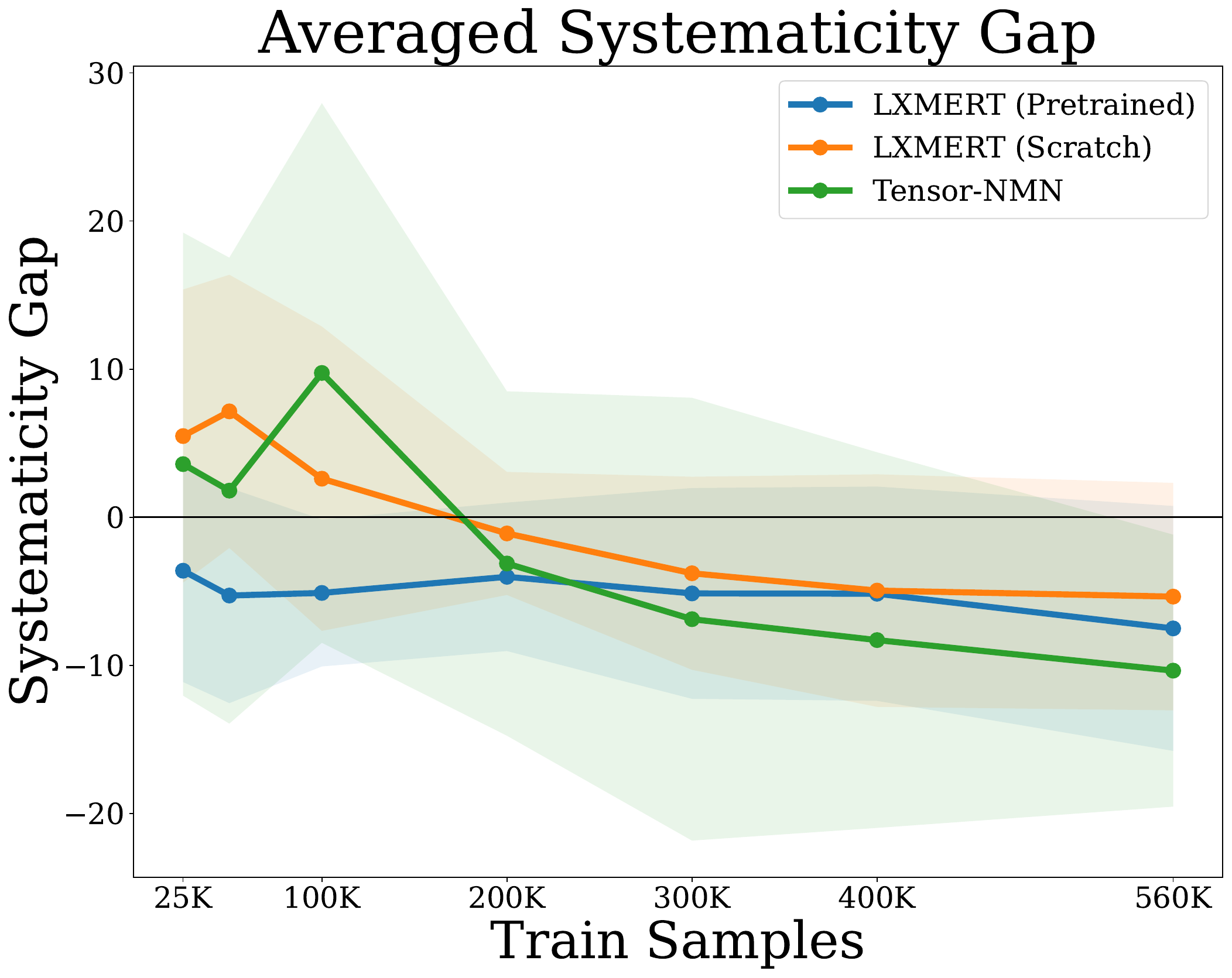}
    \caption{Average systematicity gap on \textbf{minimal} examples (i.e., \atomHo accuracy minus \atomNonHo accuracy) with 1 standard deviation; averaged over 3 runs on \textit{only} the first 6 {\ho}s.
    }
    \label{fig:ho05_sys_gap_minimal}
\end{figure}

As outlined in Section \ref{sec:sysGap}, we find that, on all models, averaged over {\ho}s, the gap between performance on complex questions involving IID vs. OOD attribute combinations does not trend to zero. Instead, it plateaus (see Figures \ref{fig:sys_gap_complex} and \ref{fig:ho05_sys_gap_complex}). In comparison, the performance gap on minimal questions plateaus or decreases gently (see Figures \ref{fig:sys_gap_minimal} and \ref{fig:ho05_sys_gap_minimal}).

\begin{figure*}[ht!]
  \centering
  \includegraphics[width=\textwidth]{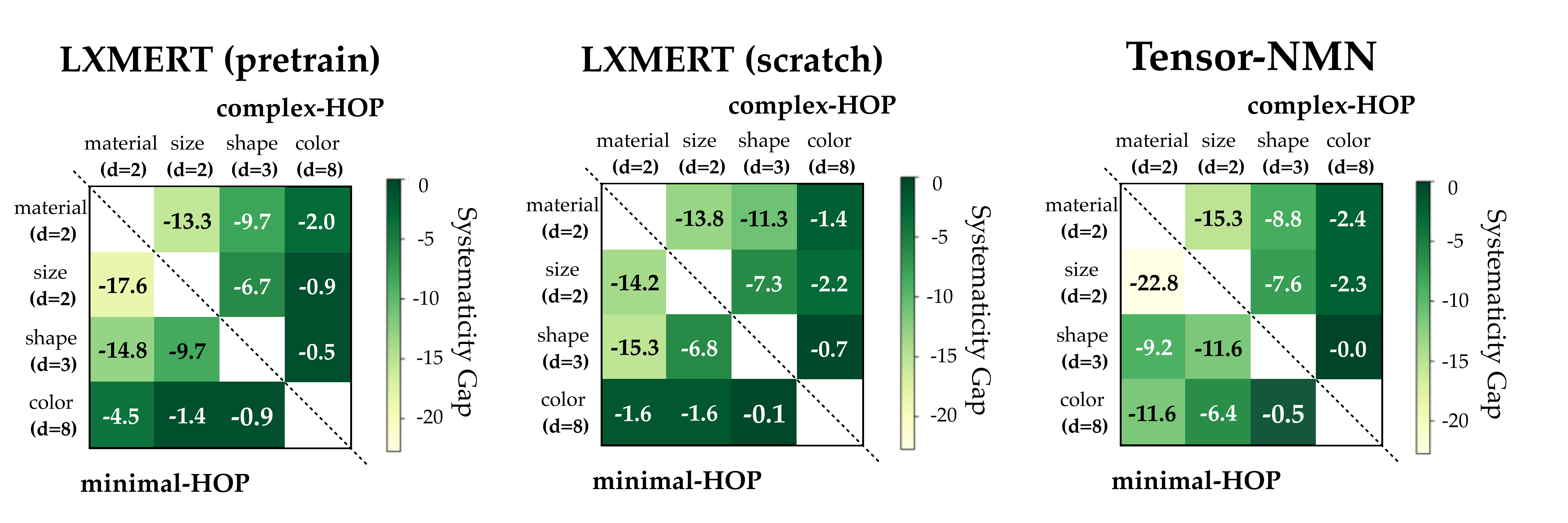}
  \caption{Systematicity gap on the complex splits (top corner) and minimal splits (bottom corner) for all models trained on 560k training examples. The systematicity gap is averaged according to the attribute types of the {\ho}s, all 29 {\ho}s for LXMERT, {\ho}s 0-5 for Tensor-NMN---attributes are sorted by increasing diversity on the axes (e.g., \attType{shape} has 2 possible values, \attType{color} has 8 possible values). As expected, we see a worse systematicity gap (i.e. lighter colors) in the top left (low-diversity combinations), and better systematicity gap (i.e., darker colors) in the bottom right (high-diversity combinations).}
    \label{fig:delta_matrix}
\end{figure*}

In~\Cref{fig:delta_matrix} we visualize the systematicity gap by attribute-types in the pair on both LXMERT and Tensor-NMN. It can be seen that the systematicity gaps are still sorted by the diversity of the attribute pairs (i.e., we see lighter colours in the top left, and darker colours in the bottom right).

\begin{figure*}[t]
    \centering

    \begin{subfigure}[b]{0.49\textwidth}
         \centering
         \includegraphics[width=0.725\textwidth]{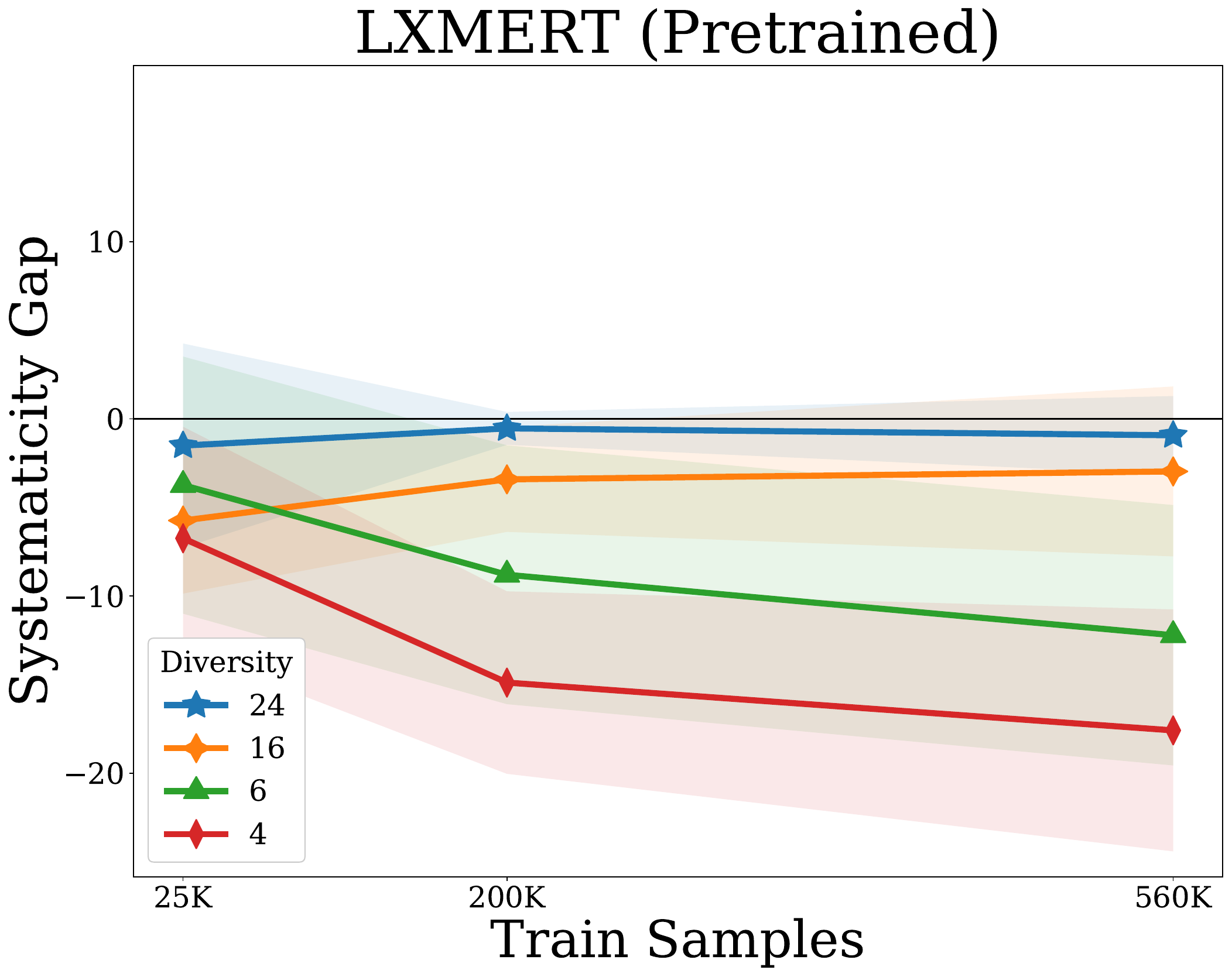}

     \end{subfigure}
     \hfill
     \begin{subfigure}[b]{0.49\textwidth}
         \centering

         \includegraphics[width=0.725\textwidth]{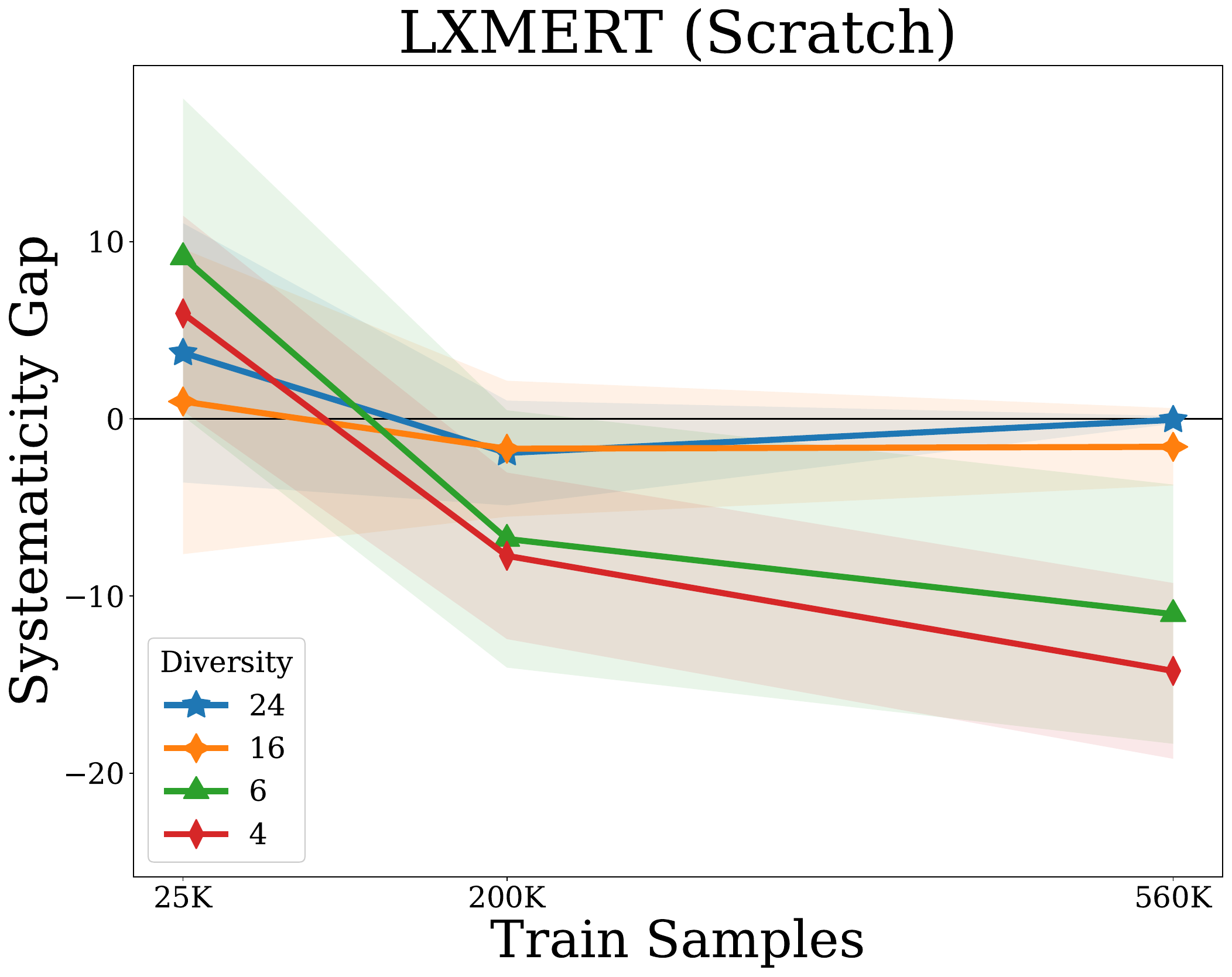}

     \end{subfigure}

    \caption{\vspace{-1mm}Systematicity gap (difference between OOD and IID model accuracy) on the {\bf minimal} split, averaged by held-out pair (\ho) diversity over 29 {\ho}s, each with 3 runs.}
    \label{fig:delta_atom_iid_ood}
\end{figure*}

The exact average systematicity gaps and standard deviations over 3 runs are in Tables~\ref{tab:sysgap_lxmert_ft_acc_aggregate_test_val} through \ref{tab:sysgap_nmn_acc_aggregate_atom-ho_filtered-atom-non-ho}.

\begin{figure}[h]
    \centering
    \includegraphics[width=0.5\textwidth]{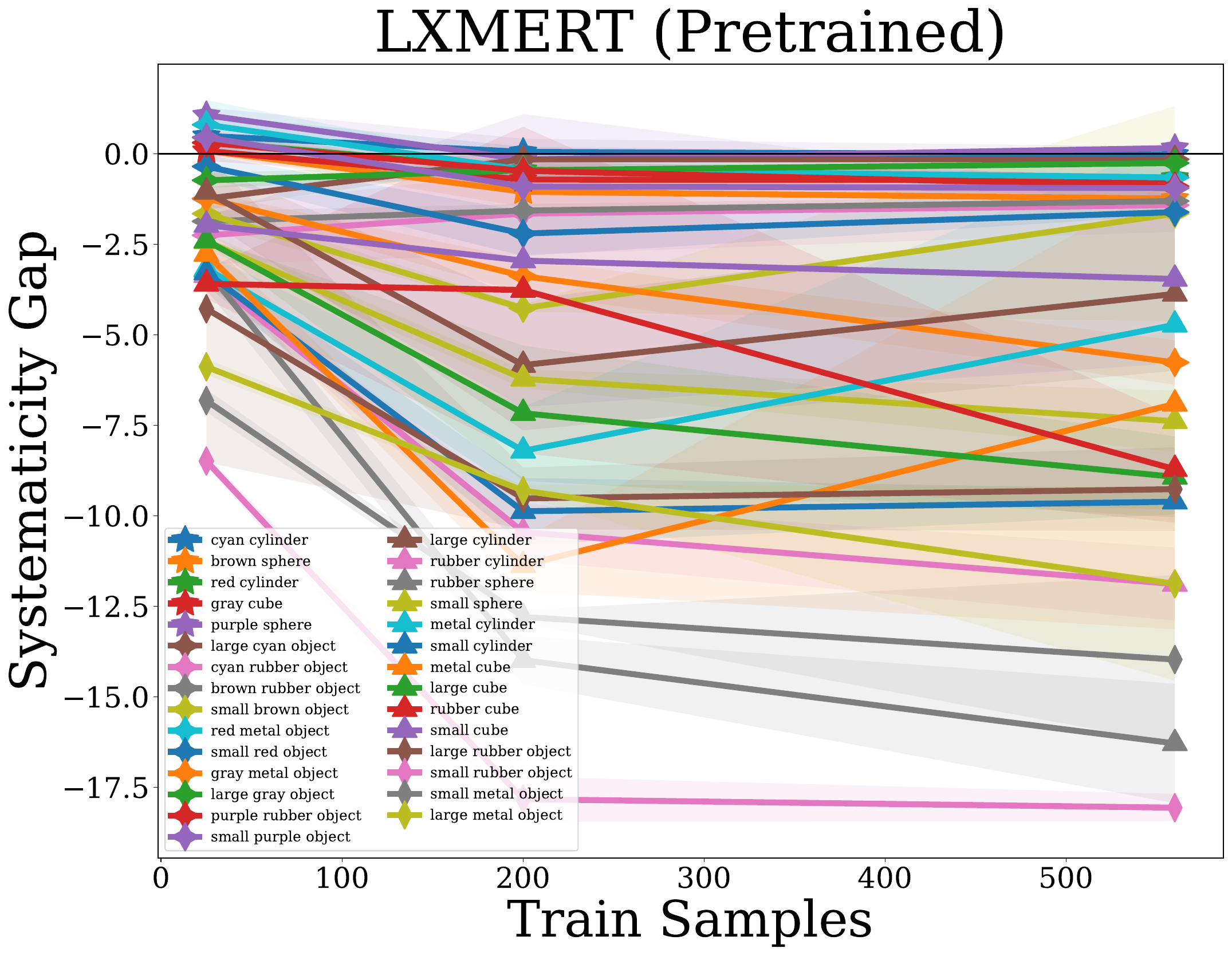}
    \includegraphics[width=0.5\textwidth]{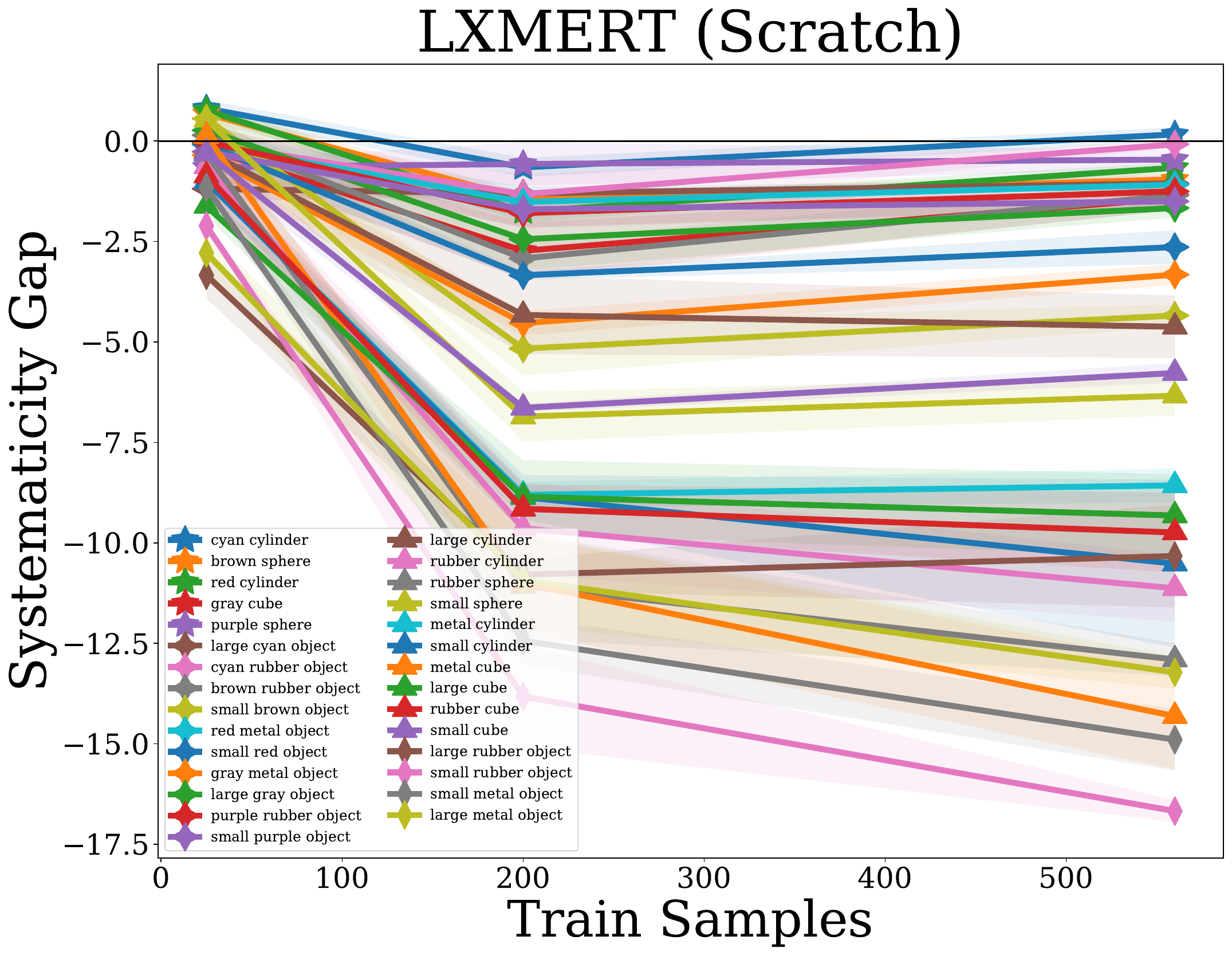}
    \includegraphics[width=0.5\textwidth]{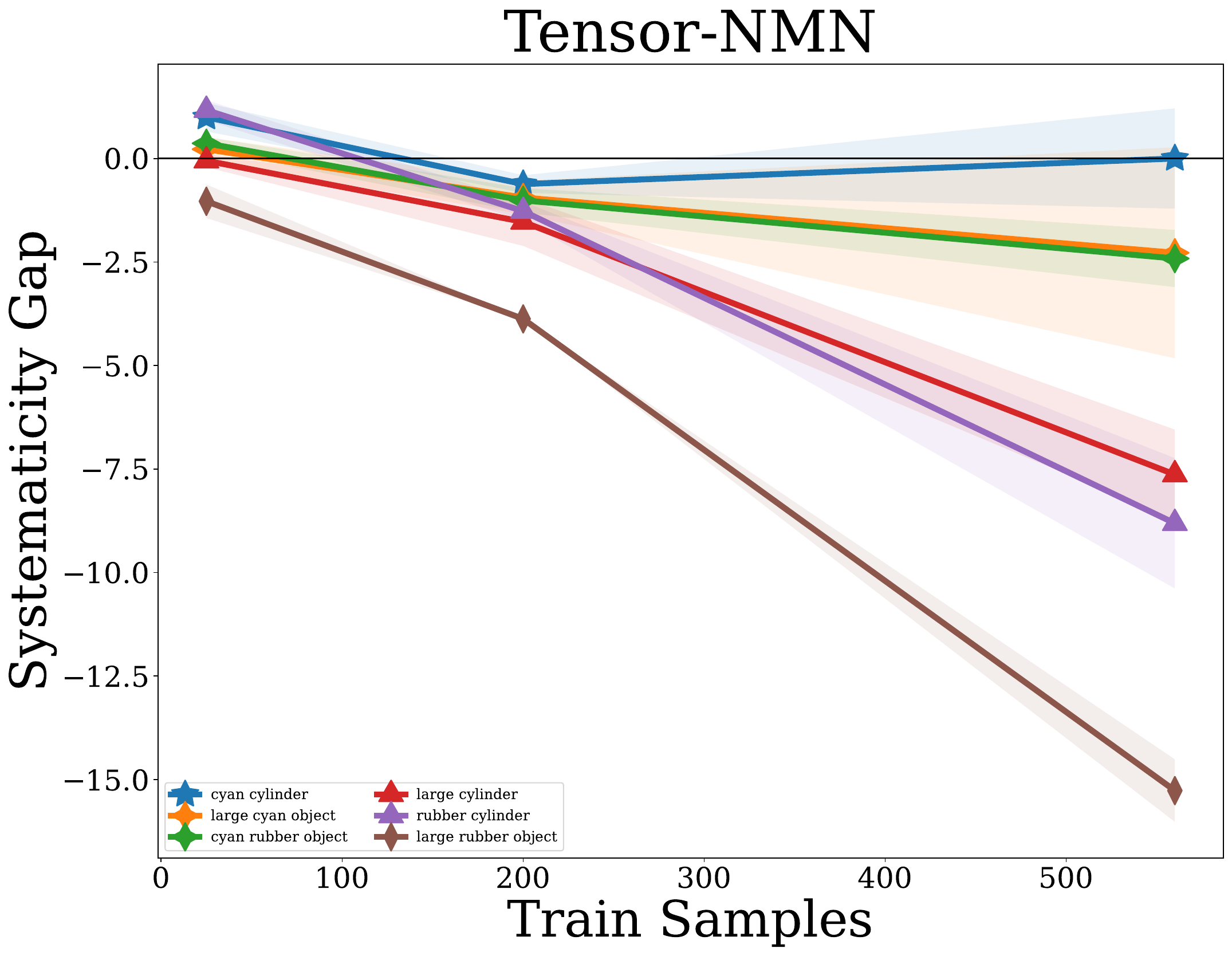}
    \caption{Systematicity gap (i.e. difference between OOD and IID model performance) for \textbf{complex} examples, averaged over 3 runs, for each \ho.}
    \label{fig:FINAL_delta_iid_ood}
\end{figure}

\begin{figure}[h]
    \centering
    \includegraphics[width=0.5\textwidth]{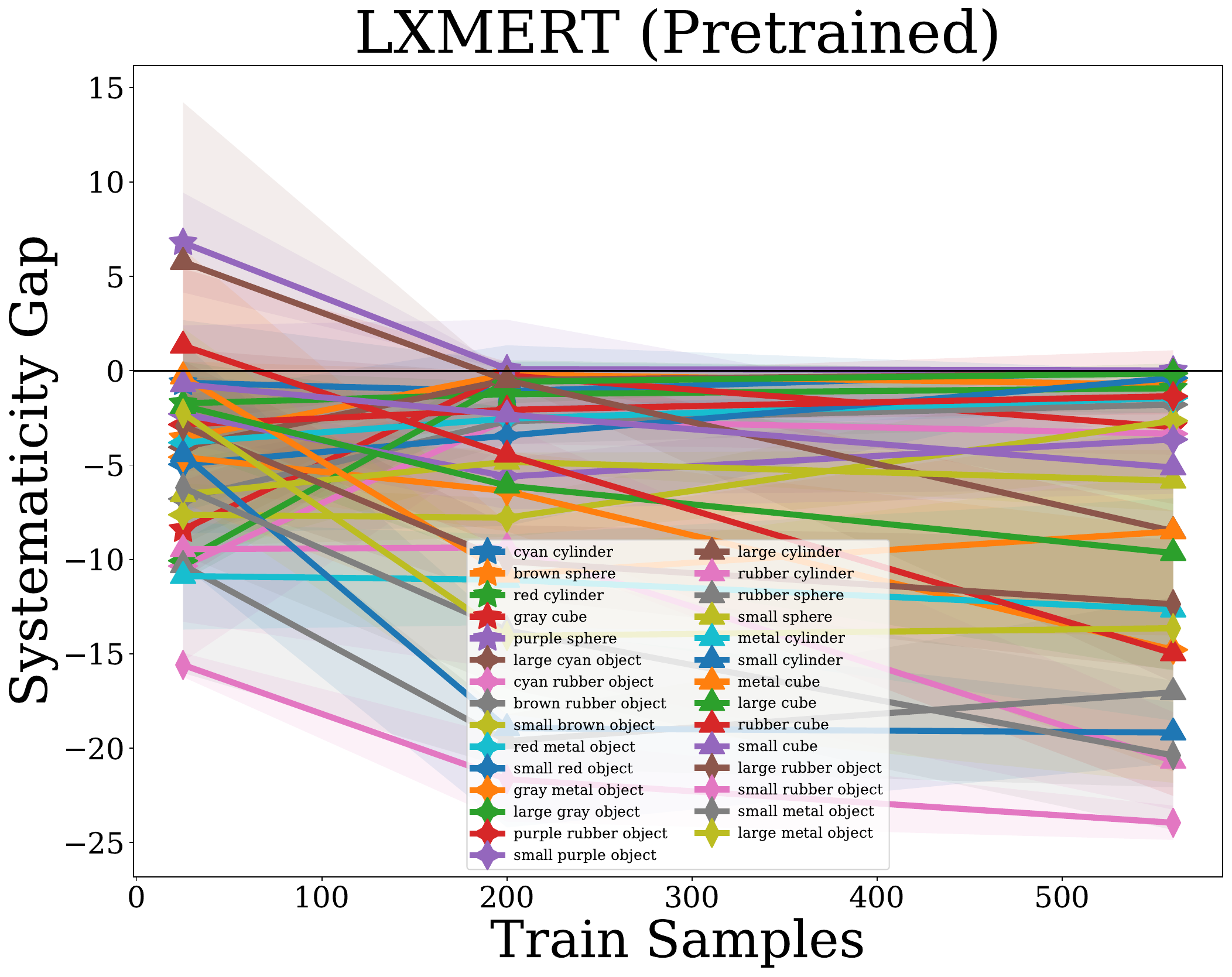}
    \includegraphics[width=0.5\textwidth]{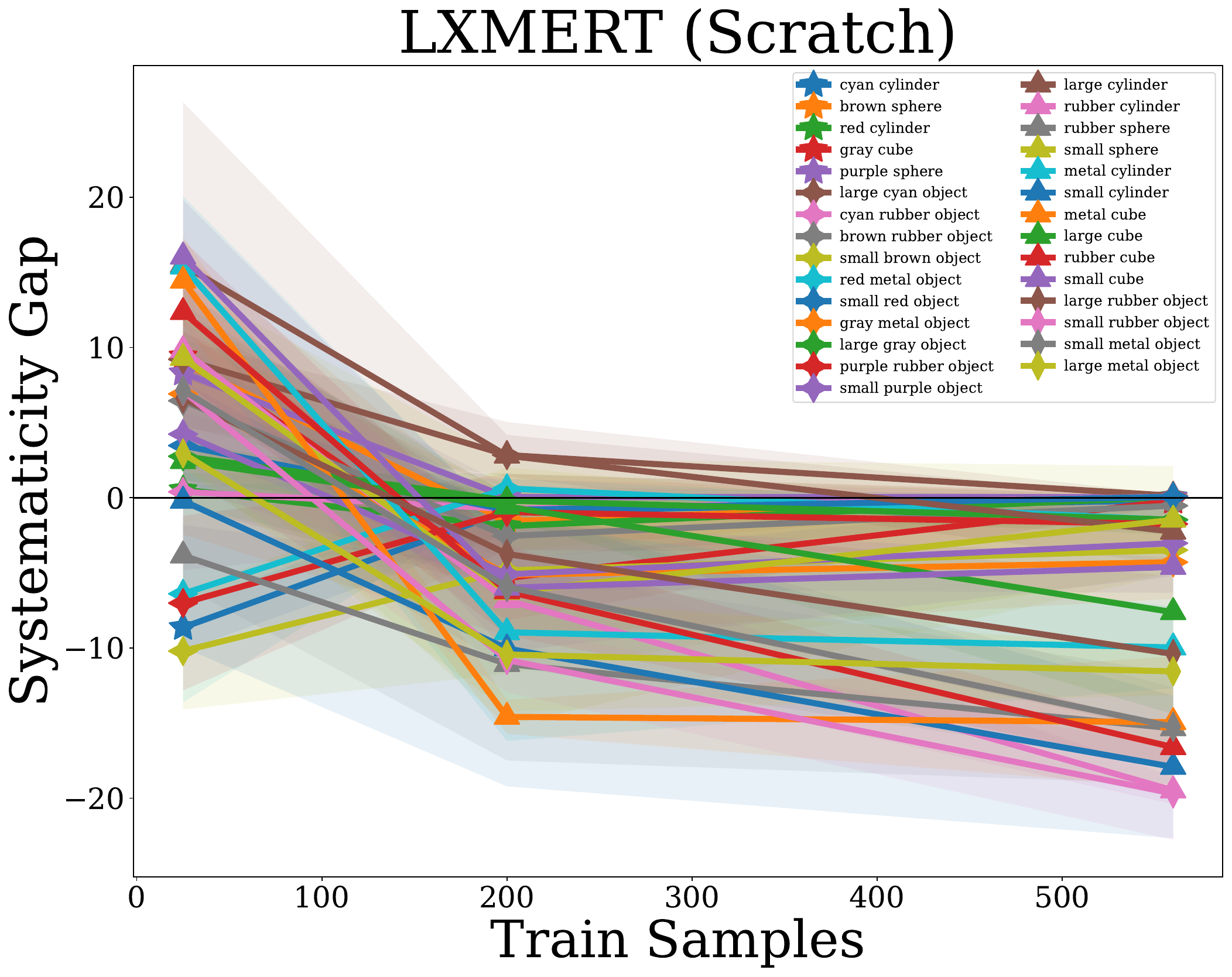}
    \includegraphics[width=0.5\textwidth]{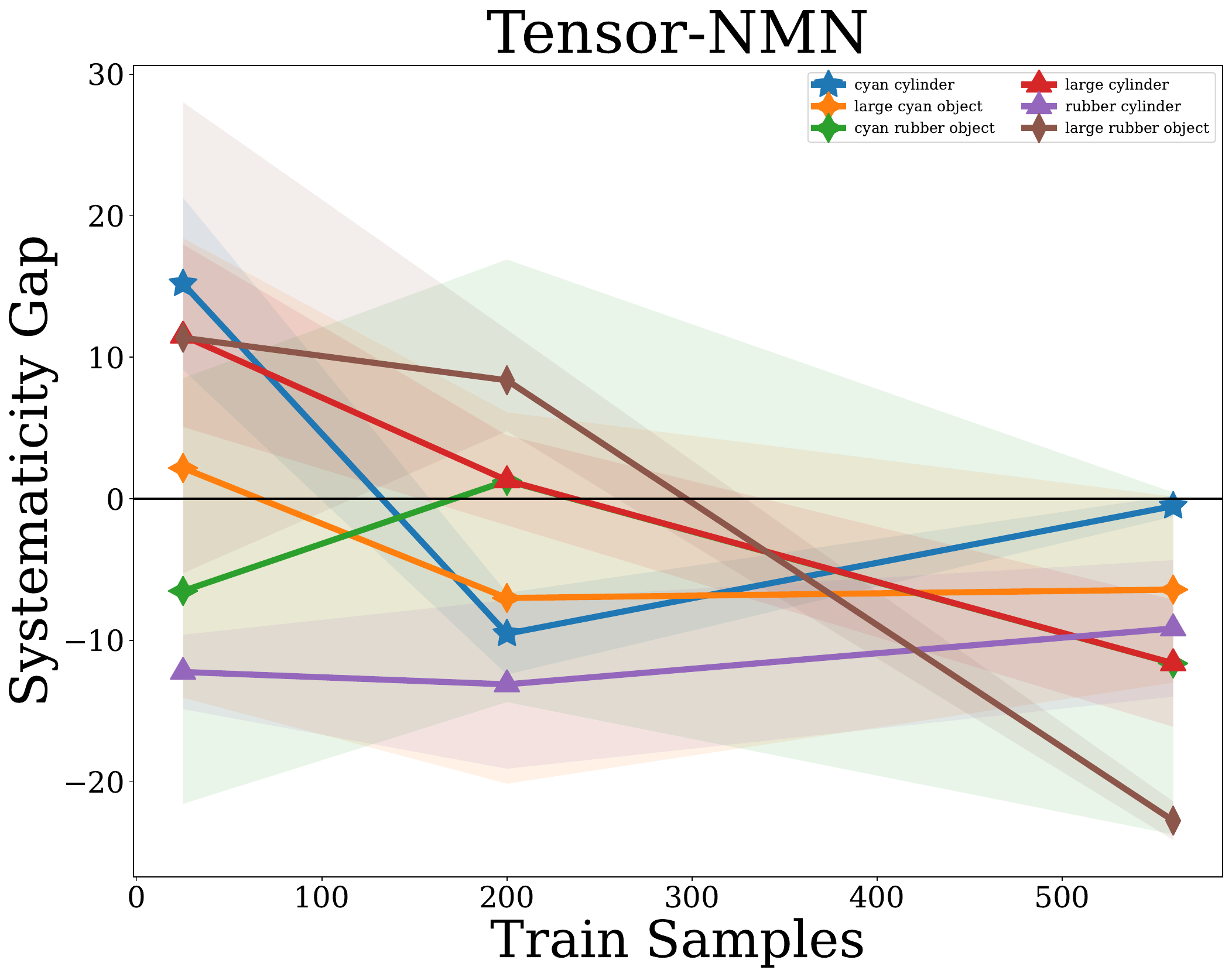}
    \caption{Systematicity gap (i.e. difference between OOD and IID model performance) for \textbf{minimal} examples, averaged over 3 runs, for each \ho.}
    \label{fig:FINAL_delta_atom_iid_ood}
\end{figure}

The systematicity gaps for each individual \ho can be found in Figures \ref{fig:FINAL_delta_iid_ood} and \ref{fig:FINAL_delta_atom_iid_ood} for the complex and minimal splits respectively.

\subsection{Detailed \nmn Systematicity Gap}

\begin{figure}[h]
    \centering
    \includegraphics[width=0.5\textwidth]{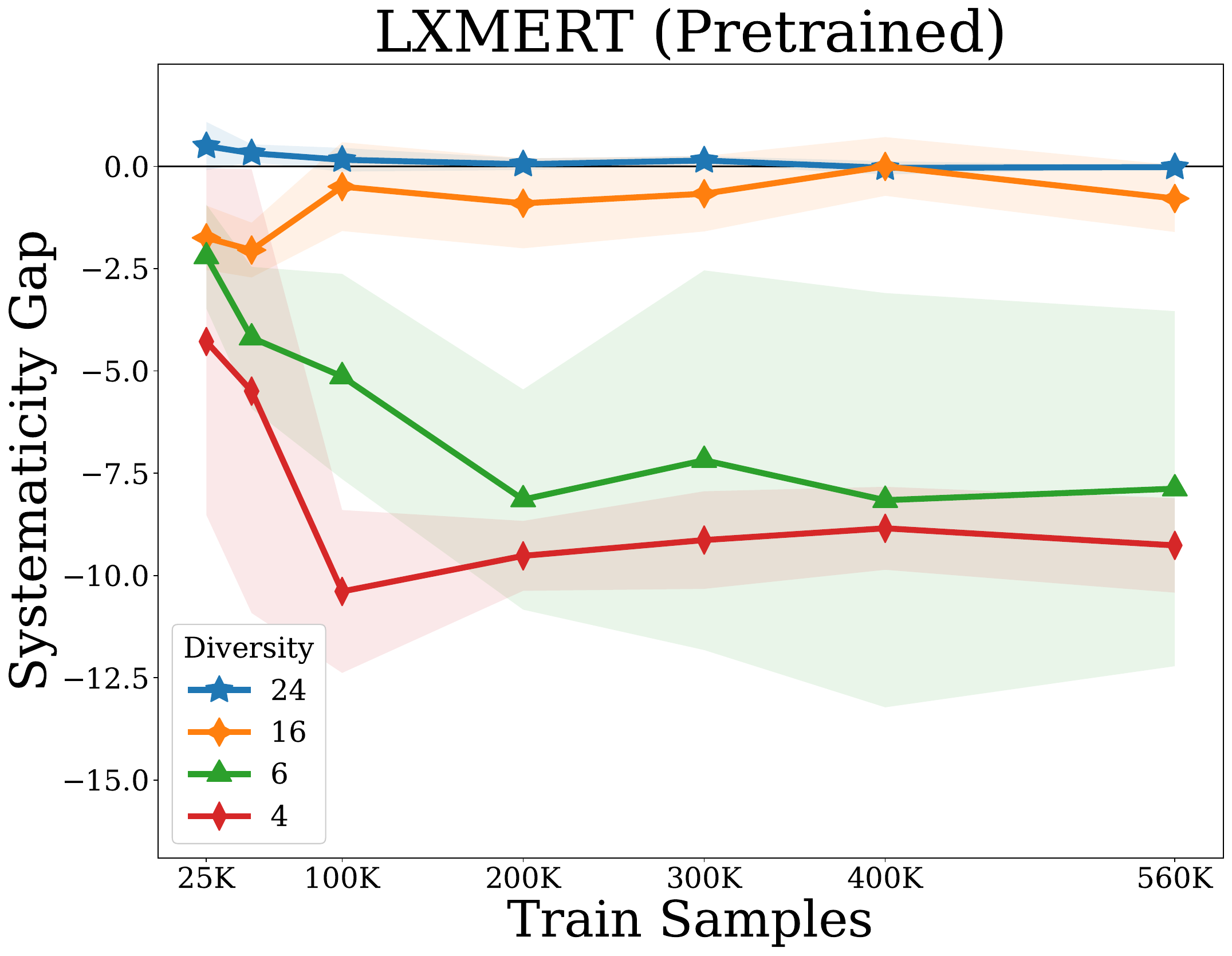}
    \includegraphics[width=0.5\textwidth]{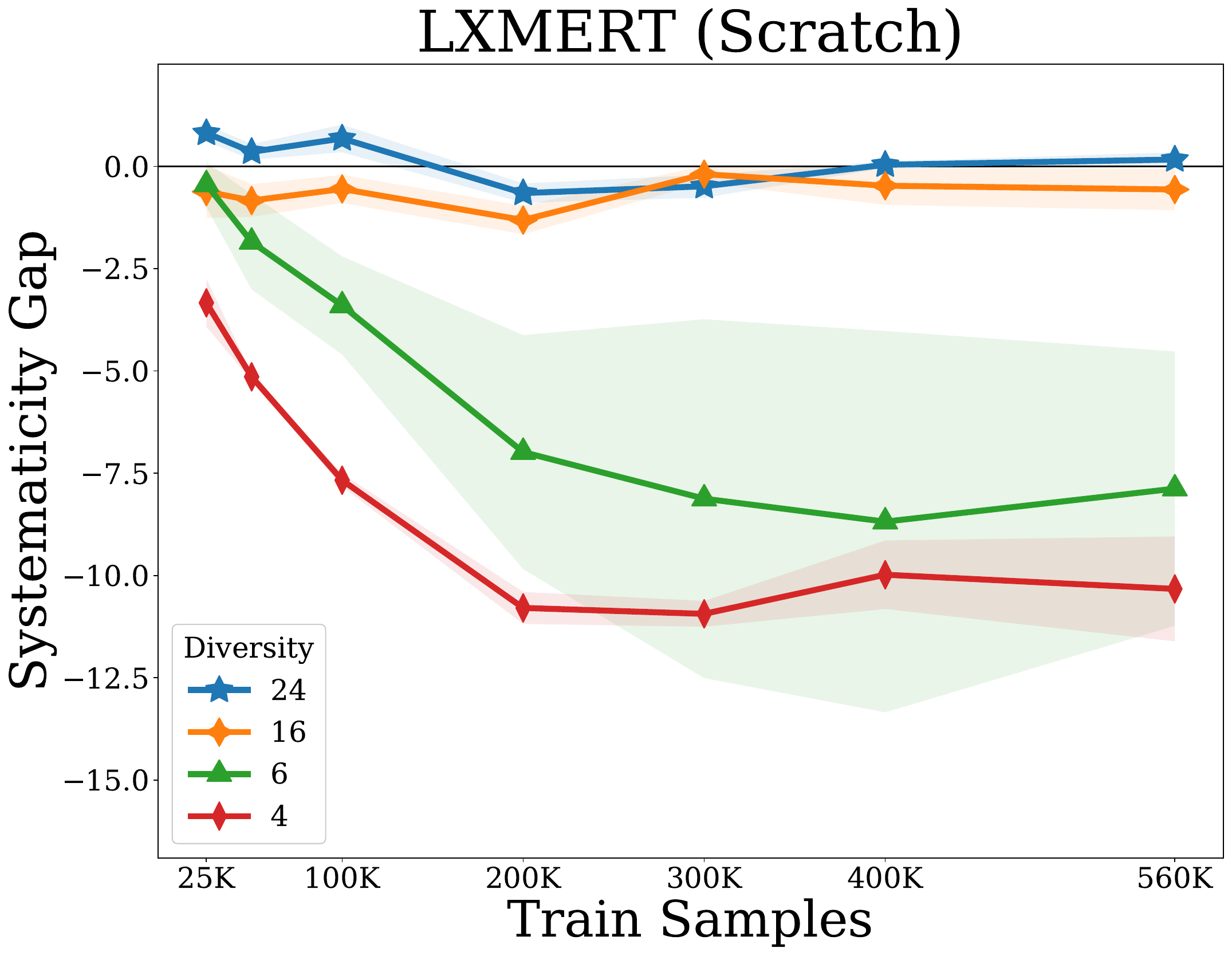}
    \includegraphics[width=0.5\textwidth]{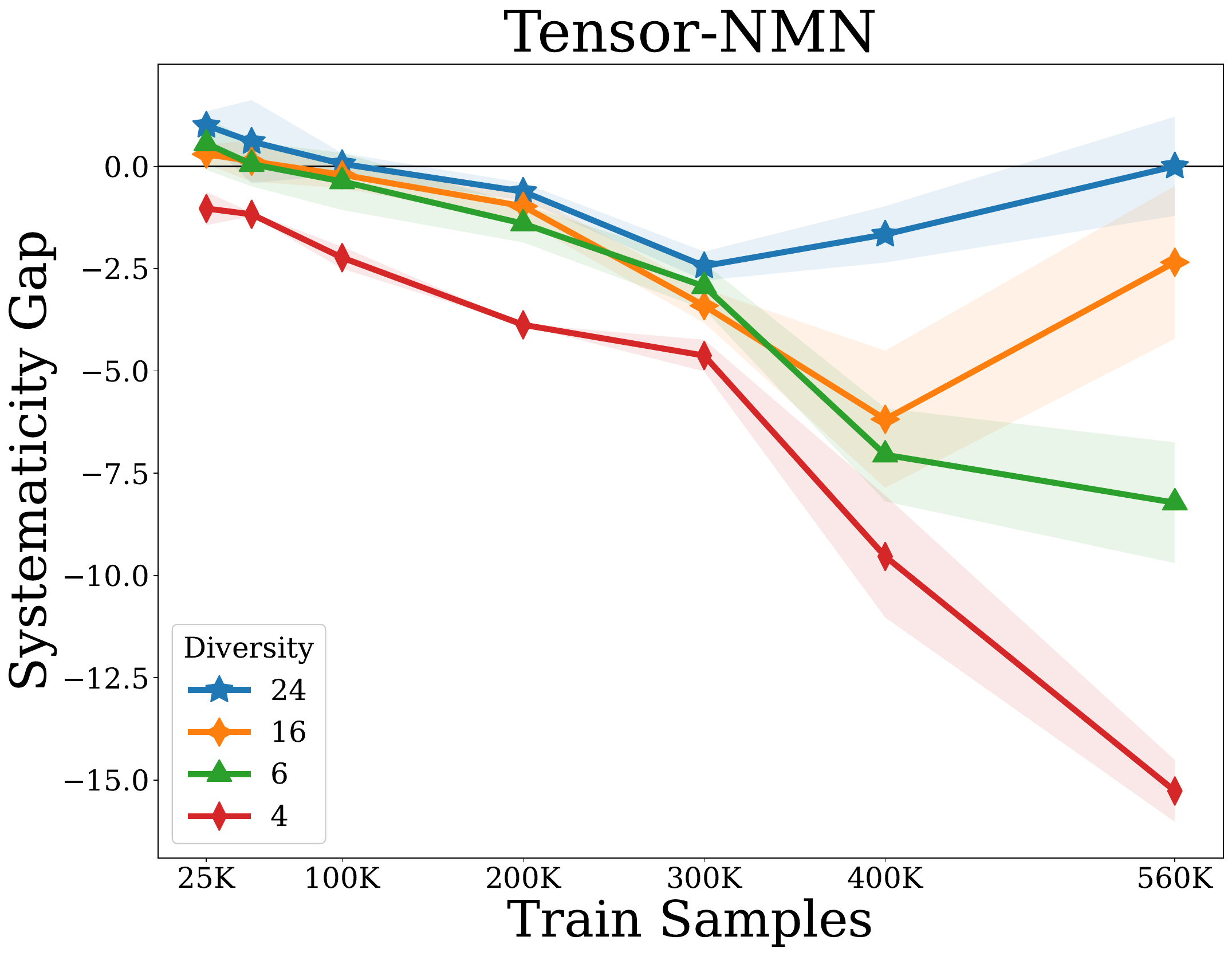}
    \caption{Systematicity gap (i.e. difference between OOD and IID model performance) for \textbf{complex} examples, averaged by \ho diversity over for the first 6 held-out attribute pairs \textit{only}, each with 3 runs.}
    \label{fig:ho05_delta_iid_ood}
\end{figure}

\begin{figure}[h]
    \centering
    \includegraphics[width=0.5\textwidth]{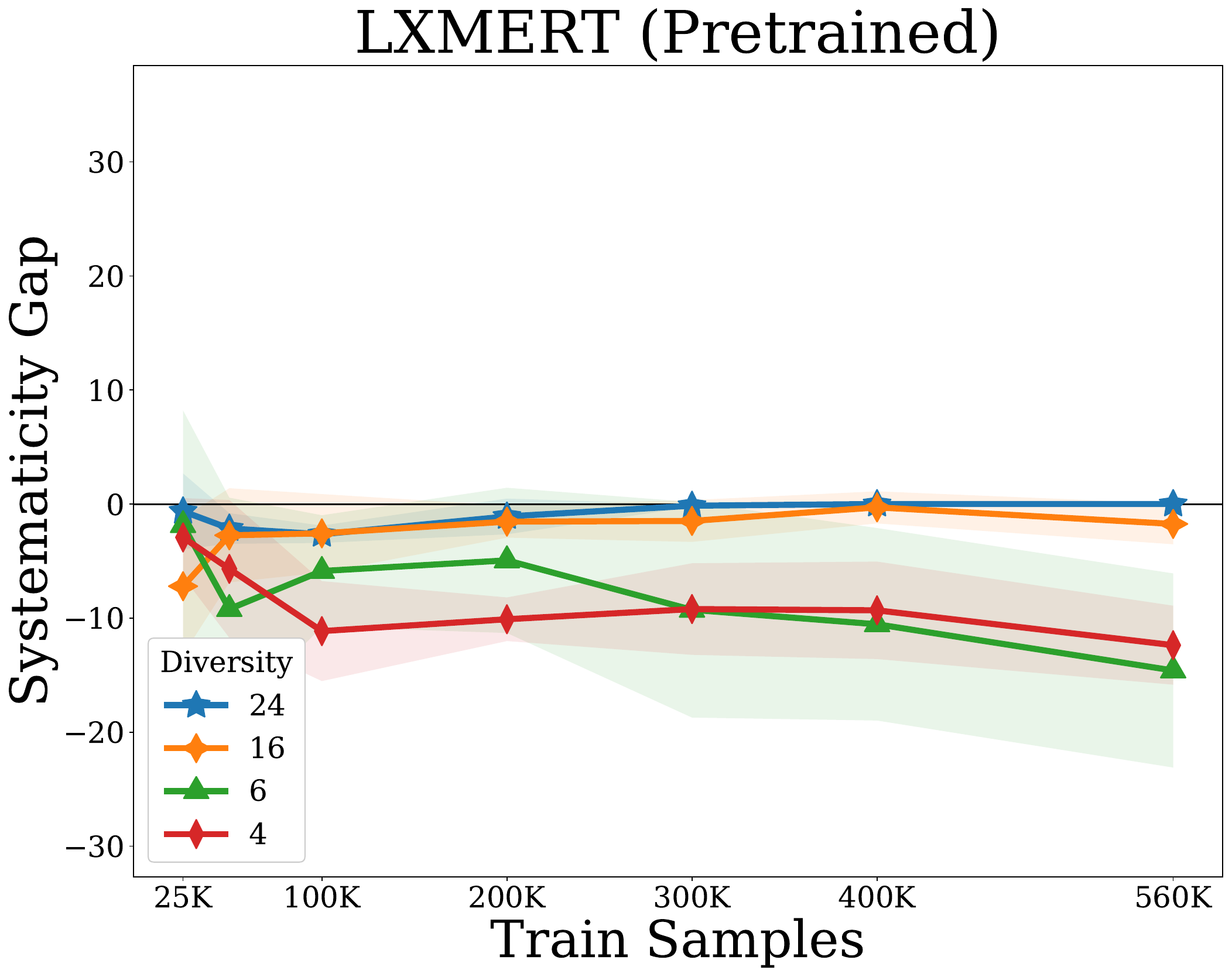}
    \includegraphics[width=0.5\textwidth]{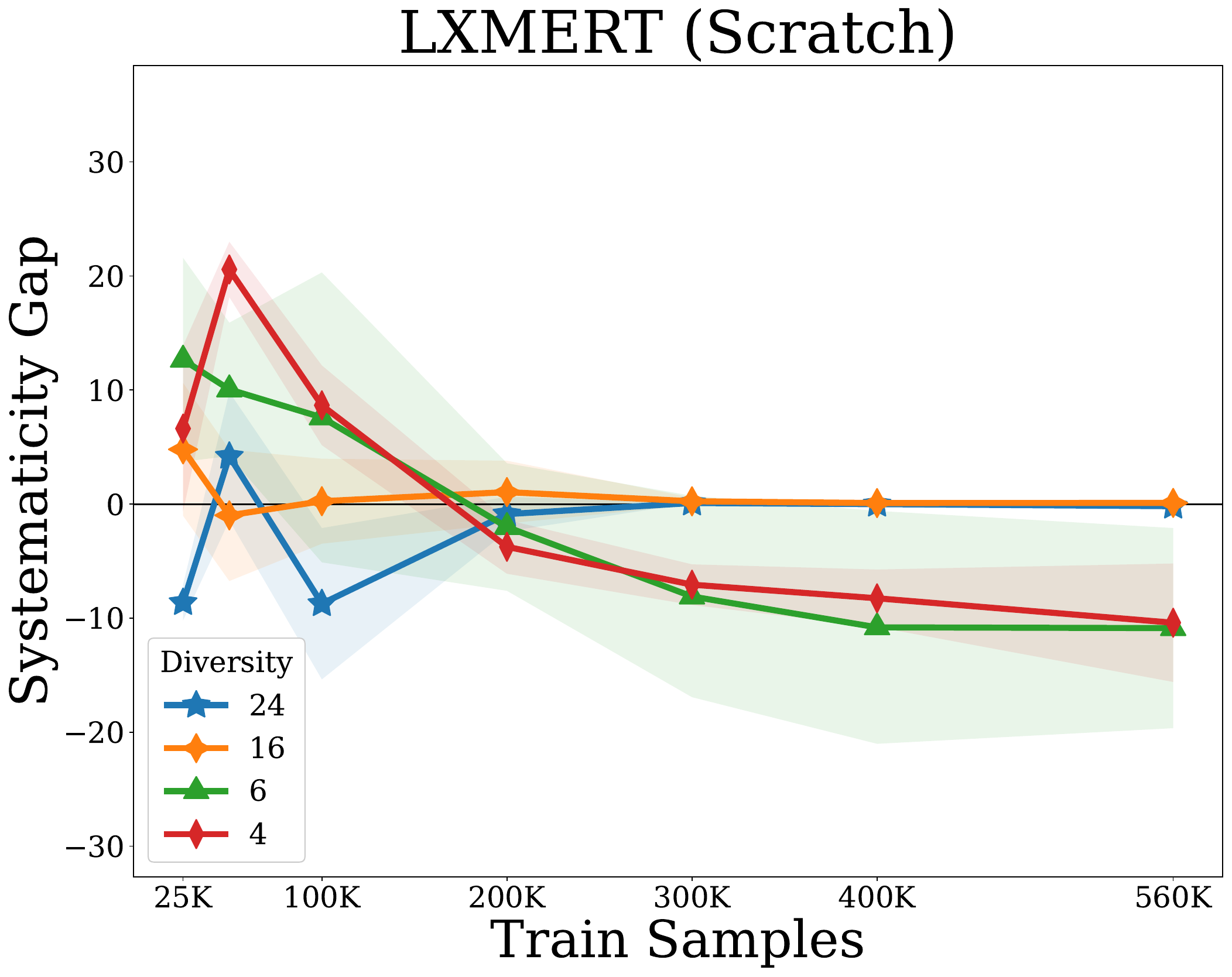}
    \includegraphics[width=0.5\textwidth]{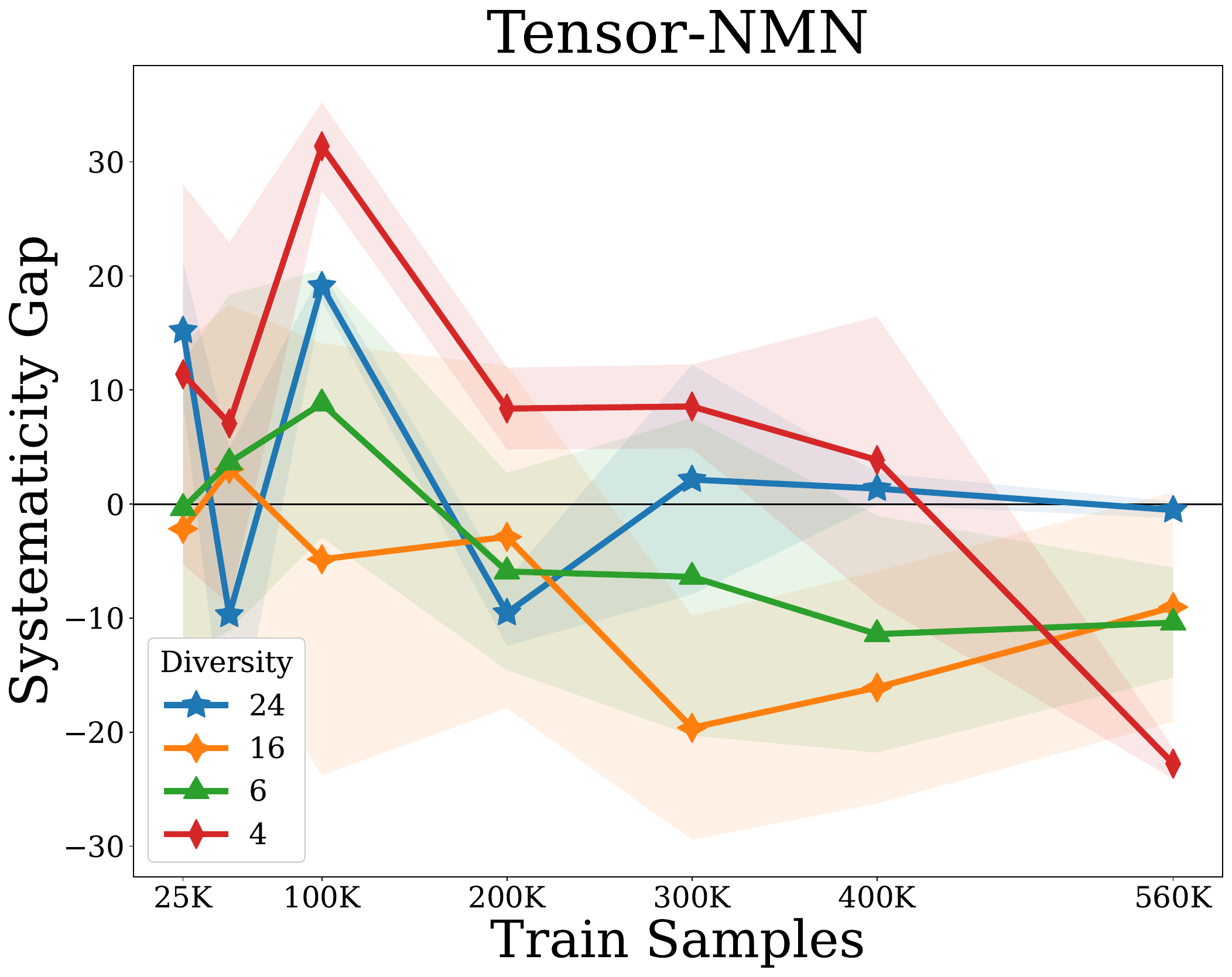}
    \caption{Systematicity gap (i.e. difference between OOD and IID model performance) for \textbf{minimal} examples, averaged by \ho diversity over for the first 6 held-out attribute pairs \textit{only}, each with 3 runs.}
    \label{fig:ho05_delta_atom_iid_ood}
\end{figure}

Averaging the systematicity gap in \nmn by diversity, we again find stratification by diversity for complex examples (see~\Cref{fig:ho05_delta_iid_ood}). The trend on minimal examples is noisier, but ultimately converges to the expected ordering (see~\Cref{fig:ho05_delta_atom_iid_ood}). Note that, as is to be expected, when limited to the first six {\ho}s the LXMERT trend is also noisier. It is therefore reasonable to expect the \nmn trend would be  
cleaner with additional {\ho}s.

\section{Summary Statistics}
The exact \lxmertFt and \lxmertScratch average accuracies and standard deviations (averaged over 3 runs) are in Tables~\ref{tab:summary_lxmert_ft_acc_aggregate_val} through \ref{tab:summary_lxmert_scratch_acc_aggregate_atom-ho}.

The exact \nmn average accuracies and standard deviations (averaged over 3 runs) are in Tables~\ref{tab:summary_nmn_acc_aggregate_val} through \ref{tab:summary_nmn_acc_aggregate_atom-ho}.

The exact average systematicity gaps and standard deviations (averaged over all runs for HOPs with the diversity in question) are in Tables~\ref{tab:sysgap_lxmert_ft_acc_aggregate_test_val} through \ref{tab:sysgap_nmn_acc_aggregate_atom-ho_filtered-atom-non-ho}.

\FloatBarrier 
\input{summary_tables}

\FloatBarrier
\section{\dataset Dataset Datasheet}\label{sec:dataset_datasheet} 
\input{datasheet}

\end{document}

%% file: model_train.tex

\textbf{Models:} Our analysis focuses on LXMERT~\cite{tan-bansal-2019-lxmert}, a multi-modal transformer-based~\cite{vaswani2017attention} architecture.
We also run experiments on a neurosymbolic model, \nmn~\cite{DBLP:conf/iccv/JohnsonHMHFZG17}, a neural module network~\cite{andreas2016neural} that  
decomposes a task into a composition of subtask-specific modules.

\textbf{Training:} For each \ho, we subsample the training set to test the impact the amount of training data has on performance. For 3 random seeds per \ho, we finetune pretrained LXMERT (\lxmertFt) and train LXMERT from scratch (\lxmertScratch). We also train \nmn from scratch, again for three runs, though only for the first 6 {\ho}s, combinations of \{\attVal{large}, \attVal{cyan}, \attVal{rubber}, \attVal{cylinder}\}.

For hyperparameter selection, we perform a grid search on the original CLEVR dataset~\cite{johnson2017clevr}. 
For further details, see~\Cref{sec:train_details}.

%% file: systematicity.tex
\begin{figure}[t]
    \centering

    \begin{subfigure}[b]{0.5\textwidth}
         \centering
         \includegraphics[width=0.725\textwidth]{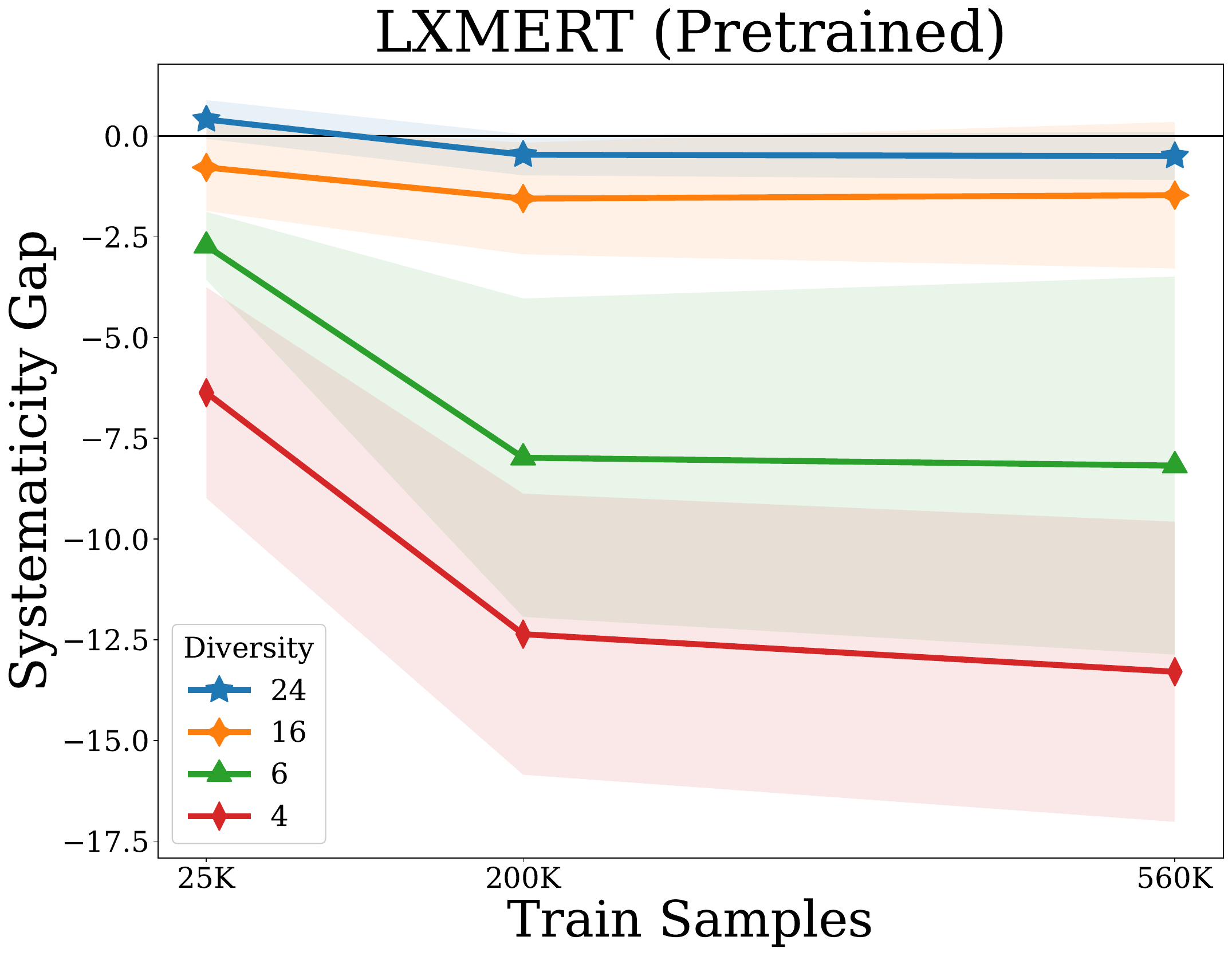}
         
         \par\medskip
         \includegraphics[width=0.725\textwidth]{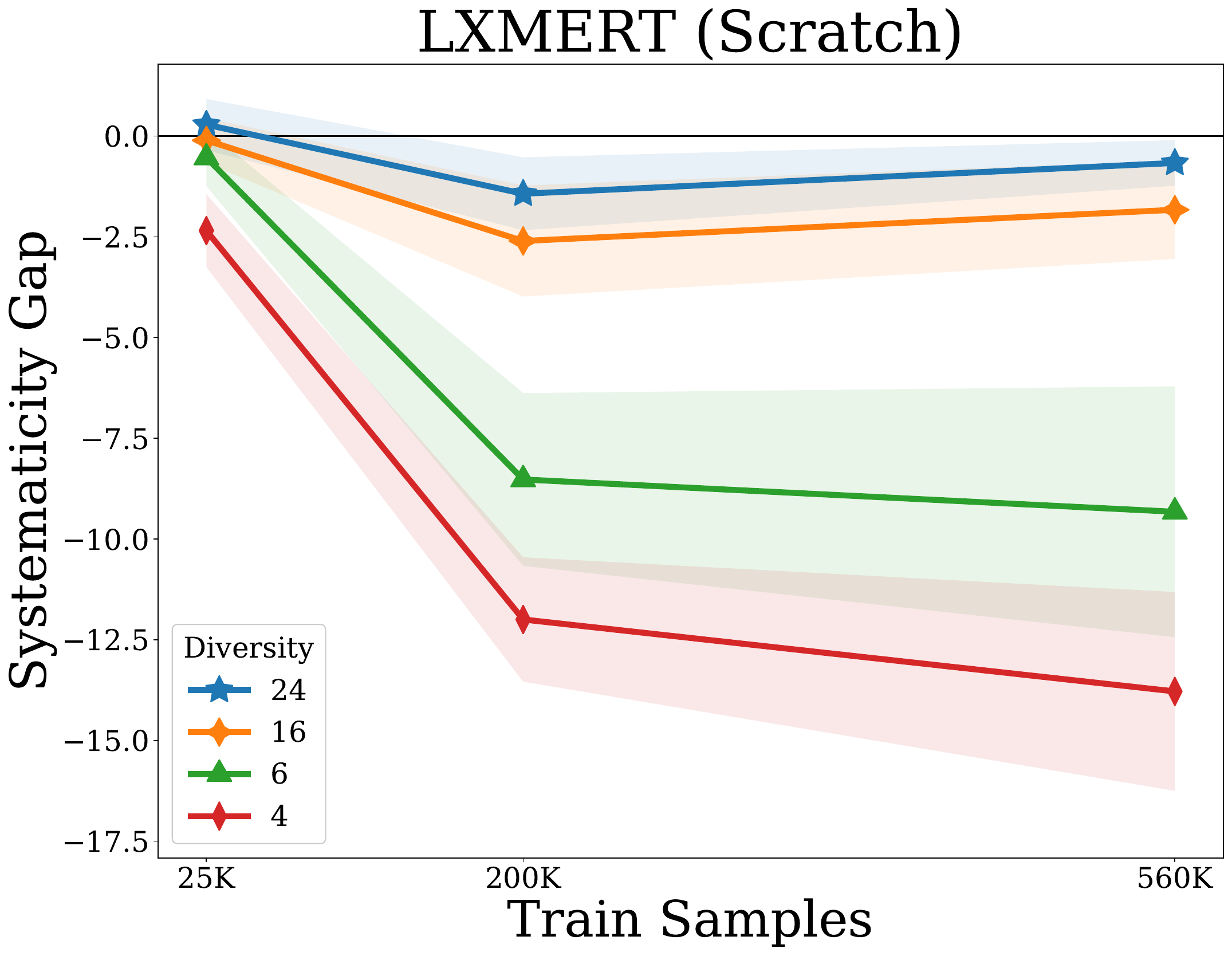}

     \end{subfigure}

    \caption{\vspace{-1mm}Systematicity gap (difference between OOD and IID accuracy) on the \textit{complex} test split, averaged by (\ho) diversity for 29 {\ho}s, each with 3 runs.}
    \label{fig:delta_iid_ood}
\end{figure}

\begin{figure}[h]
    \centering
    \includegraphics[width=0.23\textwidth]{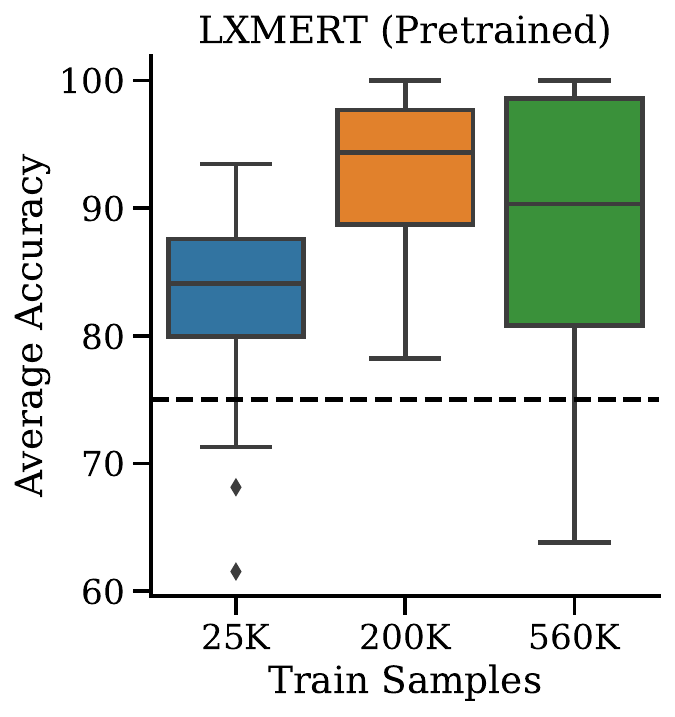}
    \includegraphics[width=0.23\textwidth]{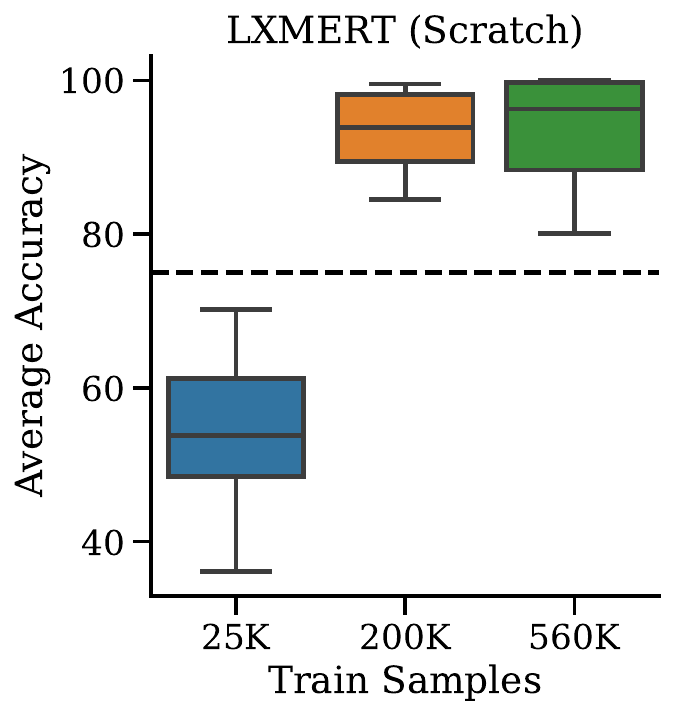}
    \caption{Box plots of \textbf{\atomHo} set performance on all 29 {\ho}s. The average performance for each \ho is produced by averaging over 3 trials. The variation captured by this boxplot is from the difference in average performance between {\ho}s, rather than from the variation within the 3 trials.}
    \label{fig:main_atom_ho}
\end{figure}

\begin{figure}[h]
    \centering
    \includegraphics[width=0.23\textwidth]{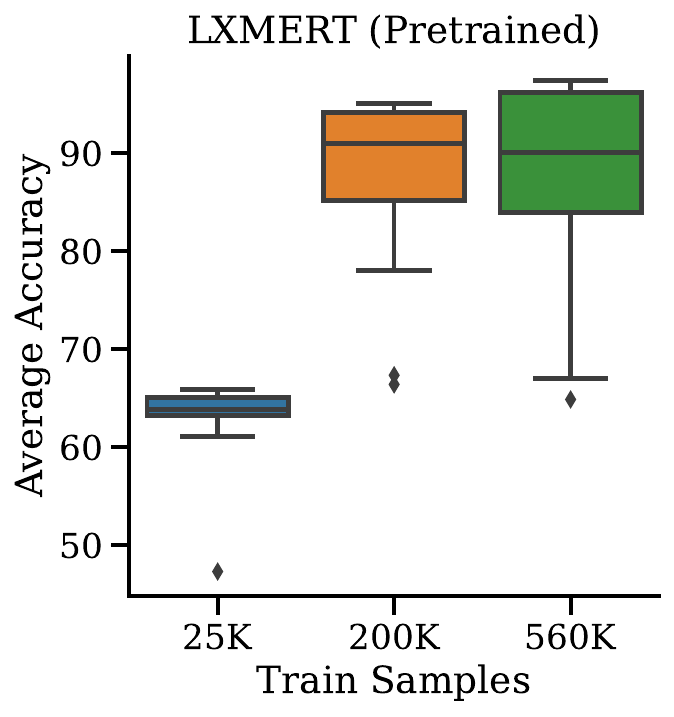}
    \includegraphics[width=0.23\textwidth]{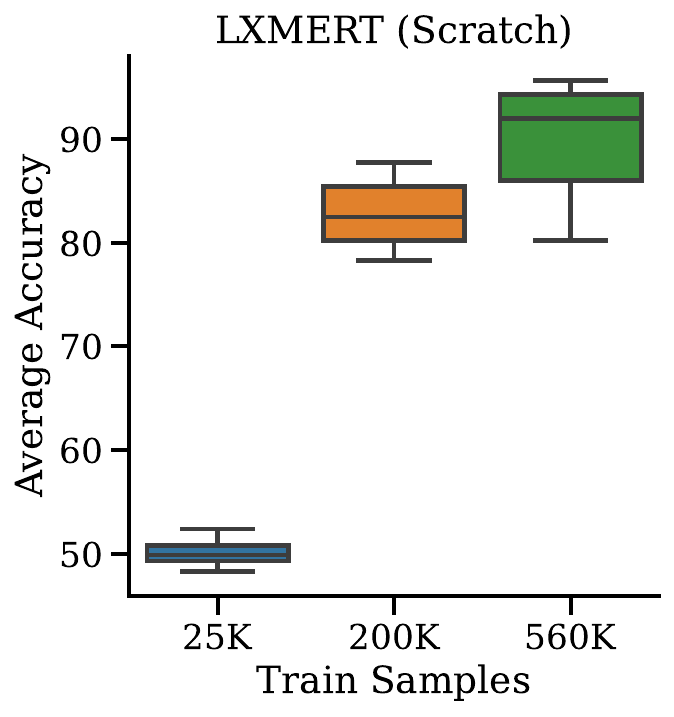}
    \caption{Box plots of \textbf{\hoTest} set performance on all 29 {\ho}s. As in \cref{fig:main_atom_ho}, each \ho is individually averaged over 3 trials. }
    \label{fig:main_ho_test}
\end{figure}

With sufficient training data, 
over 93\% of the tested model-{\ho} combinations  exceed 75\% accuracy on the \atomHo set, with some reaching 100\% (see~\Cref{fig:main_atom_ho}). 
The VQA models have a wide range of accuracies generalizing to different held out pairs. On all models tested, this accuracy varies by around 25\% across different 
{\ho}s.

Performance on the \hoTest set generally increases with the amount of training data; OOD accuracies across {\ho}s are 
similarly distributed
(see~\Cref{fig:main_ho_test}). We conclude that the models consistently exhibit at least some degree of systematicity and we observe the same trend for \nmn (see Appx.~\Cref{fig:ho05_atom_ho,fig:ho05_ho_test}).

%% file: diversity.tex
\subsection{Systematicity Gap}\label{sec:sysGap}

Knowing that our models can behave systematically, we now ask whether there is any trend in the difference between in- and out-of-distribution performance: i.e., as the size of the training set increases (and thus the model's performance generally improves), does its performance on held-out combinations approach its performance on the combinations already seen at train time? We call this performance difference, between the OOD and IID combinations, the \textit{systematicity gap}.

For example, if a model has an IID accuracy of 95\%, but only 80\% for data that requires  the model to systematically compose \attVal{rubber} and \attVal{cylinder} into the held out pair \attVal{rubber} \attVal{cylinder}, then the \textit{systematicity gap} is -15\% (i.e., a 15\% drop).

\begin{figure}[h]
    \centering
    \includegraphics[width=0.3625\textwidth]{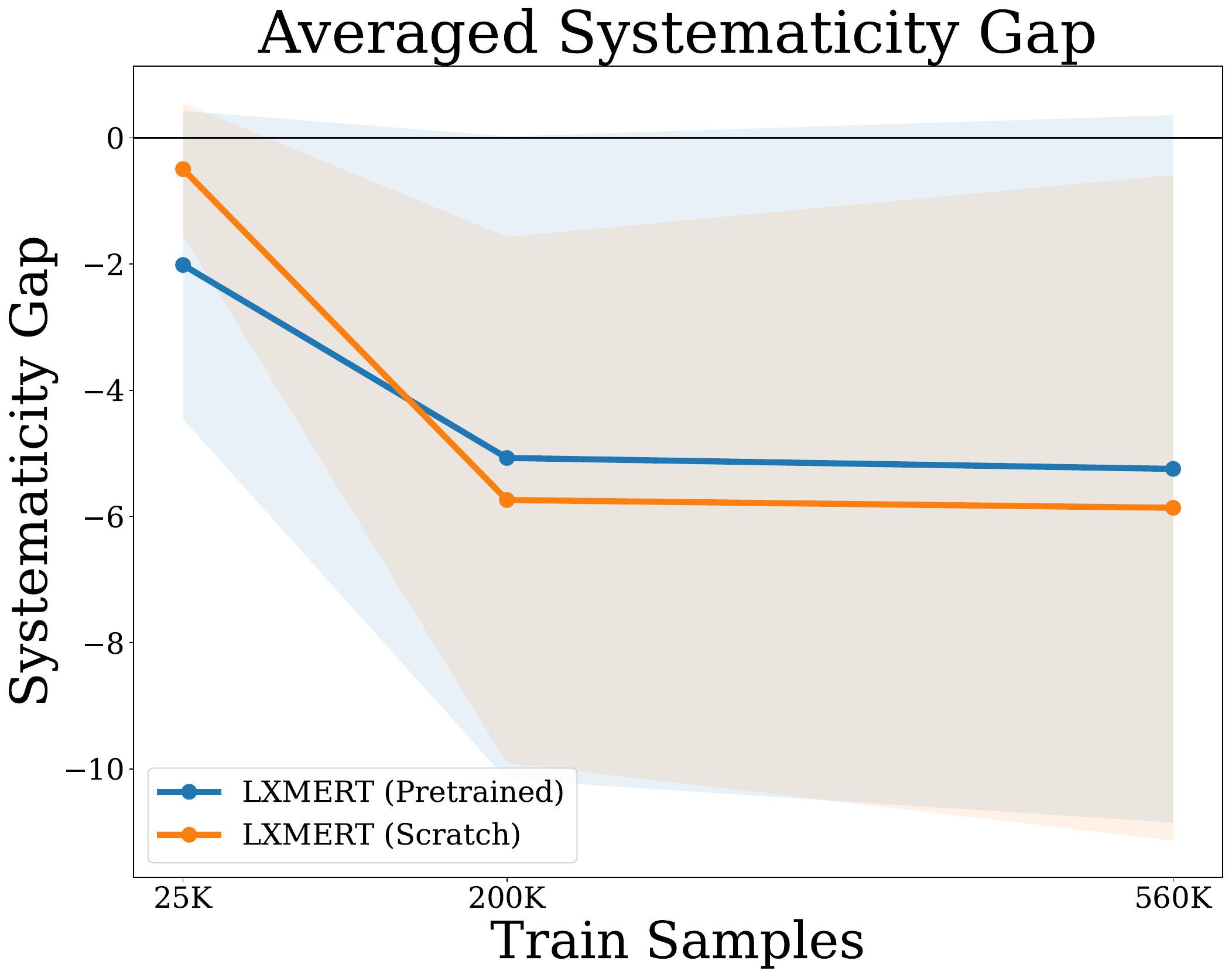}
    \caption{Average systematicity gap on \textbf{complex} examples (i.e., \hoTest accuracy minus \hoVal accuracy) with 1 standard deviation; averaged over 3 runs on each of the 29 {\ho}s. The systematicity gap plateaus, suggesting that the performance drop when generalizing to unseen combinations does not improve with additional training data. }
    \label{fig:main_sys_gap_complex}
\end{figure}

Given that the models are somewhat systematic, and that performance in general improves with more training data, one might expect that the systematicity gap would trend to zero. To the contrary, we find that, averaging over all {\ho}s, the LXMERT systematicity gap 
plateaus to a drop of 5-6\% 
(see~\Cref{fig:main_sys_gap_complex}). 
On the minimal test sets, the systematicity gap again plateaus, to a drop of 6-8\% (see
 Appx.~\Cref{fig:sys_gap_minimal}).  
The same trends are observed in \nmn (see Appx.~\Cref{fig:ho05_sys_gap_complex,fig:ho05_sys_gap_minimal}), though the systematicity gap on minimal examples widens with additional training data. 

With that said, the standard deviation of 
the observed systematicity gap is quite high -- in the following section we make the case that the nature of the training data, specifically the attribute diversity seen at train time, 
is responsible.

\subsection{Train-time conceptual diversity impacts systematicity}\label{sec:data_div}

We define \textbf{attribute diversity} as the number of possible attribute values corresponding to the unseen combination's attribute types. For example, if the unseen combination is \attVal{rubber} \attVal{cylinders}, that corresponds to the \attType{material} and \attType{shape} attribute types. Given there are 2 possible \attType{materials} and 3 possible \attType{shapes} in the training set, there are $2\times3=6$ possible \attType{material}-\attType{shape} combinations; thus the attribute diversity 
is 6.

\begin{table}[h]
    \small 
    \centering
    \begin{tabular}{c|c|c}
    \ho & Attribute Types & Diversity \\
    \hline 
    \attVal{Large rubber} & \attType{Size} + \attType{Material} & 4 \\
    \attVal{Small rubber} & \attType{Size} + \attType{Material} & 4 \\
    \attVal{Large metal} & \attType{Size} + \attType{Material} & 4 \\
    \attVal{Small metal} & \attType{Size} + \attType{Material} & 4 \\
    \hline 
    \attVal{Rubber} \attVal{cylinder} & \attType{Material} + \attType{Shape} & 6 \\
    \attVal{Metal cylinder} & \attType{Material} + \attType{Shape} & 6 \\
    \attVal{Rubber cube} & \attType{Material} + \attType{Shape} & 6 \\
    \attVal{Metal cube} & \attType{Material} + \attType{Shape} & 6 \\
    \attVal{Rubber sphere} & \attType{Material} + \attType{Shape} & 6 \\
    \hline 
    \attVal{Large cylinder} & \attType{Size} + \attType{Shape} & 6\\
    \attVal{Small cylinder} & \attType{Size} + \attType{Shape} & 6\\
    \attVal{Small cube} & \attType{Size} + \attType{Shape} & 6\\
    \attVal{Large cube} & \attType{Size} + \attType{Shape} & 6\\
    \attVal{Small sphere} & \attType{Size} + \attType{Shape} & 6\\
    \hline 
    \attVal{Rubber cyan} & \attType{Material} + \attType{Color} & 16 \\
    \attVal{Rubber brown} & \attType{Material} + \attType{Color} & 16 \\
    \attVal{Rubber purple} & \attType{Material} + \attType{Color} & 16 \\
    \attVal{Metal red} & \attType{Material} + \attType{Color} & 16 \\
    \attVal{Metal gray} & \attType{Material} + \attType{Color} & 16 \\
    \hline 
    \attVal{Large cyan} & \attType{Size} + \attType{Color} & 16 \\
    \attVal{Small brown} & \attType{Size} + \attType{Color} & 16 \\
    \attVal{Small purple} & \attType{Size} + \attType{Color} & 16 \\
    \attVal{Small red} & \attType{Size} + \attType{Color} & 16 \\
    \attVal{Large gray} & \attType{Size} + \attType{Color} & 16 \\
    \hline 
    \attVal{Cyan cylinder} & \attType{Color} + \attType{Shape} & 24 \\
    \attVal{Brown sphere} & \attType{Color} + \attType{Shape} & 24 \\
    \attVal{Red cylinder} & \attType{Color} + \attType{Shape} & 24 \\
    \attVal{Gray cube} & \attType{Color} + \attType{Shape} & 24 \\
    \attVal{Purple sphere} & \attType{Color} + \attType{Shape} & 24 \\
\end{tabular}
    \caption{{\ho} Diversity; i.e.,  number of attribute values corresponding to the {\ho}'s attribute types.}
    \label{tab:all_hos}
\end{table}

~\Cref{tab:all_hos} lists the attribute diversity of the {\ho}s.
Since the CLEVR training distribution is uniform across object attribute values, 
for a \train set of fixed size, as attribute diversity increases, 
the number of examples per combination decreases. 

\Cref{fig:delta_iid_ood} again illustrates the systematicity gap, but now only averages over {\ho}s of the same diversity (rather than over \textit{all} {\ho}s as in~\Cref{sec:sysGap}). 
With this, we see that the systematicity gap is stratified by the diversity of the combinations seen at train time. Specifically, as the diversity of the training data increases, the systematicity gap narrows. In fact, the gap is typically near or within a standard deviation of zero 
for diversities of 16 or above. In comparison, diversities of 6 show a 
a plateauing systematicity gap 
stabilizing at 7-14\%. We observe similar results with the systematicity gap of the minimal test sets (see Appx.~\Cref{fig:delta_atom_iid_ood}).

For \nmn, we also find stratification by diversity for complex examples (see Appx.~\Cref{fig:ho05_delta_iid_ood}). The trend on minimal examples is noisier,  
but converges to the expected ordering (see Appx.~\Cref{fig:ho05_delta_atom_iid_ood}).

\subsection{Controlling for attribute category}

A particular attribute can introduce confounders, such as the overall difficulty of learning its category.
We ran additional experiments explicitly controlling for the  attribute category to verify diversity's impact on the systematicity gap. In our prior experiments, attribute diversity is intrinsically tied to attribute type. As seen in~\Cref{tab:all_hos}, the most diverse pairs are always \attType{color}-\attType{shape} combinations, and the least diverse pairs are always \attType{size}-\attType{material} combinations. Thus, it is possible that we are actually measuring the effects of attribute type on generalization, rather than diversity. To address this, here we vary the attribute diversity while keeping the attribute type combination fixed. 

\begin{figure}[h]
    \centering
    \includegraphics[width=0.3625\textwidth]{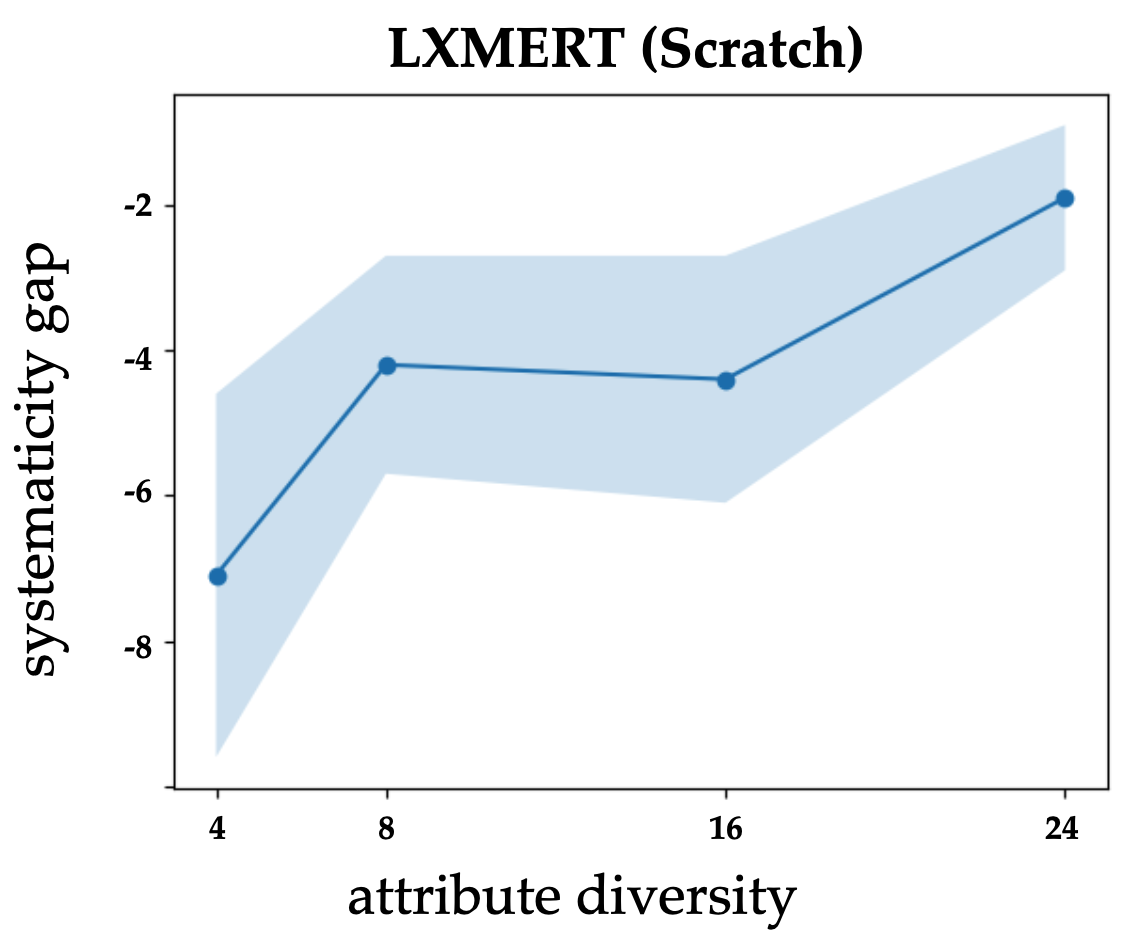}
    \caption{For attribute pair \attType{color} + \attType{shape}, we control the diversity by subsampling fixed number of combinations (one of [4, 8, 16, 24]), and finetuning the model accordingly. On the complex test sets, we observe that increasing attribute diversity reduces systematicity gap.}
    \label{fig:ablation1}
\end{figure}

We focused on \attType{shape}-\attType{color} combinations 
and generated multiple datasets with varying levels of diversity [4, 8, 16, 24] by varying the unique color-shape combinations present during training. We trained separate instances of LXMERT-s on these datasets and evaluated performance on corresponding HOPs (averaged across 3 random seeds). In~\cref{fig:ablation1}, we see that lower attribute diversity led to worse systematicity gap.

\label{sec:ablation_attr}

%% file: summary_tables.tex
\begin{table*}[h]
    \small 
    \centering
\begin{tabular}{lrlll}
\hline
 HOP                  &   Diversity & 25k                 & 200k                & 560k                \\
\hline
 cyan cylinder        &          24 & $64.80 \pm 0.13\%$  & $95.03 \pm 0.05\%$  & $97.36 \pm 0.05\%$  \\
 brown sphere         &          24 & $65.02 \pm 0.15\%$  & $95.09 \pm 0.01\%$  & $97.43 \pm 0.02\%$  \\
 red cylinder         &          24 & $65.02 \pm 0.23\%$  & $95.07 \pm 0.04\%$  & $96.25 \pm 0.97\%$  \\
 gray cube            &          24 & $65.53 \pm 0.23\%$  & $94.90 \pm 0.13\%$  & $69.88 \pm 38.90\%$ \\
 purple sphere        &          24 & $64.85 \pm 0.52\%$  & $94.71 \pm 0.03\%$  & $97.27 \pm 0.12\%$  \\
 large cyan object    &          16 & $65.32 \pm 0.22\%$  & $94.86 \pm 0.11\%$  & $97.34 \pm 0.05\%$  \\
 cyan rubber object   &          16 & $65.70 \pm 0.21\%$  & $94.35 \pm 0.69\%$  & $97.27 \pm 0.09\%$  \\
 brown rubber object  &          16 & $65.55 \pm 0.15\%$  & $94.88 \pm 0.10\%$  & $97.33 \pm 0.05\%$  \\
 small brown object   &          16 & $65.23 \pm 0.04\%$  & $95.28 \pm 0.16\%$  & $71.86 \pm 36.14\%$ \\
 red metal object     &          16 & $64.92 \pm 0.14\%$  & $95.00 \pm 0.08\%$  & $97.48 \pm 0.03\%$  \\
 small red object     &          16 & $65.19 \pm 0.15\%$  & $94.71 \pm 0.50\%$  & $97.33 \pm 0.02\%$  \\
 gray metal object    &          16 & $65.31 \pm 0.28\%$  & $94.75 \pm 0.11\%$  & $97.29 \pm 0.04\%$  \\
 large gray object    &          16 & $64.98 \pm 0.05\%$  & $94.83 \pm 0.24\%$  & $97.22 \pm 0.24\%$  \\
 purple rubber object &          16 & $65.14 \pm 0.06\%$  & $94.85 \pm 0.07\%$  & $97.31 \pm 0.07\%$  \\
 small purple object  &          16 & $64.60 \pm 0.17\%$  & $94.58 \pm 0.31\%$  & $97.37 \pm 0.07\%$  \\
 large cylinder       &           6 & $66.75 \pm 0.08\%$  & $94.44 \pm 0.93\%$  & $97.64 \pm 0.03\%$  \\
 rubber cylinder      &           6 & $66.62 \pm 0.20\%$  & $95.11 \pm 0.08\%$  & $97.35 \pm 0.22\%$  \\
 rubber sphere        &           6 & $66.38 \pm 0.21\%$  & $95.13 \pm 0.14\%$  & $97.45 \pm 0.07\%$  \\
 small sphere         &           6 & $65.65 \pm 0.28\%$  & $95.14 \pm 0.16\%$  & $97.44 \pm 0.04\%$  \\
 metal cylinder       &           6 & $66.38 \pm 0.31\%$  & $95.17 \pm 0.24\%$  & $71.77 \pm 36.57\%$ \\
 small cylinder       &           6 & $67.06 \pm 0.21\%$  & $95.07 \pm 0.31\%$  & $97.62 \pm 0.19\%$  \\
 metal cube           &           6 & $66.04 \pm 0.41\%$  & $95.18 \pm 0.10\%$  & $71.79 \pm 36.61\%$ \\
 large cube           &           6 & $66.24 \pm 0.13\%$  & $95.49 \pm 0.08\%$  & $97.88 \pm 0.02\%$  \\
 rubber cube          &           6 & $66.93 \pm 0.36\%$  & $70.18 \pm 35.34\%$ & $97.49 \pm 0.32\%$  \\
 small cube           &           6 & $65.95 \pm 0.07\%$  & $70.30 \pm 35.03\%$ & $70.67 \pm 38.16\%$ \\
 large rubber object  &           4 & $51.60 \pm 24.05\%$ & $95.23 \pm 0.15\%$  & $97.65 \pm 0.05\%$  \\
 small rubber object  &           4 & $69.59 \pm 0.18\%$  & $95.87 \pm 0.08\%$  & $97.69 \pm 0.27\%$  \\
 small metal object   &           4 & $68.69 \pm 0.31\%$  & $95.84 \pm 0.12\%$  & $97.91 \pm 0.13\%$  \\
 large metal object   &           4 & $66.96 \pm 0.52\%$  & $95.70 \pm 0.13\%$  & $97.95 \pm 0.05\%$  \\
\hline
\end{tabular}    \caption{LXMERT (Pretrained) \textbf{complex-IID} average accuracy and standard deviation over 3 runs with different random seeds. Average accuracies are reported for each HOP (row) and each training set size (column).}
    \label{tab:summary_lxmert_ft_acc_aggregate_val}
\end{table*}

\begin{table*}[h]
    \small 
    \centering
\begin{tabular}{lrlll}
\hline
 HOP                  &   Diversity & 25k                 & 200k                & 560k                \\
\hline
 cyan cylinder        &          24 & $65.29 \pm 0.48\%$  & $95.08 \pm 0.15\%$  & $97.34 \pm 0.08\%$  \\
 brown sphere         &          24 & $65.11 \pm 0.08\%$  & $94.04 \pm 0.40\%$  & $96.20 \pm 0.22\%$  \\
 red cylinder         &          24 & $65.36 \pm 0.11\%$  & $94.63 \pm 0.08\%$  & $95.59 \pm 1.32\%$  \\
 gray cube            &          24 & $65.60 \pm 0.50\%$  & $94.19 \pm 0.19\%$  & $69.15 \pm 38.40\%$ \\
 purple sphere        &          24 & $65.92 \pm 0.69\%$  & $94.55 \pm 0.57\%$  & $97.43 \pm 0.09\%$  \\
 large cyan object    &          16 & $64.08 \pm 0.30\%$  & $94.70 \pm 0.09\%$  & $97.19 \pm 0.08\%$  \\
 cyan rubber object   &          16 & $63.44 \pm 0.70\%$  & $92.69 \pm 1.82\%$  & $95.85 \pm 0.73\%$  \\
 brown rubber object  &          16 & $63.69 \pm 0.20\%$  & $93.31 \pm 0.09\%$  & $96.02 \pm 0.14\%$  \\
 small brown object   &          16 & $63.57 \pm 0.31\%$  & $91.02 \pm 0.17\%$  & $70.20 \pm 33.16\%$ \\
 red metal object     &          16 & $65.72 \pm 0.68\%$  & $94.56 \pm 0.26\%$  & $96.82 \pm 0.26\%$  \\
 small red object     &          16 & $64.84 \pm 0.45\%$  & $92.50 \pm 1.09\%$  & $95.72 \pm 0.11\%$  \\
 gray metal object    &          16 & $64.08 \pm 0.31\%$  & $91.37 \pm 0.37\%$  & $91.53 \pm 0.58\%$  \\
 large gray object    &          16 & $64.24 \pm 0.17\%$  & $94.37 \pm 0.36\%$  & $96.96 \pm 0.28\%$  \\
 purple rubber object &          16 & $65.45 \pm 0.22\%$  & $94.37 \pm 0.20\%$  & $96.41 \pm 0.38\%$  \\
 small purple object  &          16 & $65.05 \pm 0.62\%$  & $93.67 \pm 0.34\%$  & $96.42 \pm 0.33\%$  \\
 large cylinder       &           6 & $65.69 \pm 0.74\%$  & $88.60 \pm 2.68\%$  & $93.76 \pm 2.15\%$  \\
 rubber cylinder      &           6 & $63.26 \pm 0.15\%$  & $84.66 \pm 0.79\%$  & $85.46 \pm 1.23\%$  \\
 rubber sphere        &           6 & $63.17 \pm 0.57\%$  & $81.14 \pm 0.77\%$  & $81.17 \pm 1.60\%$  \\
 small sphere         &           6 & $63.23 \pm 0.33\%$  & $88.92 \pm 0.41\%$  & $90.06 \pm 0.84\%$  \\
 metal cylinder       &           6 & $63.20 \pm 0.64\%$  & $86.97 \pm 1.39\%$  & $67.05 \pm 31.47\%$ \\
 small cylinder       &           6 & $63.78 \pm 0.21\%$  & $85.20 \pm 0.91\%$  & $88.01 \pm 0.20\%$  \\
 metal cube           &           6 & $63.27 \pm 0.78\%$  & $83.82 \pm 0.68\%$  & $64.88 \pm 30.50\%$ \\
 large cube           &           6 & $63.84 \pm 0.09\%$  & $88.33 \pm 1.78\%$  & $88.95 \pm 1.11\%$  \\
 rubber cube          &           6 & $63.34 \pm 0.07\%$  & $66.41 \pm 30.84\%$ & $88.78 \pm 1.65\%$  \\
 small cube           &           6 & $63.98 \pm 0.26\%$  & $67.35 \pm 30.99\%$ & $67.21 \pm 35.83\%$ \\
 large rubber object  &           4 & $47.32 \pm 19.82\%$ & $85.71 \pm 1.01\%$  & $88.39 \pm 1.12\%$  \\
 small rubber object  &           4 & $61.10 \pm 0.32\%$  & $78.04 \pm 0.55\%$  & $79.62 \pm 0.56\%$  \\
 small metal object   &           4 & $61.87 \pm 0.54\%$  & $83.05 \pm 0.08\%$  & $83.94 \pm 2.44\%$  \\
 large metal object   &           4 & $61.07 \pm 0.59\%$  & $86.40 \pm 0.13\%$  & $86.08 \pm 2.69\%$  \\
\hline
\end{tabular}    \caption{LXMERT (Pretrained) \textbf{complex-OOD} average accuracy and standard deviation over 3 runs with different random seeds. Average accuracies are reported for each HOP (row) and each training set size (column).}
    \label{tab:summary_lxmert_ft_acc_aggregate_test}
\end{table*}

\begin{table*}[h]
    \small 
    \centering
\begin{tabular}{lrlll}
\hline
 HOP                  &   Diversity & 25k                 & 200k                & 560k                \\
\hline
 cyan cylinder        &          24 & $90.89 \pm 2.49\%$  & $99.97 \pm 0.02\%$  & $100.00 \pm 0.00\%$ \\
 brown sphere         &          24 & $92.33 \pm 1.49\%$  & $99.98 \pm 0.01\%$  & $100.00 \pm 0.00\%$ \\
 red cylinder         &          24 & $92.05 \pm 1.98\%$  & $99.99 \pm 0.00\%$  & $99.89 \pm 0.16\%$  \\
 gray cube            &          24 & $92.57 \pm 1.40\%$  & $99.95 \pm 0.03\%$  & $78.36 \pm 30.60\%$ \\
 purple sphere        &          24 & $86.66 \pm 4.47\%$  & $99.91 \pm 0.07\%$  & $99.99 \pm 0.01\%$  \\
 large cyan object    &          16 & $94.65 \pm 1.06\%$  & $99.97 \pm 0.01\%$  & $99.98 \pm 0.01\%$  \\
 cyan rubber object   &          16 & $91.62 \pm 1.09\%$  & $99.81 \pm 0.05\%$  & $99.97 \pm 0.01\%$  \\
 brown rubber object  &          16 & $91.63 \pm 1.05\%$  & $99.58 \pm 0.08\%$  & $99.93 \pm 0.01\%$  \\
 small brown object   &          16 & $90.81 \pm 1.49\%$  & $99.93 \pm 0.03\%$  & $91.23 \pm 12.40\%$ \\
 red metal object     &          16 & $91.15 \pm 1.33\%$  & $99.72 \pm 0.02\%$  & $99.97 \pm 0.01\%$  \\
 small red object     &          16 & $92.06 \pm 0.66\%$  & $98.60 \pm 1.89\%$  & $99.99 \pm 0.01\%$  \\
 gray metal object    &          16 & $90.09 \pm 1.86\%$  & $99.52 \pm 0.53\%$  & $99.98 \pm 0.01\%$  \\
 large gray object    &          16 & $94.20 \pm 1.19\%$  & $99.84 \pm 0.11\%$  & $99.98 \pm 0.02\%$  \\
 purple rubber object &          16 & $88.69 \pm 2.03\%$  & $99.77 \pm 0.05\%$  & $99.96 \pm 0.02\%$  \\
 small purple object  &          16 & $93.05 \pm 0.41\%$  & $99.97 \pm 0.02\%$  & $99.99 \pm 0.01\%$  \\
 large cylinder       &           6 & $81.81 \pm 3.51\%$  & $97.42 \pm 3.37\%$  & $99.97 \pm 0.01\%$  \\
 rubber cylinder      &           6 & $77.60 \pm 6.47\%$  & $99.61 \pm 0.15\%$  & $99.99 \pm 0.00\%$  \\
 rubber sphere        &           6 & $81.61 \pm 3.88\%$  & $99.75 \pm 0.07\%$  & $99.87 \pm 0.02\%$  \\
 small sphere         &           6 & $90.59 \pm 1.41\%$  & $99.93 \pm 0.04\%$  & $99.93 \pm 0.03\%$  \\
 metal cylinder       &           6 & $85.59 \pm 5.81\%$  & $99.84 \pm 0.10\%$  & $76.46 \pm 33.26\%$ \\
 small cylinder       &           6 & $86.79 \pm 2.68\%$  & $99.95 \pm 0.03\%$  & $99.99 \pm 0.01\%$  \\
 metal cube           &           6 & $75.06 \pm 7.55\%$  & $99.53 \pm 0.35\%$  & $77.36 \pm 31.95\%$ \\
 large cube           &           6 & $89.61 \pm 1.98\%$  & $99.98 \pm 0.02\%$  & $100.00 \pm 0.00\%$ \\
 rubber cube          &           6 & $73.00 \pm 1.91\%$  & $85.84 \pm 19.75\%$ & $99.94 \pm 0.06\%$  \\
 small cube           &           6 & $81.08 \pm 2.96\%$  & $90.02 \pm 13.74\%$ & $73.28 \pm 37.77\%$ \\
 large rubber object  &           4 & $64.46 \pm 28.99\%$ & $99.74 \pm 0.03\%$  & $99.98 \pm 0.01\%$  \\
 small rubber object  &           4 & $89.38 \pm 1.37\%$  & $99.85 \pm 0.09\%$  & $99.99 \pm 0.01\%$  \\
 small metal object   &           4 & $86.15 \pm 2.22\%$  & $99.90 \pm 0.08\%$  & $99.89 \pm 0.06\%$  \\
 large metal object   &           4 & $85.80 \pm 2.25\%$  & $99.92 \pm 0.03\%$  & $99.91 \pm 0.01\%$  \\
\hline
\end{tabular}    \caption{LXMERT (Pretrained) \textbf{minimal-IID} average accuracy and standard deviation over 3 runs with different random seeds. Average accuracies are reported for each HOP (row) and each training set size (column).}
    \label{tab:summary_lxmert_ft_acc_aggregate_filtered-atom-non-ho}
\end{table*}

\begin{table*}[h]
    \small 
    \centering
\begin{tabular}{lrlll}
\hline
 HOP                  &   Diversity & 25k                 & 200k                & 560k                \\
\hline
 cyan cylinder        &          24 & $90.25 \pm 0.82\%$  & $98.88 \pm 1.58\%$  & $100.00 \pm 0.00\%$ \\
 brown sphere         &          24 & $88.76 \pm 3.74\%$  & $99.78 \pm 0.18\%$  & $99.26 \pm 0.46\%$  \\
 red cylinder         &          24 & $90.33 \pm 1.04\%$  & $98.74 \pm 0.64\%$  & $98.96 \pm 1.47\%$  \\
 gray cube            &          24 & $84.15 \pm 1.28\%$  & $99.70 \pm 0.11\%$  & $75.37 \pm 34.67\%$ \\
 purple sphere        &          24 & $93.45 \pm 6.74\%$  & $100.00 \pm 0.00\%$ & $100.00 \pm 0.00\%$ \\
 large cyan object    &          16 & $90.60 \pm 4.23\%$  & $99.48 \pm 0.31\%$  & $99.84 \pm 0.06\%$  \\
 cyan rubber object   &          16 & $81.27 \pm 4.82\%$  & $97.22 \pm 1.23\%$  & $96.63 \pm 1.12\%$  \\
 brown rubber object  &          16 & $84.84 \pm 2.14\%$  & $96.90 \pm 1.17\%$  & $98.13 \pm 1.08\%$  \\
 small brown object   &          16 & $83.17 \pm 3.10\%$  & $92.14 \pm 0.99\%$  & $88.57 \pm 9.60\%$  \\
 red metal object     &          16 & $87.34 \pm 4.08\%$  & $97.18 \pm 0.62\%$  & $98.53 \pm 0.76\%$  \\
 small red object     &          16 & $87.10 \pm 3.48\%$  & $95.16 \pm 6.68\%$  & $99.60 \pm 0.40\%$  \\
 gray metal object    &          16 & $85.52 \pm 1.83\%$  & $93.13 \pm 2.58\%$  & $85.20 \pm 6.46\%$  \\
 large gray object    &          16 & $84.13 \pm 2.25\%$  & $99.25 \pm 1.07\%$  & $99.84 \pm 0.15\%$  \\
 purple rubber object &          16 & $85.83 \pm 4.27\%$  & $97.70 \pm 0.62\%$  & $98.61 \pm 0.95\%$  \\
 small purple object  &          16 & $90.75 \pm 1.31\%$  & $94.37 \pm 0.98\%$  & $96.35 \pm 2.66\%$  \\
 large cylinder       &           6 & $87.58 \pm 5.31\%$  & $96.91 \pm 3.45\%$  & $91.47 \pm 8.00\%$  \\
 rubber cylinder      &           6 & $68.14 \pm 2.73\%$  & $90.25 \pm 6.35\%$  & $79.31 \pm 2.58\%$  \\
 rubber sphere        &           6 & $71.30 \pm 8.29\%$  & $80.13 \pm 1.34\%$  & $82.83 \pm 5.02\%$  \\
 small sphere         &           6 & $84.04 \pm 3.47\%$  & $95.10 \pm 0.49\%$  & $94.10 \pm 1.60\%$  \\
 metal cylinder       &           6 & $74.71 \pm 6.63\%$  & $88.76 \pm 2.50\%$  & $63.80 \pm 27.45\%$ \\
 small cylinder       &           6 & $82.37 \pm 4.19\%$  & $81.02 \pm 5.24\%$  & $80.82 \pm 1.63\%$  \\
 metal cube           &           6 & $74.75 \pm 5.48\%$  & $88.72 \pm 2.93\%$  & $68.84 \pm 28.41\%$ \\
 large cube           &           6 & $87.75 \pm 3.30\%$  & $93.89 \pm 4.10\%$  & $90.34 \pm 6.22\%$  \\
 rubber cube          &           6 & $74.32 \pm 4.40\%$  & $81.38 \pm 15.25\%$ & $84.96 \pm 7.58\%$  \\
 small cube           &           6 & $80.35 \pm 0.17\%$  & $87.70 \pm 9.12\%$  & $68.15 \pm 39.08\%$ \\
 large rubber object  &           4 & $61.54 \pm 27.48\%$ & $89.64 \pm 1.88\%$  & $87.61 \pm 3.48\%$  \\
 small rubber object  &           4 & $73.79 \pm 1.94\%$  & $78.21 \pm 2.26\%$  & $76.04 \pm 0.91\%$  \\
 small metal object   &           4 & $79.95 \pm 3.57\%$  & $86.15 \pm 3.16\%$  & $79.51 \pm 3.97\%$  \\
 large metal object   &           4 & $83.54 \pm 4.87\%$  & $85.86 \pm 4.14\%$  & $86.27 \pm 8.20\%$  \\
\hline
\end{tabular}    \caption{LXMERT (Pretrained) \textbf{minimal-OOD} average accuracy and standard deviation over 3 runs with different random seeds. Average accuracies are reported for each HOP (row) and each training set size (column).}
    \label{tab:summary_lxmert_ft_acc_aggregate_atom-ho}
\end{table*}

\begin{table*}[h]
    \small 
    \centering
\begin{tabular}{lrlll}
\hline
 HOP                  &   Diversity & 25k                & 200k               & 560k               \\
\hline
 cyan cylinder        &          24 & $49.05 \pm 0.41\%$ & $86.74 \pm 1.90\%$ & $94.75 \pm 0.54\%$ \\
 brown sphere         &          24 & $48.77 \pm 0.23\%$ & $88.69 \pm 0.25\%$ & $95.60 \pm 0.23\%$ \\
 red cylinder         &          24 & $49.44 \pm 0.29\%$ & $85.45 \pm 1.99\%$ & $95.56 \pm 0.29\%$ \\
 gray cube            &          24 & $49.41 \pm 0.64\%$ & $81.59 \pm 1.95\%$ & $95.02 \pm 0.42\%$ \\
 purple sphere        &          24 & $49.60 \pm 0.94\%$ & $86.01 \pm 5.30\%$ & $95.13 \pm 0.42\%$ \\
 large cyan object    &          16 & $49.54 \pm 0.74\%$ & $83.34 \pm 1.77\%$ & $95.83 \pm 0.30\%$ \\
 cyan rubber object   &          16 & $49.59 \pm 0.70\%$ & $86.97 \pm 1.66\%$ & $95.71 \pm 0.33\%$ \\
 brown rubber object  &          16 & $49.16 \pm 0.36\%$ & $88.87 \pm 1.06\%$ & $95.52 \pm 0.65\%$ \\
 small brown object   &          16 & $49.22 \pm 0.34\%$ & $87.78 \pm 2.15\%$ & $96.21 \pm 0.17\%$ \\
 red metal object     &          16 & $49.29 \pm 0.27\%$ & $89.25 \pm 1.86\%$ & $95.70 \pm 0.14\%$ \\
 small red object     &          16 & $49.13 \pm 0.47\%$ & $87.76 \pm 1.07\%$ & $95.53 \pm 0.26\%$ \\
 gray metal object    &          16 & $48.95 \pm 0.53\%$ & $85.17 \pm 2.57\%$ & $95.88 \pm 0.27\%$ \\
 large gray object    &          16 & $50.06 \pm 0.92\%$ & $82.79 \pm 4.83\%$ & $95.77 \pm 0.07\%$ \\
 purple rubber object &          16 & $48.31 \pm 0.08\%$ & $86.51 \pm 0.25\%$ & $95.31 \pm 0.14\%$ \\
 small purple object  &          16 & $49.59 \pm 0.49\%$ & $88.13 \pm 1.41\%$ & $95.77 \pm 0.09\%$ \\
 large cylinder       &           6 & $52.66 \pm 1.68\%$ & $91.39 \pm 1.48\%$ & $96.56 \pm 0.15\%$ \\
 rubber cylinder      &           6 & $51.87 \pm 0.88\%$ & $89.82 \pm 0.64\%$ & $96.25 \pm 0.25\%$ \\
 rubber sphere        &           6 & $50.21 \pm 0.71\%$ & $90.07 \pm 0.69\%$ & $96.24 \pm 0.08\%$ \\
 small sphere         &           6 & $50.01 \pm 0.58\%$ & $91.56 \pm 0.89\%$ & $96.12 \pm 0.07\%$ \\
 metal cylinder       &           6 & $51.87 \pm 0.78\%$ & $90.57 \pm 1.05\%$ & $96.58 \pm 0.08\%$ \\
 small cylinder       &           6 & $52.01 \pm 1.18\%$ & $91.29 \pm 1.87\%$ & $96.53 \pm 0.06\%$ \\
 metal cube           &           6 & $50.34 \pm 0.33\%$ & $90.57 \pm 1.09\%$ & $96.29 \pm 0.15\%$ \\
 large cube           &           6 & $52.44 \pm 0.90\%$ & $91.34 \pm 0.92\%$ & $96.72 \pm 0.13\%$ \\
 rubber cube          &           6 & $50.38 \pm 0.76\%$ & $91.13 \pm 0.85\%$ & $96.45 \pm 0.15\%$ \\
 small cube           &           6 & $50.69 \pm 0.58\%$ & $91.75 \pm 0.47\%$ & $96.68 \pm 0.17\%$ \\
 large rubber object  &           4 & $54.28 \pm 0.47\%$ & $89.77 \pm 0.72\%$ & $96.31 \pm 0.20\%$ \\
 small rubber object  &           4 & $53.33 \pm 0.90\%$ & $92.14 \pm 0.65\%$ & $96.91 \pm 0.24\%$ \\
 small metal object   &           4 & $51.94 \pm 0.49\%$ & $90.97 \pm 0.69\%$ & $96.84 \pm 0.28\%$ \\
 large metal object   &           4 & $54.42 \pm 0.66\%$ & $89.87 \pm 2.50\%$ & $96.77 \pm 0.17\%$ \\
\hline
\end{tabular}    \caption{LXMERT (Scratch) \textbf{complex-IID} average accuracy and standard deviation over 3 runs with different random seeds. Average accuracies are reported for each HOP (row) and each training set size (column).}
    \label{tab:summary_lxmert_scratch_acc_aggregate_val}
\end{table*}

\begin{table*}[h]
    \small 
    \centering
\begin{tabular}{lrlll}
\hline
 HOP                  &   Diversity & 25k                & 200k               & 560k               \\
\hline
 cyan cylinder        &          24 & $49.86 \pm 0.31\%$ & $86.08 \pm 1.80\%$ & $94.92 \pm 0.68\%$ \\
 brown sphere         &          24 & $49.46 \pm 0.04\%$ & $87.24 \pm 0.40\%$ & $94.64 \pm 0.38\%$ \\
 red cylinder         &          24 & $50.20 \pm 0.41\%$ & $83.70 \pm 2.24\%$ & $94.90 \pm 0.19\%$ \\
 gray cube            &          24 & $49.23 \pm 0.32\%$ & $78.86 \pm 1.74\%$ & $93.60 \pm 0.55\%$ \\
 purple sphere        &          24 & $48.94 \pm 0.80\%$ & $85.44 \pm 5.63\%$ & $94.67 \pm 0.58\%$ \\
 large cyan object    &          16 & $48.35 \pm 0.43\%$ & $82.03 \pm 1.74\%$ & $94.78 \pm 0.48\%$ \\
 cyan rubber object   &          16 & $49.54 \pm 0.47\%$ & $85.65 \pm 2.12\%$ & $95.63 \pm 0.24\%$ \\
 brown rubber object  &          16 & $49.31 \pm 0.49\%$ & $85.95 \pm 1.48\%$ & $94.17 \pm 1.01\%$ \\
 small brown object   &          16 & $49.78 \pm 0.26\%$ & $82.61 \pm 2.81\%$ & $91.87 \pm 0.37\%$ \\
 red metal object     &          16 & $49.21 \pm 0.37\%$ & $87.74 \pm 2.22\%$ & $94.61 \pm 0.09\%$ \\
 small red object     &          16 & $49.04 \pm 0.09\%$ & $84.42 \pm 1.02\%$ & $92.90 \pm 0.67\%$ \\
 gray metal object    &          16 & $48.60 \pm 0.35\%$ & $80.64 \pm 2.27\%$ & $92.56 \pm 0.15\%$ \\
 large gray object    &          16 & $50.33 \pm 0.75\%$ & $80.34 \pm 4.06\%$ & $94.11 \pm 0.30\%$ \\
 purple rubber object &          16 & $48.29 \pm 0.38\%$ & $84.71 \pm 0.59\%$ & $94.06 \pm 0.31\%$ \\
 small purple object  &          16 & $49.33 \pm 0.53\%$ & $86.43 \pm 1.87\%$ & $94.27 \pm 0.10\%$ \\
 large cylinder       &           6 & $52.40 \pm 1.33\%$ & $87.06 \pm 2.46\%$ & $91.94 \pm 0.63\%$ \\
 rubber cylinder      &           6 & $51.24 \pm 0.48\%$ & $80.18 \pm 1.71\%$ & $85.12 \pm 0.67\%$ \\
 rubber sphere        &           6 & $49.89 \pm 0.55\%$ & $78.99 \pm 1.82\%$ & $83.34 \pm 0.48\%$ \\
 small sphere         &           6 & $50.54 \pm 0.41\%$ & $84.70 \pm 1.24\%$ & $89.78 \pm 0.56\%$ \\
 metal cylinder       &           6 & $50.87 \pm 0.72\%$ & $81.76 \pm 0.75\%$ & $88.00 \pm 0.49\%$ \\
 small cylinder       &           6 & $51.01 \pm 1.12\%$ & $82.43 \pm 2.17\%$ & $86.01 \pm 2.06\%$ \\
 metal cube           &           6 & $50.47 \pm 0.52\%$ & $79.56 \pm 1.94\%$ & $81.98 \pm 1.20\%$ \\
 large cube           &           6 & $50.83 \pm 0.72\%$ & $82.49 \pm 1.37\%$ & $87.40 \pm 1.06\%$ \\
 rubber cube          &           6 & $49.52 \pm 0.33\%$ & $81.98 \pm 1.08\%$ & $86.71 \pm 0.90\%$ \\
 small cube           &           6 & $50.39 \pm 0.89\%$ & $85.11 \pm 0.51\%$ & $90.91 \pm 0.09\%$ \\
 large rubber object  &           4 & $50.94 \pm 0.22\%$ & $78.98 \pm 0.83\%$ & $85.98 \pm 1.34\%$ \\
 small rubber object  &           4 & $51.22 \pm 0.87\%$ & $78.31 \pm 1.62\%$ & $80.23 \pm 0.39\%$ \\
 small metal object   &           4 & $50.78 \pm 0.36\%$ & $78.53 \pm 0.93\%$ & $81.94 \pm 0.49\%$ \\
 large metal object   &           4 & $51.63 \pm 0.47\%$ & $78.94 \pm 2.44\%$ & $83.54 \pm 0.42\%$ \\
\hline
\end{tabular}    \caption{LXMERT (Scratch) \textbf{complex-OOD} average accuracy and standard deviation over 3 runs with different random seeds. Average accuracies are reported for each HOP (row) and each training set size (column).}
    \label{tab:summary_lxmert_scratch_acc_aggregate_test}
\end{table*}

\begin{table*}[h]
    \small 
    \centering
\begin{tabular}{lrlll}
\hline
 HOP                  &   Diversity & 25k                & 200k               & 560k               \\
\hline
 cyan cylinder        &          24 & $47.40 \pm 4.40\%$ & $99.02 \pm 0.54\%$ & $99.96 \pm 0.01\%$ \\
 brown sphere         &          24 & $48.37 \pm 2.65\%$ & $98.74 \pm 0.45\%$ & $99.97 \pm 0.03\%$ \\
 red cylinder         &          24 & $60.03 \pm 5.11\%$ & $98.32 \pm 1.54\%$ & $99.95 \pm 0.03\%$ \\
 gray cube            &          24 & $60.73 \pm 3.48\%$ & $98.72 \pm 0.48\%$ & $99.93 \pm 0.03\%$ \\
 purple sphere        &          24 & $49.28 \pm 5.04\%$ & $99.44 \pm 0.21\%$ & $99.96 \pm 0.03\%$ \\
 large cyan object    &          16 & $60.52 \pm 3.26\%$ & $96.72 \pm 2.22\%$ & $99.87 \pm 0.10\%$ \\
 cyan rubber object   &          16 & $61.60 \pm 1.37\%$ & $98.60 \pm 0.32\%$ & $99.89 \pm 0.06\%$ \\
 brown rubber object  &          16 & $62.04 \pm 5.68\%$ & $99.53 \pm 0.04\%$ & $99.70 \pm 0.17\%$ \\
 small brown object   &          16 & $55.37 \pm 3.64\%$ & $98.73 \pm 0.74\%$ & $99.80 \pm 0.16\%$ \\
 red metal object     &          16 & $60.21 \pm 3.89\%$ & $98.31 \pm 0.29\%$ & $99.95 \pm 0.03\%$ \\
 small red object     &          16 & $66.29 \pm 2.51\%$ & $99.23 \pm 0.34\%$ & $99.82 \pm 0.22\%$ \\
 gray metal object    &          16 & $53.61 \pm 0.64\%$ & $98.51 \pm 0.47\%$ & $99.97 \pm 0.02\%$ \\
 large gray object    &          16 & $49.47 \pm 3.32\%$ & $99.36 \pm 0.11\%$ & $99.95 \pm 0.00\%$ \\
 purple rubber object &          16 & $57.13 \pm 5.64\%$ & $98.22 \pm 0.77\%$ & $99.92 \pm 0.04\%$ \\
 small purple object  &          16 & $62.36 \pm 4.10\%$ & $99.35 \pm 0.44\%$ & $99.97 \pm 0.03\%$ \\
 large cylinder       &           6 & $48.47 \pm 7.39\%$ & $95.77 \pm 1.04\%$ & $99.92 \pm 0.07\%$ \\
 rubber cylinder      &           6 & $38.64 \pm 3.31\%$ & $98.71 \pm 0.73\%$ & $99.90 \pm 0.04\%$ \\
 rubber sphere        &           6 & $39.95 \pm 6.05\%$ & $98.12 \pm 0.59\%$ & $99.72 \pm 0.05\%$ \\
 small sphere         &           6 & $48.61 \pm 3.31\%$ & $99.13 \pm 0.52\%$ & $97.38 \pm 2.38\%$ \\
 metal cylinder       &           6 & $38.36 \pm 1.81\%$ & $94.38 \pm 2.12\%$ & $99.96 \pm 0.00\%$ \\
 small cylinder       &           6 & $39.51 \pm 5.54\%$ & $96.51 \pm 1.99\%$ & $99.97 \pm 0.01\%$ \\
 metal cube           &           6 & $40.55 \pm 4.83\%$ & $99.11 \pm 0.27\%$ & $99.92 \pm 0.02\%$ \\
 large cube           &           6 & $43.91 \pm 4.48\%$ & $99.24 \pm 0.95\%$ & $99.97 \pm 0.01\%$ \\
 rubber cube          &           6 & $48.91 \pm 0.93\%$ & $98.90 \pm 0.60\%$ & $99.91 \pm 0.08\%$ \\
 small cube           &           6 & $36.78 \pm 1.94\%$ & $99.68 \pm 0.37\%$ & $99.88 \pm 0.15\%$ \\
 large rubber object  &           4 & $37.95 \pm 4.58\%$ & $93.24 \pm 3.61\%$ & $99.93 \pm 0.02\%$ \\
 small rubber object  &           4 & $44.15 \pm 1.84\%$ & $96.51 \pm 1.36\%$ & $99.81 \pm 0.10\%$ \\
 small metal object   &           4 & $43.83 \pm 1.89\%$ & $94.47 \pm 1.34\%$ & $99.94 \pm 0.08\%$ \\
 large metal object   &           4 & $44.12 \pm 4.62\%$ & $99.05 \pm 0.76\%$ & $99.93 \pm 0.03\%$ \\
\hline
\end{tabular}    \caption{LXMERT (Scratch) \textbf{minimal-IID} average accuracy and standard deviation over 3 runs with different random seeds. Average accuracies are reported for each HOP (row) and each training set size (column).}
    \label{tab:summary_lxmert_scratch_acc_aggregate_filtered-atom-non-ho}
\end{table*}

\begin{table*}[h]
    \small 
    \centering
\begin{tabular}{lrlll}
\hline
 HOP                  &   Diversity & 25k                 & 200k               & 560k                \\
\hline
 cyan cylinder        &          24 & $38.76 \pm 5.93\%$  & $98.14 \pm 1.00\%$ & $99.78 \pm 0.32\%$  \\
 brown sphere         &          24 & $57.37 \pm 4.03\%$  & $97.17 \pm 2.20\%$ & $100.00 \pm 0.00\%$ \\
 red cylinder         &          24 & $60.57 \pm 5.86\%$  & $96.43 \pm 3.01\%$ & $100.00 \pm 0.00\%$ \\
 gray cube            &          24 & $70.16 \pm 2.97\%$  & $93.38 \pm 3.11\%$ & $99.70 \pm 0.28\%$  \\
 purple sphere        &          24 & $57.59 \pm 8.41\%$  & $99.48 \pm 0.74\%$ & $100.00 \pm 0.00\%$ \\
 large cyan object    &          16 & $69.72 \pm 1.99\%$  & $99.56 \pm 0.30\%$ & $100.00 \pm 0.00\%$ \\
 cyan rubber object   &          16 & $61.98 \pm 4.66\%$  & $97.86 \pm 1.69\%$ & $99.96 \pm 0.06\%$  \\
 brown rubber object  &          16 & $68.49 \pm 4.69\%$  & $96.98 \pm 1.46\%$ & $99.17 \pm 0.70\%$  \\
 small brown object   &          16 & $45.16 \pm 7.22\%$  & $93.89 \pm 6.27\%$ & $96.31 \pm 1.35\%$  \\
 red metal object     &          16 & $53.81 \pm 7.51\%$  & $98.93 \pm 0.93\%$ & $98.45 \pm 0.83\%$  \\
 small red object     &          16 & $69.76 \pm 4.76\%$  & $98.41 \pm 0.66\%$ & $99.88 \pm 0.10\%$  \\
 gray metal object    &          16 & $60.52 \pm 9.88\%$  & $93.37 \pm 4.08\%$ & $95.67 \pm 2.43\%$  \\
 large gray object    &          16 & $52.22 \pm 6.42\%$  & $99.17 \pm 0.59\%$ & $98.49 \pm 0.99\%$  \\
 purple rubber object &          16 & $50.12 \pm 6.01\%$  & $97.26 \pm 2.13\%$ & $98.17 \pm 2.00\%$  \\
 small purple object  &          16 & $66.59 \pm 6.63\%$  & $94.25 \pm 0.62\%$ & $96.94 \pm 3.29\%$  \\
 large cylinder       &           6 & $63.96 \pm 11.55\%$ & $98.57 \pm 1.29\%$ & $97.66 \pm 2.14\%$  \\
 rubber cylinder      &           6 & $48.46 \pm 8.28\%$  & $91.89 \pm 3.47\%$ & $80.42 \pm 0.91\%$  \\
 rubber sphere        &           6 & $36.09 \pm 5.65\%$  & $87.04 \pm 6.71\%$ & $84.36 \pm 3.71\%$  \\
 small sphere         &           6 & $57.90 \pm 5.78\%$  & $92.84 \pm 8.36\%$ & $95.91 \pm 1.82\%$  \\
 metal cylinder       &           6 & $53.75 \pm 6.33\%$  & $85.42 \pm 5.21\%$ & $89.99 \pm 2.71\%$  \\
 small cylinder       &           6 & $39.30 \pm 13.90\%$ & $86.47 \pm 7.28\%$ & $82.04 \pm 4.75\%$  \\
 metal cube           &           6 & $54.99 \pm 5.96\%$  & $84.52 \pm 1.40\%$ & $84.97 \pm 4.35\%$  \\
 large cube           &           6 & $46.34 \pm 3.32\%$  & $98.65 \pm 0.49\%$ & $92.34 \pm 6.77\%$  \\
 rubber cube          &           6 & $61.26 \pm 5.22\%$  & $92.67 \pm 2.57\%$ & $83.29 \pm 1.18\%$  \\
 small cube           &           6 & $52.83 \pm 5.33\%$  & $93.68 \pm 4.05\%$ & $95.27 \pm 0.69\%$  \\
 large rubber object  &           4 & $44.57 \pm 10.97\%$ & $89.50 \pm 2.31\%$ & $89.53 \pm 5.20\%$  \\
 small rubber object  &           4 & $51.26 \pm 4.02\%$  & $85.70 \pm 0.79\%$ & $80.11 \pm 3.01\%$  \\
 small metal object   &           4 & $50.97 \pm 6.85\%$  & $88.53 \pm 4.26\%$ & $84.69 \pm 2.55\%$  \\
 large metal object   &           4 & $47.07 \pm 3.88\%$  & $88.61 \pm 3.05\%$ & $88.37 \pm 1.54\%$  \\
\hline
\end{tabular}    \caption{LXMERT (Scratch) \textbf{minimal-OOD} average accuracy and standard deviation over 3 runs with different random seeds. Average accuracies are reported for each HOP (row) and each training set size (column).}
    \label{tab:summary_lxmert_scratch_acc_aggregate_atom-ho}
\end{table*}

\begin{sidewaystable}[h]
    \small 
    \centering
\begin{tabular}{lrlllllll}
\hline
 HOP                 &   Diversity & 25k                & 50k                & 100k               & 200k               & 300k               & 400k               & 560k               \\
\hline
 cyan cylinder       &          24 & $45.01 \pm 0.06\%$ & $45.87 \pm 0.20\%$ & $47.51 \pm 0.10\%$ & $50.08 \pm 0.11\%$ & $56.24 \pm 0.51\%$ & $81.98 \pm 0.88\%$ & $90.43 \pm 0.14\%$ \\
 large cyan object   &          16 & $45.28 \pm 0.08\%$ & $46.26 \pm 0.13\%$ & $47.75 \pm 0.22\%$ & $50.38 \pm 0.21\%$ & $57.26 \pm 0.70\%$ & $82.27 \pm 1.07\%$ & $90.54 \pm 0.22\%$ \\
 cyan rubber object  &          16 & $45.21 \pm 0.07\%$ & $46.09 \pm 0.07\%$ & $47.67 \pm 0.15\%$ & $50.71 \pm 0.30\%$ & $58.21 \pm 1.11\%$ & $83.52 \pm 0.39\%$ & $90.72 \pm 0.15\%$ \\
 large cylinder      &           6 & $45.96 \pm 0.09\%$ & $46.86 \pm 0.11\%$ & $48.11 \pm 0.22\%$ & $50.62 \pm 0.33\%$ & $55.39 \pm 0.25\%$ & $73.21 \pm 2.52\%$ & $90.64 \pm 0.23\%$ \\
 rubber cylinder     &           6 & $45.64 \pm 0.07\%$ & $46.93 \pm 0.20\%$ & $48.40 \pm 0.13\%$ & $51.89 \pm 0.09\%$ & $57.07 \pm 1.20\%$ & $82.07 \pm 1.39\%$ & $90.54 \pm 0.03\%$ \\
 large rubber object &           4 & $46.68 \pm 0.14\%$ & $47.84 \pm 0.13\%$ & $49.62 \pm 0.18\%$ & $52.78 \pm 0.12\%$ & $55.26 \pm 0.24\%$ & $66.30 \pm 3.40\%$ & $89.30 \pm 0.29\%$ \\
\hline
\end{tabular}    \caption{Tensor-NMN \textbf{complex-IID} average accuracy and standard deviation over 3 runs with different random seeds. Average accuracies are reported for each HOP (row) and each training set size (column).}
    \label{tab:summary_nmn_acc_aggregate_val}
\end{sidewaystable}

\begin{sidewaystable}[h]
    \small 
    \centering
\begin{tabular}{lrlllllll}
\hline
 HOP                 &   Diversity & 25k                & 50k                & 100k               & 200k               & 300k               & 400k               & 560k               \\
\hline
 cyan cylinder       &          24 & $46.01 \pm 0.40\%$ & $46.47 \pm 0.82\%$ & $47.57 \pm 0.22\%$ & $49.46 \pm 0.32\%$ & $53.81 \pm 0.16\%$ & $80.31 \pm 0.26\%$ & $90.43 \pm 1.18\%$ \\
 large cyan object   &          16 & $45.50 \pm 0.21\%$ & $45.98 \pm 0.46\%$ & $47.71 \pm 0.14\%$ & $49.44 \pm 0.51\%$ & $53.94 \pm 0.81\%$ & $75.33 \pm 2.55\%$ & $88.26 \pm 2.44\%$ \\
 cyan rubber object  &          16 & $45.57 \pm 0.15\%$ & $46.62 \pm 0.03\%$ & $47.30 \pm 0.26\%$ & $49.70 \pm 0.56\%$ & $54.72 \pm 0.91\%$ & $78.09 \pm 1.29\%$ & $88.30 \pm 0.68\%$ \\
 large cylinder      &           6 & $45.90 \pm 0.15\%$ & $46.42 \pm 0.13\%$ & $47.18 \pm 0.53\%$ & $49.09 \pm 0.49\%$ & $52.61 \pm 0.21\%$ & $67.20 \pm 2.55\%$ & $83.00 \pm 0.86\%$ \\
 rubber cylinder     &           6 & $46.81 \pm 0.17\%$ & $47.48 \pm 0.21\%$ & $48.60 \pm 0.37\%$ & $50.62 \pm 0.20\%$ & $53.99 \pm 0.91\%$ & $73.98 \pm 0.79\%$ & $81.73 \pm 1.59\%$ \\
 large rubber object &           4 & $45.65 \pm 0.35\%$ & $46.67 \pm 0.09\%$ & $47.39 \pm 0.21\%$ & $48.90 \pm 0.12\%$ & $50.63 \pm 0.30\%$ & $56.76 \pm 2.17\%$ & $74.04 \pm 0.71\%$ \\
\hline
\end{tabular}    \caption{Tensor-NMN \textbf{complex-OOD} average accuracy and standard deviation over 3 runs with different random seeds. Average accuracies are reported for each HOP (row) and each training set size (column).}
    \label{tab:summary_nmn_acc_aggregate_test}
\end{sidewaystable}

\begin{sidewaystable}[h]
    \small 
    \centering
\begin{tabular}{lrlllllll}
\hline
 HOP                 &   Diversity & 25k                & 50k                & 100k               & 200k               & 300k               & 400k                & 560k                \\
\hline
 cyan cylinder       &          24 & $52.43 \pm 2.95\%$ & $51.22 \pm 2.45\%$ & $54.24 \pm 1.57\%$ & $60.14 \pm 5.56\%$ & $76.80 \pm 4.99\%$ & $97.75 \pm 1.27\%$  & $100.00 \pm 0.00\%$ \\
 large cyan object   &          16 & $58.06 \pm 2.50\%$ & $60.52 \pm 2.34\%$ & $63.58 \pm 2.52\%$ & $62.37 \pm 3.52\%$ & $80.83 \pm 4.52\%$ & $99.15 \pm 0.49\%$  & $99.99 \pm 0.01\%$  \\
 cyan rubber object  &          16 & $49.69 \pm 2.29\%$ & $46.07 \pm 2.12\%$ & $45.42 \pm 2.43\%$ & $52.02 \pm 1.85\%$ & $75.78 \pm 3.72\%$ & $99.01 \pm 0.72\%$  & $99.93 \pm 0.01\%$  \\
 large cylinder      &           6 & $55.84 \pm 3.04\%$ & $53.73 \pm 0.89\%$ & $57.62 \pm 0.88\%$ & $49.78 \pm 1.33\%$ & $69.48 \pm 5.76\%$ & $97.50 \pm 1.03\%$  & $99.99 \pm 0.00\%$  \\
 rubber cylinder     &           6 & $44.85 \pm 2.73\%$ & $42.56 \pm 5.64\%$ & $43.80 \pm 5.43\%$ & $64.17 \pm 4.29\%$ & $86.06 \pm 7.60\%$ & $96.40 \pm 2.19\%$  & $99.87 \pm 0.07\%$  \\
 large rubber object &           4 & $34.92 \pm 1.40\%$ & $47.56 \pm 5.53\%$ & $41.39 \pm 2.08\%$ & $55.39 \pm 2.52\%$ & $64.32 \pm 2.90\%$ & $75.34 \pm 13.59\%$ & $99.78 \pm 0.14\%$  \\
\hline
\end{tabular}    \caption{Tensor-NMN \textbf{minimal-IID} average accuracy and standard deviation over 3 runs with different random seeds. Average accuracies are reported for each HOP (row) and each training set size (column).}
    \label{tab:summary_nmn_acc_aggregate_filtered-atom-non-ho}
\end{sidewaystable}

\begin{sidewaystable}[h]
    \small 
    \centering
\begin{tabular}{lrlllllll}
\hline
 HOP                 &   Diversity & 25k                 & 50k                 & 100k                & 200k                & 300k                & 400k                & 560k                \\
\hline
 cyan cylinder       &          24 & $67.63 \pm 5.43\%$  & $41.52 \pm 14.81\%$ & $73.36 \pm 1.21\%$  & $50.60 \pm 6.08\%$  & $78.94 \pm 5.26\%$  & $99.11 \pm 0.18\%$  & $99.48 \pm 0.74\%$  \\
 large cyan object   &          16 & $60.24 \pm 15.01\%$ & $74.92 \pm 0.06\%$  & $54.09 \pm 15.68\%$ & $55.36 \pm 16.45\%$ & $64.33 \pm 14.35\%$ & $73.29 \pm 3.71\%$  & $93.57 \pm 6.58\%$  \\
 cyan rubber object  &          16 & $43.17 \pm 12.85\%$ & $37.82 \pm 14.10\%$ & $45.24 \pm 20.55\%$ & $53.29 \pm 14.42\%$ & $53.06 \pm 9.63\%$  & $92.70 \pm 1.56\%$  & $88.29 \pm 12.08\%$ \\
 large cylinder      &           6 & $67.37 \pm 8.26\%$  & $48.28 \pm 14.17\%$ & $65.53 \pm 5.37\%$  & $51.08 \pm 1.88\%$  & $69.62 \pm 19.86\%$ & $83.36 \pm 10.07\%$ & $88.36 \pm 4.50\%$  \\
 rubber cylinder     &           6 & $32.60 \pm 0.65\%$  & $55.30 \pm 9.10\%$  & $53.52 \pm 20.84\%$ & $51.05 \pm 10.23\%$ & $73.15 \pm 8.19\%$  & $87.73 \pm 12.04\%$ & $90.70 \pm 4.84\%$  \\
 large rubber object &           4 & $46.30 \pm 16.02\%$ & $54.62 \pm 20.75\%$ & $72.76 \pm 1.91\%$  & $63.77 \pm 5.81\%$  & $72.88 \pm 4.02\%$  & $79.20 \pm 1.28\%$  & $77.03 \pm 1.35\%$  \\
\hline
\end{tabular}    \caption{Tensor-NMN \textbf{minimal-OOD} average accuracy and standard deviation over 3 runs with different random seeds. Average accuracies are reported for each HOP (row) and each training set size (column).}
    \label{tab:summary_nmn_acc_aggregate_atom-ho}
\end{sidewaystable}

\begin{table*}[h]
    \small 
    \centering
\begin{tabular}{rlll}
\hline
   Diversity & 25k                & 200k                & 560k                \\
\hline
          24 & $0.41 \pm 0.48\%$  & $-0.46 \pm 0.51\%$  & $-0.50 \pm 0.59\%$  \\
          16 & $-0.78 \pm 1.08\%$ & $-1.55 \pm 1.39\%$  & $-1.47 \pm 1.82\%$  \\
           6 & $-2.72 \pm 0.84\%$ & $-7.98 \pm 3.95\%$  & $-8.18 \pm 4.69\%$  \\
           4 & $-6.37 \pm 2.62\%$ & $-12.36 \pm 3.49\%$ & $-13.29 \pm 3.72\%$ \\
\hline
\end{tabular}    \caption{LXMERT (Pretrained) \textbf{complex} systematicity gap (complex-OOD accuracy minus complex-IID accuracy). Average systematicity gap and standard deviation are on the differences, over all 3 runs (with different random seeds) of all HOPs with the stated diversity. Average accuracies are reported for each diversity (row) and each training set size (column).}
    \label{tab:sysgap_lxmert_ft_acc_aggregate_test_val}
\end{table*}

\begin{table*}[h]
    \small 
    \centering
\begin{tabular}{rlll}
\hline
   Diversity & 25k                & 200k                & 560k                \\
\hline
          24 & $-1.51 \pm 5.76\%$ & $-0.54 \pm 0.93\%$  & $-0.93 \pm 2.22\%$  \\
          16 & $-5.74 \pm 4.13\%$ & $-3.42 \pm 2.96\%$  & $-2.97 \pm 4.80\%$  \\
           6 & $-3.74 \pm 7.26\%$ & $-8.80 \pm 7.31\%$  & $-12.22 \pm 7.35\%$ \\
           4 & $-6.74 \pm 6.30\%$ & $-14.89 \pm 5.15\%$ & $-17.59 \pm 6.83\%$ \\
\hline
\end{tabular}    \caption{LXMERT (Pretrained) \textbf{minimal} systematicity gap (minimal-OOD accuracy minus minimal-IID accuracy). Average systematicity gap and standard deviation are on the differences, over all 3 runs (with different random seeds) of all HOPs with the stated diversity. Average accuracies are reported for each diversity (row) and each training set size (column).}
    \label{tab:sysgap_lxmert_ft_acc_aggregate_atom-ho_filtered-atom-non-ho}
\end{table*}

\begin{table*}[h]
    \small 
    \centering
\begin{tabular}{rlll}
\hline
   Diversity & 25k                & 200k                & 560k                \\
\hline
          24 & $0.28 \pm 0.63\%$  & $-1.43 \pm 0.90\%$  & $-0.67 \pm 0.57\%$  \\
          16 & $-0.11 \pm 0.55\%$ & $-2.60 \pm 1.38\%$  & $-1.83 \pm 1.22\%$  \\
           6 & $-0.53 \pm 0.71\%$ & $-8.52 \pm 2.14\%$  & $-9.32 \pm 3.11\%$  \\
           4 & $-2.35 \pm 0.91\%$ & $-12.00 \pm 1.54\%$ & $-13.78 \pm 2.47\%$ \\
\hline
\end{tabular}    \caption{LXMERT (Scratch) \textbf{complex} systematicity gap (complex-OOD accuracy minus complex-IID accuracy). Average systematicity gap and standard deviation are on the differences, over all 3 runs (with different random seeds) of all HOPs with the stated diversity. Average accuracies are reported for each diversity (row) and each training set size (column).}
    \label{tab:sysgap_lxmert_scratch_acc_aggregate_test_val}
\end{table*}

\begin{table*}[h]
    \small 
    \centering
\begin{tabular}{rlll}
\hline
   Diversity & 25k               & 200k               & 560k                \\
\hline
          24 & $3.72 \pm 7.32\%$ & $-1.93 \pm 2.96\%$ & $-0.06 \pm 0.22\%$  \\
          16 & $0.98 \pm 8.62\%$ & $-1.69 \pm 3.84\%$ & $-1.58 \pm 2.18\%$  \\
           6 & $9.12 \pm 8.97\%$ & $-6.78 \pm 7.26\%$ & $-11.03 \pm 7.32\%$ \\
           4 & $5.95 \pm 5.51\%$ & $-7.73 \pm 4.70\%$ & $-14.23 \pm 4.96\%$ \\
\hline
\end{tabular}    \caption{LXMERT (Scratch) \textbf{minimal} systematicity gap (minimal-OOD accuracy minus minimal-IID accuracy). Average systematicity gap and standard deviation are on the differences, over all 3 runs (with different random seeds) of all HOPs with the stated diversity. Average accuracies are reported for each diversity (row) and each training set size (column).}
    \label{tab:sysgap_lxmert_scratch_acc_aggregate_atom-ho_filtered-atom-non-ho}
\end{table*}

\begin{sidewaystable}[h]
    \small 
    \centering
\begin{tabular}{rlllllll}
\hline
   Diversity & 25k                & 50k                & 100k               & 200k               & 300k               & 400k               & 560k                \\
\hline
          24 & $1.00 \pm 0.34\%$  & $0.60 \pm 1.01\%$  & $0.06 \pm 0.26\%$  & $-0.62 \pm 0.21\%$ & $-2.44 \pm 0.35\%$ & $-1.66 \pm 0.69\%$ & $0.00 \pm 1.21\%$   \\
          16 & $0.30 \pm 0.24\%$  & $0.12 \pm 0.50\%$  & $-0.21 \pm 0.33\%$ & $-0.98 \pm 0.36\%$ & $-3.41 \pm 0.42\%$ & $-6.18 \pm 1.67\%$ & $-2.35 \pm 1.87\%$  \\
           6 & $0.56 \pm 0.64\%$  & $0.05 \pm 0.54\%$  & $-0.37 \pm 0.70\%$ & $-1.40 \pm 0.46\%$ & $-2.93 \pm 0.56\%$ & $-7.05 \pm 1.14\%$ & $-8.22 \pm 1.48\%$  \\
           4 & $-1.03 \pm 0.40\%$ & $-1.17 \pm 0.05\%$ & $-2.23 \pm 0.27\%$ & $-3.87 \pm 0.02\%$ & $-4.63 \pm 0.39\%$ & $-9.54 \pm 1.49\%$ & $-15.27 \pm 0.75\%$ \\
\hline
\end{tabular}    \caption{Tensor-NMN \textbf{complex} systematicity gap (complex-OOD accuracy minus complex-IID accuracy). Average systematicity gap and standard deviation are on the differences, over all 3 runs (with different random seeds) of all HOPs with the stated diversity. Average accuracies are reported for each diversity (row) and each training set size (column).}
    \label{tab:sysgap_nmn_acc_aggregate_test_val}
\end{sidewaystable}

\begin{sidewaystable}[h]
    \small 
    \centering
\begin{tabular}{rlllllll}
\hline
   Diversity & 25k                 & 50k                 & 100k                & 200k                & 300k                & 400k                 & 560k                \\
\hline
          24 & $15.20 \pm 6.08\%$  & $-9.70 \pm 14.59\%$ & $19.12 \pm 1.31\%$  & $-9.54 \pm 2.88\%$  & $2.15 \pm 10.09\%$  & $1.36 \pm 1.39\%$    & $-0.52 \pm 0.74\%$  \\
          16 & $-2.17 \pm 16.24\%$ & $3.07 \pm 14.40\%$  & $-4.84 \pm 18.92\%$ & $-2.87 \pm 15.02\%$ & $-19.61 \pm 9.83\%$ & $-16.08 \pm 10.18\%$ & $-9.03 \pm 10.07\%$ \\
           6 & $-0.36 \pm 12.87\%$ & $3.64 \pm 14.73\%$  & $8.82 \pm 11.73\%$  & $-5.91 \pm 8.65\%$  & $-6.38 \pm 13.92\%$ & $-11.40 \pm 10.38\%$ & $-10.40 \pm 4.82\%$ \\
           4 & $11.38 \pm 16.64\%$ & $7.06 \pm 15.89\%$  & $31.37 \pm 3.85\%$  & $8.37 \pm 3.59\%$   & $8.56 \pm 3.68\%$   & $3.86 \pm 12.58\%$   & $-22.75 \pm 1.36\%$ \\
\hline
\end{tabular}    \caption{Tensor-NMN \textbf{minimal} systematicity gap (minimal-OOD accuracy minus minimal-IID accuracy). Average systematicity gap and standard deviation are on the differences, over all 3 runs (with different random seeds) of all HOPs with the stated diversity. Average accuracies are reported for each diversity (row) and each training set size (column).}
    \label{tab:sysgap_nmn_acc_aggregate_atom-ho_filtered-atom-non-ho}
\end{sidewaystable}

%% file: datasheet.tex
\definecolor{darkblue}{RGB}{46,25, 110}

\newcommand{\dssectionheader}[1]{%
   \noindent\framebox[\columnwidth]{%
      {\fontfamily{phv}\selectfont \textbf{\textcolor{darkblue}{#1}}}
   }
}

\newcommand{\dsquestion}[1]{%
    {\noindent \fontfamily{phv}\selectfont \textcolor{darkblue}{\textbf{#1}}}
}

\newcommand{\dsquestionex}[2]{%
    {\noindent \fontfamily{phv}\selectfont \textcolor{darkblue}{\textbf{#1} #2}}
}

\newcommand{\dsanswer}[1]{%
   {\noindent #1 \medskip}
}

\dssectionheader{Motivation for Dataset Creation}

\dsquestionex{Why was the dataset created?}{(e.g., were there specific tasks in mind, or a specific gap that needed to be filled?)}

\dsanswer{The \dataset diagnostic dataset was created to study systematicity with respect to held-out pairs of attribute values in a controlled setting. These held-out pairs include various color-shape, color-material, color-size, size-shape, size-material, and shape-material pairs; each of the 29 pairs has a dedicated train set and four dedicated test sets. The specific task is visual question answering (VQA), in the form of 28-way classification. 

To the best of the author's knowledge, this was a specific gap that needed to be filled. The closest prior work is the CLEVR-CoGenT dataset: ~\citet{johnson2017clevr} created a train-test CLEVR split where at train time 
cubes and cylinders are restricted to limited color palettes, that are
reversed at test time. Unlike \dataset, CLEVR-CoGenT does not change the question distribution at train time --- held-out combinations can leak by appearing in text at train time.  
Furthermore, CLEVR-CoGenT has only a single train set with held-out \attType{color}-\attType{shape} combinations --- whereas \dataset expands the set of held-out combinations to 29 train sets, covering all possible pairs of attribute types.  
\dataset also independently assesses each \ho, including in a minimal setting.  
In combination, these improvements allows the use of \dataset to study the impact of train-time diversity on systematicity.}

\dsquestionex{What (other) tasks could the dataset be used for?}{Are there obvious tasks for which it should not be used?}

\dsanswer{
\dataset can also be useful for studying model transfer from another domain (e.g., natural images) to the synthetic CLEVR domain. \dataset is a diagnostic dataset only, it is not intended as a thorough evaluation of a model's systematicity.
}

\dsquestionex{Has the dataset been used for any tasks already?}{If so, where are the results so others can compare (e.g., links to published papers)?}

\dsanswer{
\dataset has only been used in this paper. A GitHub repo for recording works using this dataset will be provided. It is redacted at present to preserve anonymity.
}

\dsquestionex{Who funded the creation of the dataset?}{If there is an associated grant, provide the grant number.}

\dsanswer{
Resources used in preparing this research were provided, in part, by the Department of Computer Science at the University of Toronto, the Province of Ontario, the Government of Canada through CIFAR, companies sponsoring the Vector Institute (\url{www.vectorinstitute.ai/partnerships/current-partners/}), the Hyundai Motor Company (under the project Uncertainty in Neural Sequence Modeling), the Samsung Advanced Institute of Technology (under the project Next Generation Deep Learning: From Pattern Recognition to AI), and by a gift from the Chan Zuckerberg Initiative Foundation to establish the Kempner Institute for the Study of Natural and Artificial Intelligence.

Ian Berlot-Attwell is funded by a Natural Sciences and Engineering Research Council of Canada Postgraduate Scholarship-Doctoral, and a Vector Institute Research Grant. A. Michael Carrell is funded in part by a Microsoft Research scholarship. The authors thank the International Max Planck Research School for Intelligent Systems (IMPRS-IS) for supporting Yash Sharma. 
}

\dsquestion{Any other comments?}
N/A

\bigskip
\dssectionheader{Dataset Composition}

\dsquestionex{What are the instances?}{(that is, examples; e.g., documents, images, people, countries) Are there multiple types
of instances? (e.g., movies, users, ratings; people, interactions between them; nodes, edges)}

\dsanswer{
Each instance is comparable to a CLEVR instance. i.e., each instance consists of an image (a rendered blender scene of colored blocks on a plain background in the style of the CLEVR dataset), an English question, and a 1-word answer (there are 28 possible answers, exactly the same as in the original CLEVR). Scene graphs and the question's corresponding functional program (specified with the CLEVR question primitives) are also provided.

For each of the 29 held-out pairs (HOPs) in \dataset, train instances are of comparable complexity to CLEVR and do not contain the HOP in the image, or the question. 

Of the four test sets: The \hoVal and \hoTest sets have images and questions of comparable complexity to CLEVR. The \atomHo and \atomNonHo sets contain minimal examples; the images are of only a single object, and the questions ask whether there is an object in the scene matching a specific pair of attribute values -- e.g., ``Are there any rubber cylinders?''. Of these four test sets, the IID sets are like the train set in that the images and questions do not contain the HOP. The OOD test sets contain the HOP in both the question, and in at least one object in the image.

For more details see Sections~\ref{sec:dataset} and \ref{sec:dataset_details}. Example images and questions are visualized in~\Cref{fig:hope}.
}

\dsquestionex{Are relationships between instances made explicit in the data}{ (e.g., social network links, user/movie ratings, etc.)?}

\dsanswer{
The only relationships between instances are that some instances re-use images (see~\Cref{sec:dataset_details} for further details), and some instances use questions generated from the same base template.
In both cases, these relationships are available in the data. Instances reusing images refer to the same image index, and each question records its question family, as in CLEVR. 
}

\dsquestion{How many instances of each type are there?}

\dsanswer{
For each of the 29 held-out pairs (HOPs) in \dataset, the approximate size of the corresponding splits is outlined below:

\begin{itemize}
    \item \train set: 62k images, and 560k image-question pairs
    \item \hoVal set: 13k images, 120k image-question pairs
    \item \hoTest set: 15k images, 15k image-question pairs
    \item \atomNonHo set: 2576-3200 images, 8640-11970 image-question pairs (depending on \ho)
    \item \atomHo set: 448-3840 images, 448-3840 image-question pairs (depending on \ho) 
\end{itemize}
}

In general, for every HOP, each image in the \train, and \hoVal has 9 matching questions. Each image in \hoTest has 1 corresponding question. 

The number of questions per image for \atomNonHo and \atomHo varies depending on the HOP -- see Section \ref{sec:dataset_details} for details on the construction of the \atomNonHo and \atomHo datasets.

\dsquestionex{What data does each instance consist of?}{ “Raw” data
(e.g., unprocessed text or images)? Features/attributes?
Is there a label/target associated with instances? If the
instances are related to people, are subpopulations identified (e.g., by age, gender, etc.) and what is their distribution?}

\dsanswer{
For every instance, the image is a $320\times 480$ pixels. Images are 
aggregated over all HOPs in three HDF5 files (corresponding to train, IID test sets, and OOD test sets, respectively), which can be easily converted back to individual images in the PNG format.

The scene graphs are represented as .json files, following the CLEVR specification.

Questions, programs, and answer labels are provided in HDF5 files. Functional programs are encoded as a sequence of integers, the vocabulary mapping these integers to their English equivalents is provided in a JSON file. Questions are similarly encoded. Questions have undergone minimal tokenization, and the raw English questions are available in a separate JSON file. The only tokenization performed is the treating of ``,'' and ``;'' as separate tokens, the removal of ``.'' and ``?'' characters, and separation by white space. Answers are encoded as a single integer; the mapping to English is again in the JSON vocab file.

Instances are not related to people.
}

\dsquestionex{Is everything included or does the data rely on external
resources?}{ (e.g., websites, tweets, datasets) If external
resources, a) are there guarantees that they will exist, and
remain constant, over time; b) is there an official archival
version. Are there licenses, fees or rights associated with
any of the data?}

\dsanswer{
\dataset does not rely on external resources.
}

\dsquestionex{Are there recommended data splits or evaluation measures?}{ (e.g., training, development, testing; accuracy/AUC)}

\dsanswer{
The dataset comes with recommended train/test splits that ensure no images are shared between the train and test splits, and the held-out pair only occurs in given test sets. It is recommended that hyperparameter tuning be done on the original CLEVR dataset. The intended evaluation is to report accuracy.
}

\dsquestionex{What experiments were initially run on this dataset?}{Have a summary of those results and, if available, provide
the link to a paper with more information here.}

Initial experiments were the fitting of LXMERT (both finetuned, and from scratch) on each of the 29 held-out pairs. Tensor-NMN was also fit to the first 6 HOPs. Models were trained using the full training set (560k image-question pairs), as well as subsets of size 25k and 200k. 

In all cases, models exhibited some degree of systematicity, but performance degraded on OOD test sets. Furthermore, studying the systematicity gap (the difference between OOD and IID test performance) it was clear that the systematicity gap narrrowed as the train-time diversity of the HOP (i.e., the number of pairs of the same attribute types but different values) increased. See Sections~\ref{sec:evidence_of_sys} and \ref{sec:sysGap} for details.

\bigskip
\dssectionheader{Data Collection Process}

\dsquestionex{How was the data collected?}{ (e.g., hardware apparatus/sensor, manual human curation, software program, software interface/API; how were these constructs/measures/methods validated?)}

\dsanswer{
Data was generated via computer program. The code was modified from the original CLEVR codebase, and tested via code review among the authors, and manual inspection of the output.
}

\dsquestionex{Who was involved in the data collection process? }{(e.g.,
students, crowdworkers) How were they compensated? (e.g.,
how much were crowdworkers paid?)}

\dsanswer{
N/A: Only the authors were involved.
}

\dsquestionex{Over what time-frame was the data collected?}{ Does the
collection time-frame match the creation time-frame?
How was the data associated with each instance acquired? Was the data directly observable (e.g., raw text,
movie ratings), reported by subjects (e.g., survey responses),
or indirectly inferred/derived from other data (e.g., part of
speech tags; model-based guesses for age or language)? If
the latter two, were they validated/verified and if so how?
Does the dataset contain all possible instances? Or is
it, for instance, a sample (not necessarily random) from a
larger set of instances?}

\dsanswer{
N/A: The data was generated by python program, and the images rendered with Blender 2.7.
}

\dsquestionex{If the dataset is a sample, then what is the population?}{
What was the sampling strategy (e.g., deterministic, probabilistic with specific sampling probabilities)? Is the sample
representative of the larger set (e.g., geographic coverage)?
If not, why not (e.g., to cover a more diverse range of instances)? How does this affect possible uses?}

\dsanswer{
For each of the 29 HOPs:

For the \train, and \hoVal the full population of images is the space of all valid CLEVR images such that no object matches the HOP (e.g., if the HOP is rubber cylinder, then there must be no rubber cylinders in the scene). The \hoTest population of images is valid CLEVR images such that at least one object matches the HOP. The \atomHo and \atomNonHo are similar to \hoVal and \hoTest respectively, but always have exactly 1 object in the scene.

The key constraints that valid CLEVR images must meet are that at least 100 pixels of each object must be visible, and that there must be 3-10 objects in the scene. 

The sampling of images was probabilistic, uniformly at random. 

The space of questions is the space of all instantiations of the CLEVR templates that produce well-formed questions (the key constraint being that questions are unambiguously answerable from the scenegraph and the functional form of the question).
The sampling method was probabilistic in all cases. Following CLEVR, question templates were sampled randomly, and instantiations found via depth first search with randomized ordering of possibilities. Following CLEVR, sampling probabilities shift over time to encourage distribution balance with respect to question templates.
}

\dsquestionex{Is there information missing from the dataset and why?}{
(this does not include intentionally dropped instances; it
might include, e.g., redacted text, withheld documents) Is
this data missing because it was unavailable?}

\dsanswer{
No.
}

\dsquestion{Are there any known errors, sources of noise, or redundancies in the data?}

\dsanswer{
No.
}

\bigskip
\dssectionheader{Data Preprocessing}

\dsquestionex{What preprocessing/cleaning was done?}{ (e.g., discretization or bucketing, tokenization, part-of-speech tagging, SIFT feature extraction, removal of instances, processing of
missing values, etc.)}

\dsanswer{
The English questions were tokenized.  The only tokenization performed is the treating of ``,'' and ``;'' as separate tokens, the removal of ``.'' and ``?'' characters, and separation by white space. Capitalization was not changed.
}

\dsquestionex{Was the “raw” data saved in addition to the preprocessed/cleaned data?}{ (e.g., to support unanticipated future uses)}

\dsanswer{
Yes.
}

\dsquestion{Is the preprocessing software available?}

\dsanswer{
Yes, the same tokenization as \cite{DBLP:conf/iccv/JohnsonHMHFZG17} was used.
}

\dsquestion{Does this dataset collection/processing procedure
achieve the motivation for creating the dataset stated
in the first section of this datasheet?}

\dsanswer{
Yes, for each of the 29 held-out pairs, we have a train set that does not contain the HOP, and test sets of minimal and comparable complexity that do or do not contain the HOP. Thus we can asses the systematicity of a model, as well as how the systematicity is affected by the exact HOP, the amount of training data, and the complexity of test data.
}

\bigskip
\dssectionheader{Dataset Distribution}

\dsquestionex{How is the dataset distributed?}{ (e.g., website, API, etc.;
does the data have a DOI; is it archived redundantly?)}

\dsanswer{

Distribution details are on the paper's official repository: \url{https://github.com/ikb-a/systematicity-gap-in-vqa}. The data is not archived redundantly.
}

\dsquestionex{When will the dataset be released/first distributed?}{ (Is
there a canonical paper/reference for this dataset?)}

\dsanswer{
\dataset will be released with the publication of this paper.
}

\dsquestionex{What license (if any) is it distributed under?}{ Are there any copyrights on the data?}

\dsanswer{

\dataset is shared under a \href{https://creativecommons.org/licenses/by/4.0/}{Creative Commons CC BY 4.0} license. 

Note that \dataset contains images from the original CLEVR dataset \cite{johnson2017clevr} which is also shared under a CC BY 4.0 license, and \dataset was created using a modified version of the CLEVR generation code which was shared under a \href{https://github.com/facebookresearch/clevr-dataset-gen/blob/f0ce2c81750bfae09b5bf94d009f42e055f2cb3a/LICENSE}{BSD license}.
}

\dsquestion{Are there any fees or access/export restrictions?}

\dsanswer{
No.
}

\bigskip
\dssectionheader{Dataset Maintenance}

\dsquestion{Who is supporting/hosting/maintaining the dataset?}

\dsanswer{
Hosting TBD, the details will be on the paper's official repository \url{https://github.com/ikb-a/systematicity-gap-in-vqa}. The lead author is maintaining the dataset.
}

\dsquestionex{How does one contact the owner/curator/manager of the dataset}{ (e.g. email address, or other contact info)?}

\dsanswer{
Contact the lead author via email at \href{mailto:ianberlot@cs.toronto.edu}{ianberlot@cs.toronto.edu}.
}

\dsquestionex{Will the dataset be updated?}{ How often and by whom? How will updates/revisions be documented and communicated (e.g., mailing list, GitHub)? Is there an erratum?}

\dsanswer{
There are no plans for the dataset to be updated. If needed, it will be updated by the lead author, and changes documented via GitHub.
}

\dsquestion{If the dataset becomes obsolete how will this be communicated?}

\dsanswer{
The GitHub page will be updated to reflect this.
}

\dsquestion{Is there a repository to link to any/all papers/systems that use this dataset?}

\dsanswer{
Works using this work can be linked to on this page in the repository: \url{https://github.com/ikb-a/systematicity-gap-in-vqa/blob/main/FOLLOWUP.md}.
}

\dsquestionex{If others want to extend/augment/build on this dataset,
is there a mechanism for them to do so?}{ If so, is there
a process for tracking/assessing the quality of those contributions. What is the process for communicating/distributing
these contributions to users?}

\dsanswer{
There is no provided mechanism, but they are free to do so under the license, and enouraged to do so by the authors.
}

\dsquestion{Any other comments?}

\dsanswer{
Due to the size of the dataset, we are currently exploring hosting options.
}

\bigskip
\dssectionheader{Legal \& Ethical Considerations}

\dsquestionex{If the dataset relates to people (e.g., their attributes) or
was generated by people, were they informed about the
data collection?}{ (e.g., datasets that collect writing, photos, interactions, transactions, etc.)}

\dsanswer{
N/A
}

\dsquestionex{If it relates to other ethically protected subjects, have
appropriate obligations been met?}{ (e.g., medical data
might include information collected from animals)
If it relates to people, were there any ethical review applications/reviews/approvals? (e.g. Institutional Review
Board applications)}

\dsanswer{
N/A
}

\dsquestionex{If it relates to people, were they told what the dataset
would be used for and did they consent?}{ What community norms exist for data collected from human communications? If consent was obtained, how? Were the people
provided with any mechanism to revoke their consent in the
future or for certain uses?}

\dsanswer{
N/A
}

\dsquestionex{If it relates to people, could this dataset expose people
to harm or legal action? }{ (e.g., financial social or otherwise) What was done to mitigate or reduce the potential for harm?}

\dsanswer{
N/A
}

\dsquestionex{If it relates to people, does it unfairly advantage or disadvantage a particular social group? }{In what ways? How was this mitigated?}

\dsanswer{
N/A
}

\dsquestionex{If it relates to people, were they provided with privacy
guarantees? }{If so, what guarantees and how are these
ensured?}

\dsanswer{
N/A
}

\dsquestionex{Does the dataset comply with the EU General Data Protection Regulation (GDPR)? }{Does it comply with any other
standards, such as the US Equal Employment Opportunity
Act?}

\dsanswer{
N/A

}

\dsquestionex{Does the dataset contain information that might be considered sensitive or confidential? }{(e.g., personally identifying information)}

\dsanswer{
No.
}

\dsquestion{Does the dataset contain information that might be considered inappropriate or offensive?}

\dsanswer{No.}